\documentclass{article}

\usepackage{microtype}
\usepackage{graphicx}
\usepackage{subfigure}
\usepackage{booktabs} 
\usepackage{enumitem}
\usepackage{subcaption}

\usepackage{hyperref}

\usepackage{tabularx}

\usepackage[accepted]{icml2024}

\usepackage{algorithm}
\usepackage[noend]{algpseudocode}
\renewcommand{\algorithmicrequire}{\textbf{Input:}}
\renewcommand{\algorithmicensure}{\textbf{Output:}}

\usepackage{amsmath}
\usepackage{amssymb}
\usepackage{mathtools}
\usepackage{amsthm}

\usepackage[capitalize,noabbrev]{cleveref}

\theoremstyle{plain}

\theoremstyle{definition}

\theoremstyle{remark}

\newcommand{\crossprompt}{\textsc{LocoGen}}
\newcommand{\crossedit}{\textsc{LocoEdit}}

\newcommand{\clean}{\text{clean}}
\newcommand{\altered}{\text{altered}}

\usepackage{hyperref}
\usepackage{cleveref}
\crefname{figure}{Fig}{Figs}%
\crefname{algorithm}{Algorithm}{Algo}%

\usepackage{booktabs}
\usepackage{esrelation}
\usepackage{tikz}
\usepackage{graphics}

\usepackage{graphicx}
\usepackage{xcolor}
\usepackage{amsfonts} 
\usepackage{cleveref} 
\usepackage{amsmath}
\usepackage{siunitx}
\usepackage{physics}
\AtBeginDocument{\RenewCommandCopy\qty\SI}

\usepackage{newfloat}
\usepackage{listings}
\usepackage{color, colortbl}

\usepackage[textsize=tiny]{todonotes}

\icmltitlerunning{Preprint}

\DeclareMathOperator*{\argmin}{arg\,min}

\begin{document}

\twocolumn[
\icmltitle{On Mechanistic Knowledge Localization in Text-to-Image Generative Models}



\icmlsetsymbol{equal}{*}

\begin{icmlauthorlist}
\icmlauthor{Samyadeep Basu}{equal,yyy}
\icmlauthor{Keivan Rezaei}{equal,yyy}
\icmlauthor{Priyatham Kattakinda}{yyy}
\icmlauthor{Ryan Rossi}{comp}
\icmlauthor{Cherry Zhao}{comp}
\icmlauthor{Vlad Morariu}{comp}
\icmlauthor{Varun Manjunatha}{comp}
\icmlauthor{Soheil Feizi}{yyy}
\end{icmlauthorlist}

\icmlaffiliation{yyy}{University of Maryland}
\icmlaffiliation{comp}{Adobe Research}

\icmlcorrespondingauthor{Samyadeep Basu}{sbasu12@umd.edu}

\icmlkeywords{Machine Learning, ICML}

\vskip 0.3in
]


\printAffiliationsAndNotice{\icmlEqualContribution} 

\begin{abstract}
\vspace{-0.1cm}
Identifying layers within text-to-image models which control visual attributes can facilitate efficient model editing through closed-form updates.
Recent work,
leveraging causal tracing show that early Stable-Diffusion variants confine knowledge primarily to the first layer of the CLIP text-encoder,
while it diffuses throughout the UNet.
Extending this framework, we observe that for recent models (e.g., SD-XL, DeepFloyd), causal tracing fails in pinpointing localized knowledge,
highlighting challenges in model editing.
To address this issue, we introduce the concept of {\it mechanistic localization} in text-to-image models,
where knowledge about various visual attributes (e.g., ``style", ``objects", ``facts") can be {\it mechanistically} localized
to a small fraction of layers in the UNet, thus facilitating efficient model editing.
We localize knowledge using our method~\crossprompt{} which measures the direct effect of intermediate layers to output generation by performing interventions in the cross-attention layers of the UNet. 
We then employ \crossedit{}, a fast closed-form editing method across popular open-source text-to-image models (including the latest SD-XL)
and explore the possibilities of neuron-level model editing.
Using {\it mechanistic localization}, our work offers a better view of successes and failures in localization-based text-to-image model editing. Code will be available at \href{https://github.com/samyadeepbasu/LocoGen}{https://github.com/samyadeepbasu/LocoGen}.

\end{abstract}

\section{Introduction}
In recent years, substantial strides in conditional image generation have been made through diffusion-based text-to-image generative models, including notable examples like Stable-Diffusion~\citep{sd_main}, Imagen~\citep{saharia2022photorealistic}, and DALLE~\citep{dalle}. These models have captured widespread attention owing to their impressive image generation and editing capabilities, as evidenced by leading FID scores on prominent benchmarks such as MS-COCO~\citep{coco}. Typically trained on extensive billion-scale image-text pairs like LAION-5B~\citep{schuhmann2022laion5b}, these models encapsulate a diverse array of visual concepts, encompassing color, artistic styles, objects, and renowned personalities.

\begin{figure*}
    \hskip 0.0cm
  \includegraphics[width=17.5cm, height=5.4cm]{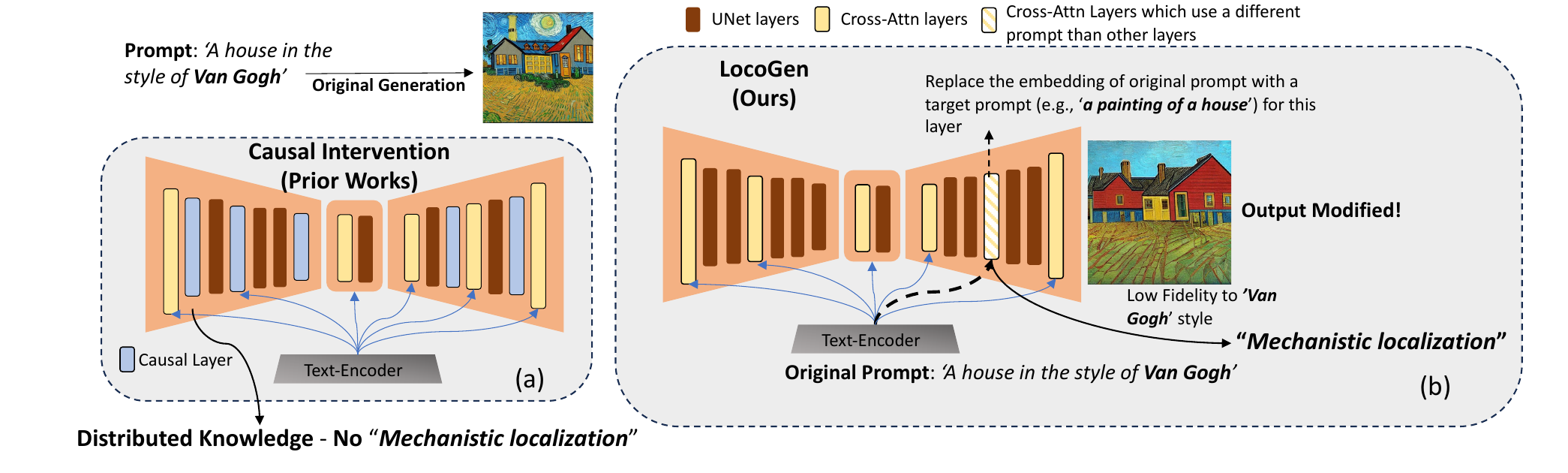}
  \vspace{-0.5cm}
    \caption{\label{teaser}
    \textbf{\crossprompt: Identifying UNet layers that, when given different input, can alter visual attributes (e.g., style, objects, facts).}
    (a) Earlier works~\citep{basu2023localizing} which show distributed knowledge using causal interventions. (b) \crossprompt{} where a few cross-attention layers receive a different prompt-embedding than the original, leading to generation of images without the particular style.
    }%
    \vspace{-0.3cm}
\end{figure*}

A recent work~\citep{basu2023localizing} designs an interpretability framework
using causal tracing~\citep{pearl2013direct} to trace the location of knowledge about various styles, objects or facts in text-to-image generative models. 
Essentially, causal tracing finds the indirect effects of intermediate layers~\citep{pearl2013direct},
by finding layers which can restore a model with corrupted inputs to its original state.
Using this framework, the authors find that knowledge about various visual attributes is distributed in the UNet,
whereas, there exists a unique causal state in the CLIP text-encoder where knowledge is localized.
This unique causal state in the text-encoder can be leveraged to edit text-to-image models in order to remove style, objects or update facts effectively. 
However, we note that their framework is restricted to early Stable-Diffusion variants such as Stable-Diffusion-v1-5. 

In our paper, we first revisit knowledge localization for text-to-image generative models,
specifically examining the effectiveness of causal tracing beyond Stable-Diffusion-v1-5.
While causal tracing successfully identifies unique localized states in the text-encoder for Stable-Diffusion variants,
including v1-5 and v2-1, it fails to do so for recent models like SD-XL \cite{podell2023sdxl} and DeepFloyd\footnote{https://github.com/deep-floyd/IF} across different visual attributes.
In the UNet, causal states are distributed across a majority of open-source text-to-image models (excluding DeepFloyd), aligning with findings in~\citet{basu2023localizing}.
Notably, for DeepFloyd, we observe a lack of strong causal states corresponding to visual attributes in the UNet.

To address the \textit{universal} knowledge localization framework absence across different text-to-image models,
we introduce the concept of {\it mechanistic localization} that aims to identify a small number of layers which control the generation of distinct visual attributes,
across a spectrum of text-to-image models. 
To achieve this, we propose~\crossprompt{}, a method that finds a subset of cross-attention layers in the UNet such that when the input to their key and value matrices is changed,
output generation for a given visual attribute (e.g., ``style") is modified (see Figure~\ref{teaser}).
This intervention in the intermediate layers has a direct effect on the output -- therefore~\crossprompt \hspace{0.05cm} measures the direct effect of intermediate layers, as opposed to indirect effects in causal tracing. 

Leveraging~\crossprompt{}, we probe knowledge locations for different visual attributes across popular open-source text-to-image models such as Stable-Diffusion-v1, Stable-Diffusion-v2, OpenJourney\footnote{https://huggingface.co/prompthero/openjourney}, SD-XL~\citep{podell2023sdxl} and DeepFloyd.
For all models, we find that unique locations can be identified for visual attributes (e.g., ``style", ``objects", ``facts").  
Using these locations, we then perform weight-space model editing to remove artistic ``styles", modify trademarked ``objects" and update outdated ``facts" in text-to-image models.
This weight-space editing is performed using~\crossedit{}  which updates the key and value matrices using a closed-form update in the locations identified by \crossprompt.
Moreover, for certain attributes such as ``style", we show that knowledge can be traced and edited to a subset of neurons, therefore highlighting the possibilities of neuron-level model editing.
\paragraph{Contributions.} In summary, our contributions include:
\begin{itemize}
    \item  We highlight the drawbacks of existing interpretability methods such as causal tracing for localizing knowledge in latest text-to-image models.
    \item We introduce~\crossprompt{} which can universally identify layers that control for visual attributes across a large spectrum of open-source text-to-image models.  
    \item By examining edited models using \crossedit{} along with \crossprompt, we observe that this efficient approach is successful across a majority of text-to-image models.
\end{itemize}
\label{submission}
\vspace{-0.4cm}
\section{Related Works}
\textbf{Intepretability of Text-to-Image Models. } 
To our understanding, there's limited exploration into the inner workings of text-to-image models, such as Stable-Diffusion. DAAM~\citep{tang2022daam, hertz2022prompttoprompt} scrutinizes diffusion models through the analysis of cross-attention maps between text tokens and images, highlighting their semantic precision. ~\citep{chefer2023hidden} understand the decomposition of concepts in diffusion models. ~\citep{basu2023localizing} leverage causal tracing to understand how knowledge is stored in text-to-image models such as Stable-Diffusion-v1. 

\textbf{Editing Text-to-Image Models.}  
The capacity to modify a diffusion model's behavior without starting from scratch was initially investigated in Concept-Ablation~\citep{kumari2023ablating} and Concept-Erasure~\citep{gandikota2023erasing}. Another method, TIME~\citep{orgad2023editing}, alters all the cross-attention layers' key and value matrices to translate between concepts, though lacks interpretability and applications on a real-use case of model editing. ~\citep{basu2023localizing} edits text-to-image models in the text-encoder space by leveraging a singular causal state. 
However, existing works overlook newer text-to-image models (e.g., SD-XL and DeepFloyd), which we delve into in detail.
\section{Preliminaries}
Diffusion models start with an initial random real image $\vb{x}_{0}$, the noisy image at time step $t$ is expressed as
$\vb{x}_{t} = \sqrt{\alpha_{t}}\vb{x}_{0} + \sqrt{(1-\alpha_{t})}\vb{\epsilon}$.
Here, $\alpha_{t}$ determines the strength of the random Gaussian noise,
gradually diminishing as the time step increases,
ensuring that $\vb{x}_{T}\sim \mathcal{N}(0, I)$.
The denoising network $\epsilon_{\theta}(\vb{x}_{t}, \vb{c}, t)$, is pre-trained to denoise the noisy image $\vb{x}_{t}$ and produce $\vb{x}_{t-1}$.
Typically, the conditional input $\vb{c}$ for the denoising network $\epsilon_{\theta}(.)$ is a text-embedding derived from a caption $c$ through a text-encoder,
denoted as $\vb{c} = v_{\gamma}(c)$. 
\begin{figure}
    \hskip 0.3cm
    \centering
  \includegraphics[width=0.8\columnwidth]{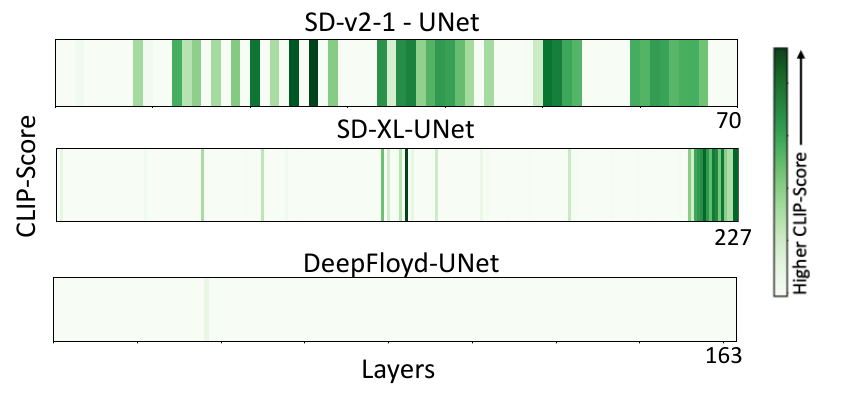}
  \vspace{-0.3cm}
    \caption{\label{causal_trace_unet}\textbf{Causal tracing for UNet. } Similar to~\citep{basu2023localizing}, we find that knowledge is causally distributed across the UNet for text-to-image models such as SD-v2-1 and SD-XL. For DeepFloyd we do not observe any significant causal state in the UNet.}%
    \vspace{-0.2cm}
\end{figure}
\begin{figure}
    \hskip 0.4cm
    \centering
  \includegraphics[width=0.8\columnwidth]{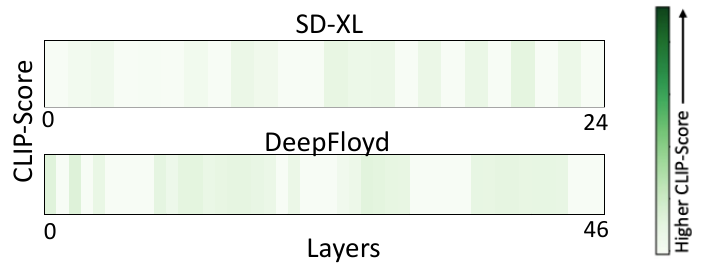}
  \vspace{-0.3cm}
    \caption{\label{causal_trace_text}\textbf{Causal tracing for text-encoder.} Unlike SD-v1-5 and SD-v2-1, we find that a singular causal states does not exist in the text-encoder for SD-XL and DeepFloyd.}%
    \vspace{-0.6cm}
\end{figure}
The noising as well as the denoising operation can also occur in a latent space defined by $\vb{z} = \mathcal{E}(\vb{x})$~\cite{sd_main} for better efficiency. The pre-training objective learns to denoise in the latent space as denoted by: 
\begin{align*}
    \mathcal{L}(\vb{z}, \vb{c}) = \mathbb{E}_{\epsilon, t} || \epsilon - \epsilon_{\theta}(\vb{z}_{t}, \vb{c}, t) ||_{2}^{2},
\end{align*}
where $\vb{z}_{t} = \mathcal{E}(\vb{x}_{t})$ and $\mathcal{E}$ is an encoder such as VQ-VAE~\citep{oord2018neural}. 
\begin{figure*}
    \hskip 0.2cm
    \includegraphics[width=17cm, height=10cm]{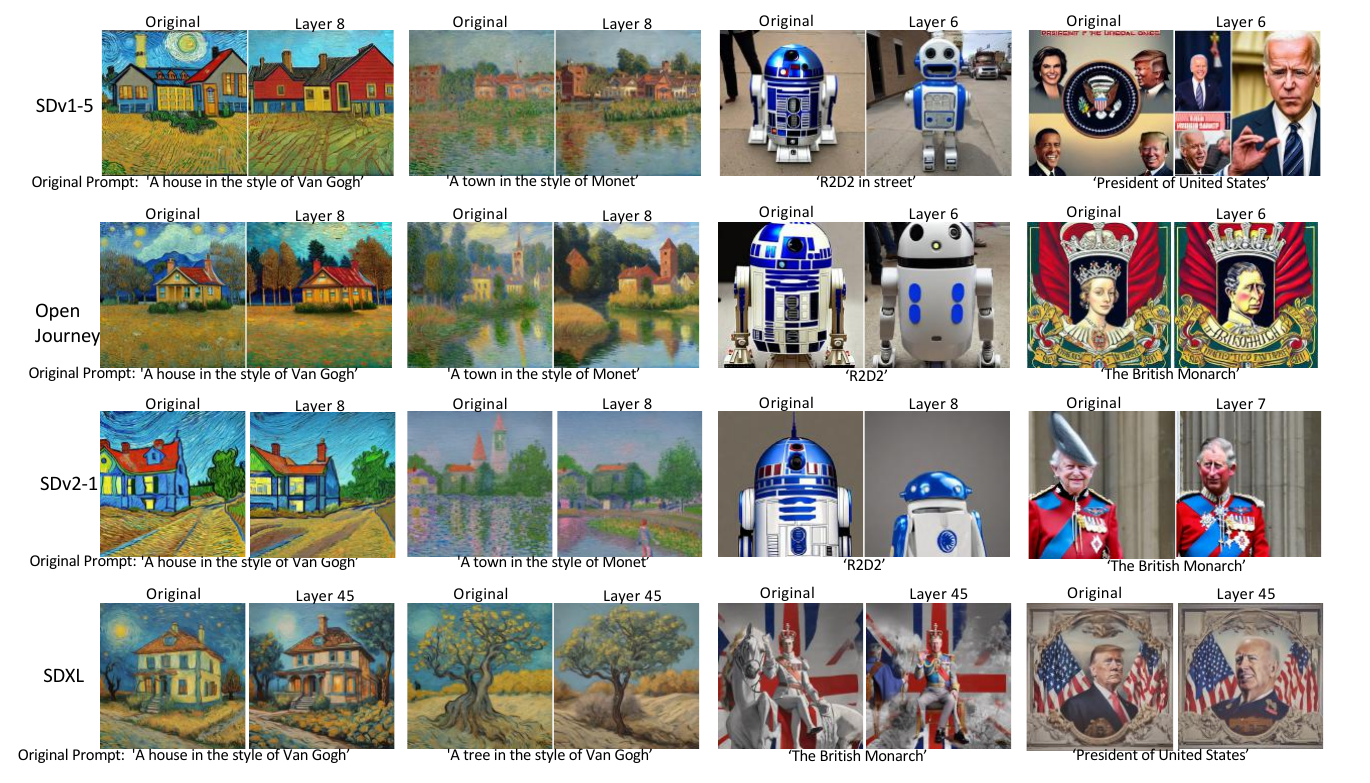}
    \vspace{-0.8cm}
    \caption{\label{viz_prompt} \textbf{Interpretability Results: Images generated by intervening on the layers identified by~\crossprompt \hspace{0.05cm} across various open-source text-to-image models.} We compare the original generation vs. generation by intervening on the layers identified with \crossprompt \hspace{0.05cm} along with a target prompt. We find that across various text-to-image models, visual attributes such as {\it style, objects, facts} can be manipulated by intervening only on a very small fraction of cross-attention layers. }%
    \vspace{-0.4cm}
\end{figure*}
\vspace{-0.25cm}
\section{On the Effectiveness of Causal Tracing for Text-to-Image Models} 
\label{causal_trace_ineffective}
In this section, we empirically observe the effectiveness of causal tracing to models beyond Stable-Diffusion-v1-5. In particular, we find the ability of causal tracing to identify localized control points in Stable-Diffusion-v2-1, OpenJourney, SD-XL and DeepFloyd. 

\textbf{Causal Tracing in UNet.}  In Figure~\ref{causal_trace_unet}, we find that knowledge across different visual attributes is distributed in the UNet for all the text-to-image models (except for DeepFloyd), similar to Stable-Diffusion-v1-5. However, the degree of distribution varies between different text-to-image models. While knowledge about various visual attributes is densely distributed in Stable-Diffusion variants, for SD-XL we find that the distribution is extremely sparse (e.g., only 5$
\%$ of the total layers are causal). For DeepFloyd, we observe that there are no strong causal states in the UNet. We provide more qualitative visualizations on causal tracing across the these text-to-image models in~\Cref{causal_tracing_viz}. Overall, these results reinforce the difficulty of editing knowledge in the UNet directly due to (i) distribution of causal states or (ii) absence of any. 

\textbf{Causal Tracing in Text-Encoder. } ~\citet{basu2023localizing} show that there exists a unique causal state in the text-encoder for Stable-Diffusion-v1-5 and Stable-Diffusion-v2-1 which can be used to perform fast model editing. In Figure~\ref{causal_trace_text}, we find that such an unique causal state is absent in the text-encoder for DeepFloyd and SD-XL.
We note that DeepFloyd uses a T5-text encoder, whereas SD-XL uses a a combination of CLIP-ViT-L and OpenCLIP-ViT-G \cite{radford2021learning}.
Our empirical results indicate that an unique causal state arises only when a CLIP text-encoder is used by itself in a text-to-image model.
\vspace{-0.25cm}
\section{\crossprompt: Towards Mechanistic Knowledge Localization}
Given the lack of generalizability of knowledge localization using causal tracing as shown in~\Cref{causal_trace_ineffective}, we introduce~\crossprompt \hspace{0.05cm}, which can identify localized control regions for visual attributes across {\it all} text-to-image models. 
\vspace{-0.6cm}
\subsection{Knowledge Control in Cross-Attention Layers}
During the inference process, the regulation of image generation involves the utilization of classifier-free guidance, as outlined in \citet{ho2022classifierfree}
which incorporates scores from both the conditional and unconditional diffusion models at each time-step.
Specifically, the classifier-free guidance is applied at each time-step to combine the conditional ($\epsilon_{\theta}(\vb{z}_{t}, \vb{c}, t)$) and
unconditional score estimates ($\epsilon_{\theta}(\vb{z}_{t},t)$).
The result is a combined score denoted as $\hat{\epsilon}(\vb{z}_{t}, \vb{c}, t)$.
\vspace{-0.05cm}
\begin{equation}
    \label{cf_free_guidance}
    \resizebox{0.48\textwidth}{!}{
        \begin{math}
            \hat{\epsilon}(\vb{z}_{t}, \vb{c}, t) = \epsilon_{\theta}(\vb{z_{t}}, \vb{c}, t) + \alpha \left(\epsilon_{\theta}\left(\vb{z_{t}}, \vb{c}, t\right) - \epsilon_{\theta}(\vb{z_{t}}, t)\right), \hspace{0.5em}   \forall t\in \left[T, 1\right]
        \end{math}
    }.
\end{equation}
This combined score is used to update the latent $\vb{z}_{t}$ using DDIM sampling~\citep{DBLP:journals/corr/abs-2010-02502} at each time-step to obtain the final latent code $\vb{z}_{0}$. We term the model $\epsilon_{\theta}(\vb{z}_{t}, \vb{c},t)$ as the \texttt{Clean Model} and the final image generated as $I_{\clean}$.
We note that text is incorporated in the process of generation using cross-attention layers denoted by $\{C_l\}_{l=1}^{M}$ within $\epsilon_{\theta}(\vb{z}_{t}, \vb{c},t) \hspace{0.1cm} \forall t \in [T,1]$.
These layers include key and value matrices -- $\{W_{l}^{K}, W_{l}^{V}\}_{l=1}^{M}$ that take text-embedding $\vb{c}$ of the input prompt and guide the generation toward the text prompt.
Generally, the text-embedding $\vb{c}$ is same across all these layers. However, in order to localize and find control points for different visual attributes,
we replace the original text-embedding $\vb{c}$ with a target prompt embedding $\vb{c}'$ across a small subset of the cross-attention layers and measure its direct effect on the generated image. 
\vspace{-0.25cm}
\subsubsection{Altered Inputs}
We say that a model receives \textit{altered input} when a subset of cross-attention layers $C' \subset \{C_{l}\}_{l=1}^{M}$
receive a different text-embedding $\vb{c}'$ than the other cross-attention layers that take $\vb{c}$ as input.
We name these layers as \textit{controlling layers}.
We denote by $I_{\altered}$ the image generated using this model and~\Cref{cf_free_guidance} with altered inputs when $\vb{z}_{T}$ is given as the initial noise.
We denote the model $\epsilon_{\theta}(\vb{z}_{t}, \vb{c}, \vb{c}', t)$ with the altered inputs as the \texttt{Altered Model} with the following inference procedure:
\begin{align*}
\vspace{-0.0cm}
    \resizebox{0.48\textwidth}{!}{
        \begin{math}
            \hat{\epsilon}(\vb{z}_{t}, \vb{c},\vb{c}', t) = \epsilon_{\theta}(\vb{z_{t}}, \vb{c}, \vb{c}', t) + \alpha (\epsilon_{\theta}(\vb{z_{t}}, \vb{c},\vb{c}', t) - \epsilon_{\theta}(\vb{z_{t}}, t))   
        \end{math}
    }.
\end{align*}
\begin{figure*}
    \hskip -0.25cm
  \includegraphics[width=17.5cm, height=4.8cm]{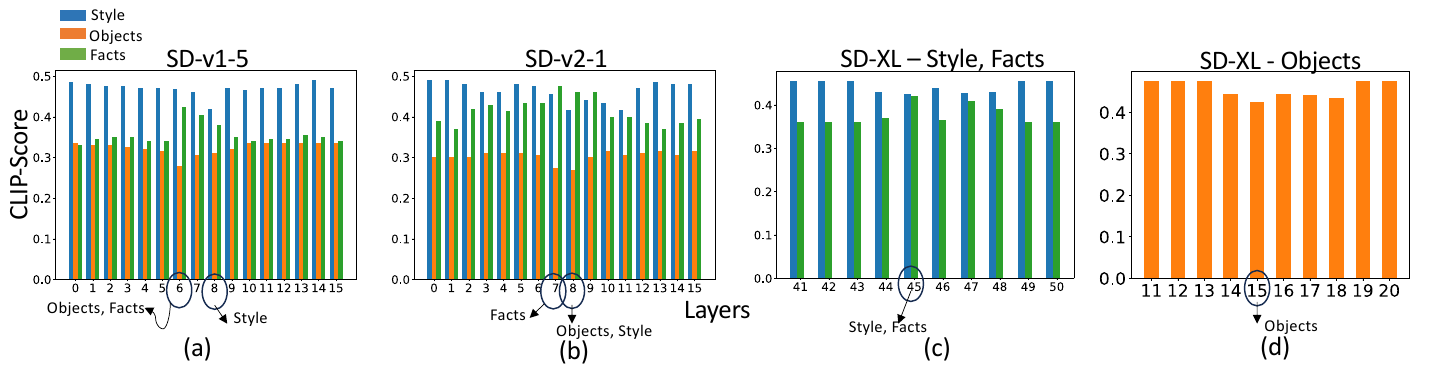}
  \vspace{-0.7cm}
    \caption{\label{clip_score_v1} \textbf{CLIP-Score of the generated images with original prompt for {\it style, objects} and target prompt for {\it facts}~after intervening on layers through~\crossprompt{}.} Lower CLIP-Score for {\it objects, style} indicate correct localization, whereas a higher CLIP-Score indicates such for {\it facts}. (a) For SD-v1-5 (m=2), {\it objects, facts} can be controlled from Layer 6, whereas {\it style} can be controlled from Layer 8. (b) For SD-v2-1(m=3), {\it facts} are controlled from Layer 7, {\it style} and {\it objects} from Layer 8. (c,d): For SD-XL, {\it style (m=3), facts(m=5)} are controlled from Layer 45, whereas {\it objects} are controlled from Layer 15. 
    }%
    \vspace{-0.4cm}
\end{figure*}
As an example, to find the layers where {\it style} knowledge corresponding to a particular artist is stored,
$\{C_{l}\}_{l=1}^{M} - C'$ receive text-embeddings corresponding to the prompt {\it `An $<$object$>$ in the style of $<$artist$>$'}, whereas the layers in $C'$ receive text-embeddings corresponding to the prompt {\it `An $<$object$>$ in the style of painting'}.
If the generated image with these inputs do not have that particular style,
we realize that controlling layers $C'$ are responsible for incorporating that specified style in the output (see Figure~\ref{teaser}).
In fact, this replacement operation enables finding locations across different cross-attention layers where various visual attribute knowledge is localized.
\subsubsection{\crossprompt \hspace{0.04cm} Algorithm}
\label{sec:crossprompt}
Our goal is to find controlling layers $C'$ for different visual attributes. 
We note that the cardinality of the set $|C'| = m$ is a hyper-parameter and the search space for $C'$ is exponential.
Given $|C'| = m$, there are $\binom{M}{m}$ possibilities for $C'$, thus, we restrict our search space to only adjacent cross-attention layers.
In fact, we consider all $C'$ such that  $C' = \{C_l\}_{l=j}^{j+m-1}$ for $j \in [1, M-m+1]$.

\textbf{Selecting the hyper-parameter m}.
To select the cardinality of the set $C'$, 
we run an iterative hyper-parameter search with $m \in [1, M]$, where $M$ is selected based on the maximum number of cross-attention layers in a given text-to-image generative model.
At each iteration of the hyper-parameter search,
we investigate whether there exists a set of $m$ adjacent cross-attention layers that are responsible for the generation of the specific visual attribute.
We find minimum $m$ that such controlling layers for the particular attribute exists.

To apply \crossprompt{} for a particular attribute,
we obtain a set of input prompts $T = \{T_i\}_{i=1}^N$ that include the particular attribute and corresponding set of prompts $T' = \{T'_i\}_{i=1}^N$
where $T'_i$ is analogous to $T_i$ except that the particular attribute is removed/updated.
These prompts serve to create altered images and assess the presence of the specified attribute within them.
Let $\vb{c}_i$ be the text-embedding of $T_i$ and $\vb{c}'_i$ be that of $T'_i$.

Given $m$, we examine all $M-m+1$ possible candidates for controlling layers.
For each of them, we generate $N$ altered images where $i$-th image is generated by giving $\vb{c}'_i$ as the input embedding to selected $m$ layers and $\vb{c}_i$ to other ones.
Then we measure the CLIP-Score \citep{hessel2021clipscore} of original text prompt $T_i$ to the generated image for {\it style, objects} and target text prompt $T_{i}'$ to the generated image for {\it facts}.  For {\it style} and {\it objects}, drop in CLIP-Score shows the removal of the attribute while for {\it facts} increase in score shows similarity to the updated fact. We take the average of the mentioned score across all $1 \leq i \leq N$.
By doing that for all candidates,
we report the one with minimum average CLIP-Score for {\it style, objects} and maximum average CLIP-Score for {\it facts}.
These layers could be candidate layers controlling the generation of the specific attribute.
Algorithm~\ref{alg:cap} provides the pseudocode to find the best candidate.
Figure~\ref{clip_score_v1} shows CLIP-Score across different candidates.

\begin{algorithm}
\caption{\crossprompt \label{alg:cap}}
\begin{algorithmic}
\Require{$m, \{T_i\}_{i=1}^N, \{T'_i\}_{i=1}^N, \{\vb{c}_i\}_{i=1}^N, \{\vb{c}'_i\}_{i=1}^N$}
\Ensure{\text{Candidate controlling set}}
\For{$j\gets 1, \dots , M-m$}
    \State $C' \gets \{C_l\}_{l=j}^{j+m-1}$
    \For{$i\gets 1, \dots , N$}
        \State $s_i \gets \textsc{CLIP-Score}\left(T_i, I_{\altered}\right)$ 
        \State $s'_i \gets \textsc{CLIP-Score}\left(T'_i, I_{\altered}\right)$ 
        
    \EndFor
    \State $a_j \gets \textsc{Average}\left(\{s_i\}_{i=1}^N\right)$ \Comment{for objects, style}
    \State $a_j \gets \textsc{Average}\left(\{s'_i\}_{i=1}^N\right)$ \Comment{for facts}
    
    \EndFor
\State $j^* \gets \argmin_{j} a_j$  \hspace{0.3cm} \Comment{for objects, style}

\State $j^* \gets \arg \max_{j} a_j$  \hspace{0.3cm} \Comment{for facts}
\State \Return $a_{j^*}, \{C_l\}_{l=j^*}^{j^*+m-1}$
\end{algorithmic}
\end{algorithm}
\begin{figure*}
    \hskip -0.4cm
  \includegraphics[width=17.5cm, height=8.8cm]{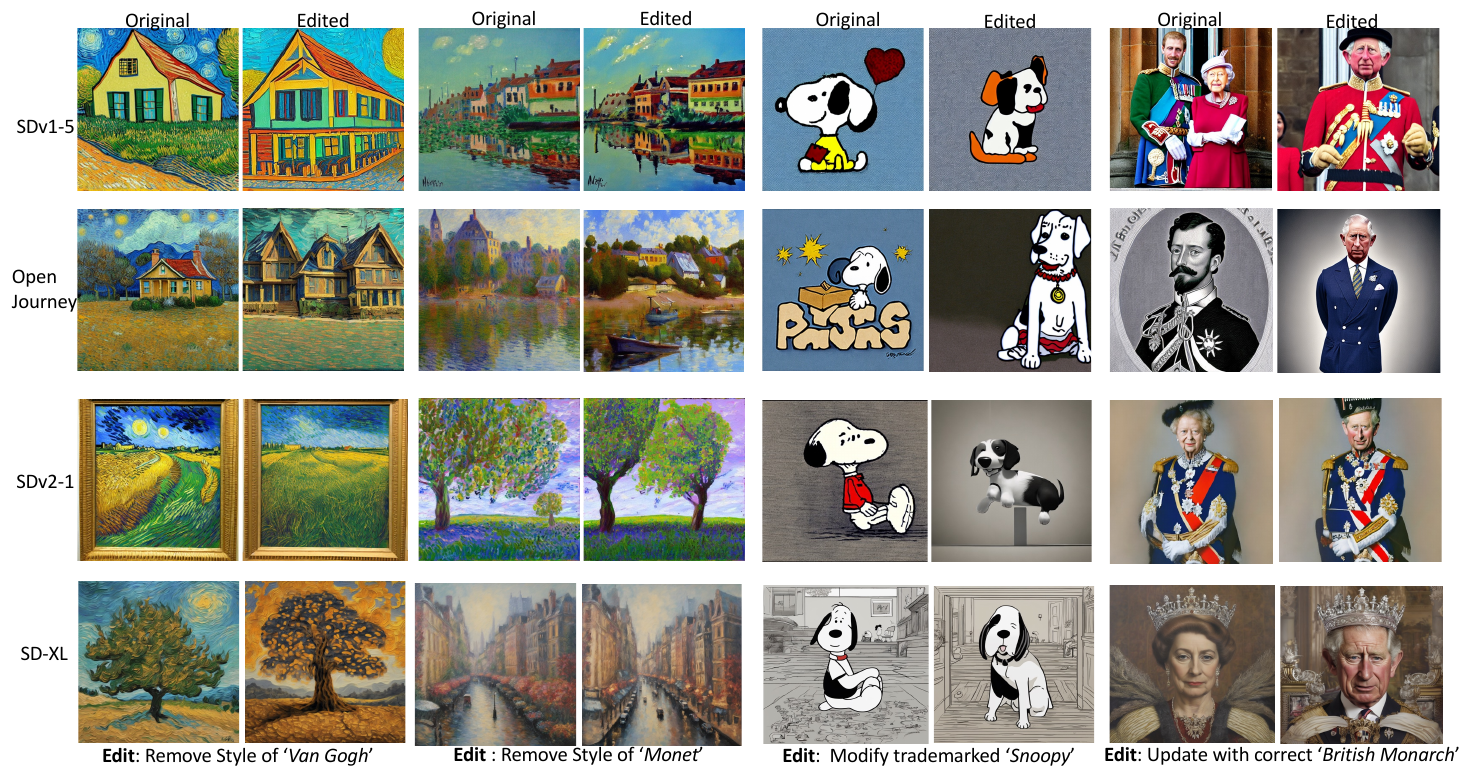}
  \vspace{-0.6cm}
    \caption{\label{viz_prompt_edits} \textbf{\crossedit \hspace{0.05cm} (Model editing) results at locations identified by \crossprompt \hspace{0.05cm} across various open-source text-to-image models.} We observe that locations identified by our interpretability framework can be edited effectively to remove {\it styles}, {\it objects} and update {\it facts} in text-to-image models. We provide more visualizations in~\Cref{viz_prompt_interpret}.}%
    \vspace{-0.5cm}
\end{figure*}

We set a threshold for average CLIP-Score and find the minimum $m$ such that there exists
$m$ adjacent cross-attention layers whose corresponding CLIP-Score meets the requirement.
We point the reader to ~\Cref{hyperparameter} for the values of $m$ selected for different models and thresholds.

\begin{figure}
    \hskip -0.1cm
  \includegraphics[width=\columnwidth]{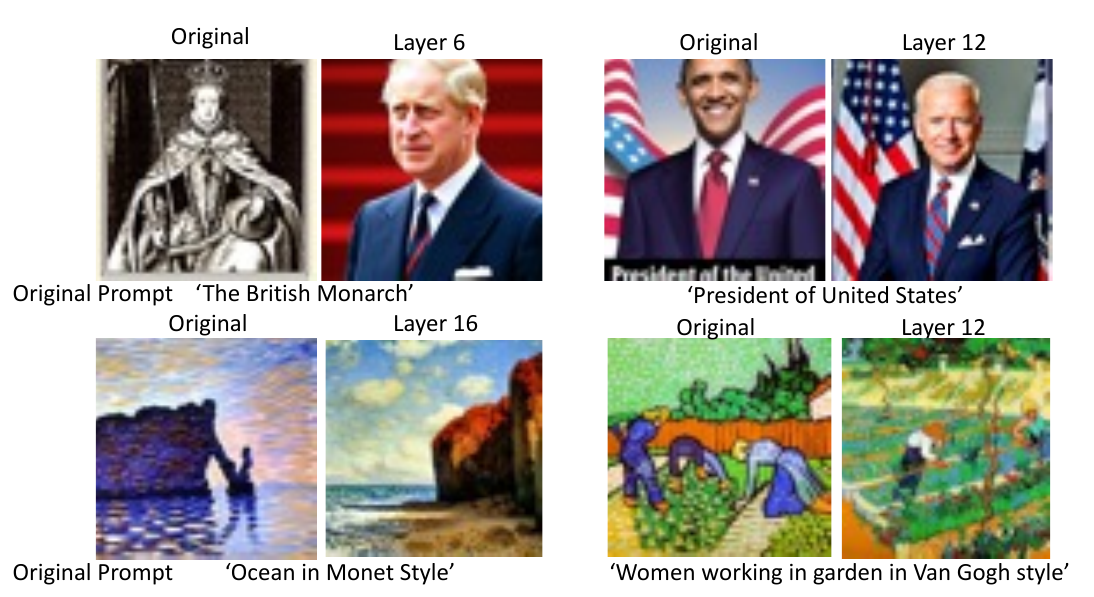}
  \vspace{-0.3cm}
    \caption{\label{deepfloyd_viz} \textbf{Interpretability Results for DeepFloyd.} We find the control points for visual attributes to be dependent on the underlying prompts, rather than the visual attribute. }%
    \vspace{-0.6cm}
\end{figure}

\textbf{Dataset for Prompts}. We use the prompts used in~\citep{basu2023localizing, kumari2023ablating} to extract locations in the UNet which control for various visual attributes such as {\it objects, style} and {\it facts}. More details in~\Cref{prompt_dataset}.
\vspace{-0.2cm}
\subsection{Empirical Results}
In this section, we provide empirical results highlighting the localized layers across
various open-source text-to-image generative models:

\textbf{Stable-Diffusion Variants. } 
Across both models, as depicted qualitatively in Figure~\ref{viz_prompt} and quantitatively in Figure~\ref{clip_score_v1}-(a), we observe the presence of a distinctive subset of layers that govern specific visual attributes. In the case of both SD-v1-5 and SD-v2-1, the control for ``style" is centralized at $l=8$ with $m=2$. In SD-v1-5, the control for ``objects" and ``facts" emanates from the same locations: $l=6$ and $m=2$. However, in SD-v2-1, ``objects" are controlled from $l=8$, while ``facts" are influenced by $l=7$. Despite sharing a similar UNet architecture and undergoing training with comparable scales of pre-training data, these models diverge in the text-encoder utilized. This discrepancy in text-encoder choice may contribute to the variation in how they store knowledge concerning different attributes.

\textbf{Open-Journey. } 
We note that Open-Journey exhibits control locations similar to SD-v1-5 for various visual attributes. As illustrated in Figure~\ref{viz_prompt} and Figure~\ref{clip_score_v1}-(a), ``objects" and ``facts" are governed from $l=6$, while ``style" is controlled from $l=8$. Despite the architectural resemblance between Open-Journey and SD-v1-5, it's important to highlight that Open-Journey undergoes fine-tuning on a subset of images generated from Mid-Journey. This suggests that the control locations for visual attributes are more closely tied to the underlying model architecture than to the specifics of the training or fine-tuning data.

\textbf{SD-XL. } 
Within SD-XL, our investigation reveals that both ``style" and ``facts" can be effectively controlled from $l=45$, with $m=3$ as evidenced in Figure~\ref{viz_prompt} and Figure~\ref{clip_score_v1}-(c). For the attribute ``objects," control is situated at $l=15$, albeit with a slightly larger value of $m=5$. In summary, SD-XL, consisting of a total of $70$ cross-attention layers, underscores a significant finding: \textbf{various attributes} in image generation can be governed by only \textbf{a small subset of layers}.

\textbf{DeepFloyd. } Across SD-v1-5, SD-v2-1, Open-Journey, and SD-XL, our findings indicate that visual attributes like ``style", ``objects" and ``facts," irrespective of the specific prompt used, can be traced back to control points situated within a limited number of layers. However, in the case of DeepFloyd, our observations differ. We find instead, that all attributes display localization dependent on the specific prompt employed. To illustrate, factual knowledge related to ``The British Monarch" is governed from $l=6$ with $m=3$, whereas factual knowledge tied to ``The President of the United States" is controlled from $l=12$ (see Figure~\ref{deepfloyd_viz}). This divergence in localization patterns highlights the nuanced behavior of DeepFloyd in comparison to the other models examined. More results can be referred in~\Cref{df_interpretability}.

\textbf{Human-Study Results.}
We run a human-study to verify that \crossprompt{} can effectively identify controlling layers for different visual attributes.
In our setup, evaluators assess $132$ image pairs, each comprising an image generated by \texttt{Clean Model} and
an image generated by \texttt{Altered Model} whose identified cross-attention layers takes different inputs.
Evaluators determine whether the visual attribute is changed in the image generated by \texttt{Altered Model}(for instance, the artistic Van Gogh style is removed from the original image or not).
Covering $33$ image pairs, generated with different prompts per model,
with five participating evaluators, our experiments reveal a $92.58\%$ verification rate for the impact of \crossprompt{}-identified layers on visual attributes.
See more details in Appendix~\ref{sec:human-study-apx}.
\vspace{-0.3cm}
\section{\crossedit \hspace{0.05cm}: Editing to Ablate Concepts}
\vspace{-0.01cm}
In this section, we analyse the effectiveness of closed-form edit updates in the layers identified by~\crossprompt \hspace{0.05cm} across different text-to-image generative models. 
\newcommand{\source}{\text{orig}}
\newcommand{\target}{\text{target}}
\vspace{-0.0cm}
\subsection{Method}
\Cref{alg:cap} extracts the exact set of cross-attention layers from which the knowledge about a particular visual attribute (e.g., {\it style}) is controlled.
We denote this set as $C_{\text{loc}}$, where $C_{\text{loc}} \subset C$ and $|C_{\text{loc}}| = m$.
This set of extracted cross-attention layers $C_{loc}$, each containing value and key matrices is denoted as $C_{\text{loc}} = \{\hat{W}_{l}^{K}, \hat{W}_{l}^{V} \}_{l=1}^{m}$.
The objective is to modify these weight matrices $\{\hat{W}_{l}^{K}, \hat{W}_{l}^{V} \}_{l=1}^{m}$ such that
they transform the original prompt (e.g., '{\it A house in the style of Van Gogh}') to a target prompt (e.g., '{\it A house in the style of a painting}')
in a way that the visual attribute in the generation is modified.
Similar to Section~\ref{sec:crossprompt},
we use a set of input prompts $T_{\source} = \{T_i^o\}_{i=1}^N$ consisting of prompts featuring the particular visual attribute.
Simultaneously, we create a counterpart set $T_{\target} = \{T_i^t\}_{i=1}^N$ where each $T_i^t$ is identical to $T_i^o$ but lacks the particular attribute in focus.
Let $\vb{c}_i^o \in \mathbb{R}^d$ be the text-embedding of the last subject token in $T_i^o$ and $\vb{c}_i^t \in \mathbb{R}^d$ be that of $T_i^t$.
We obtain matrix $\vb{X}_{\source} \in \mathbb{R}^{N \times d}$ by stacking vectors $\vb{c}_1^o, \vb{c}_2^o, \dots, \vb{c}_N^o$ and
matrix $\vb{X}_{\target} \in \mathbb{R}^{N \times d}$ by stacking $\vb{c}_1^t, \vb{c}_2^t, \dots, \vb{c}_N^t$.
\begin{figure}
    \hskip 0.2cm
  \includegraphics[width=\columnwidth]{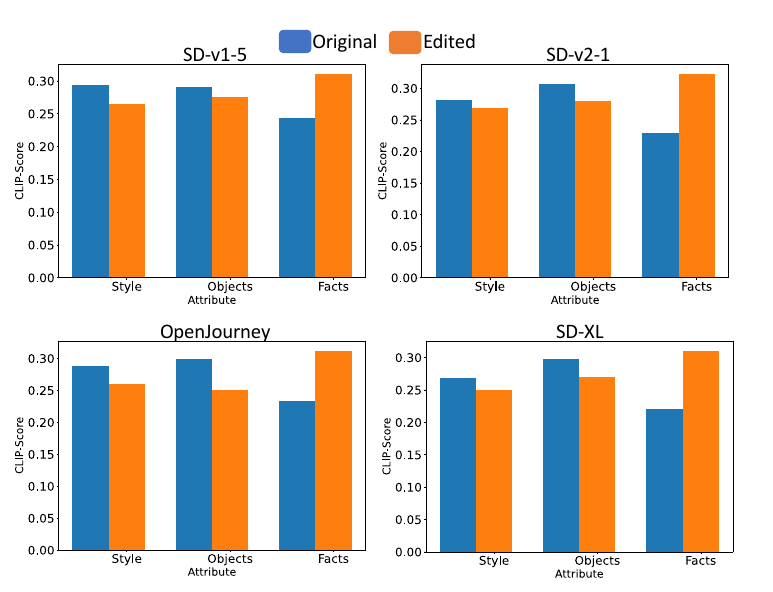}
  \vspace{-0.8cm}
    \caption{\label{model_edit_v1}
    \textbf{Quantitative Model Editing Results for Text-to-Image Models.} We observe a drop in CLIP-Score for ``style" and "objects", while an increase in CLIP-Score for ``facts" therefore highlighting correct edits. 
    }
    \vspace{-0.4cm}
\end{figure}
To learn a mapping between the key and the value embeddings, 
we solve the following optimization for each layer $l \in [1,m]$ corresponding to the key matrices as:
\begin{align*}
    \min_{W_{l}^{K}} \| \vb{X}_{\source}W_{l}^{K} - \vb{X}_{\target}\hat{W}_{l}^{K} \|_{2}^{2} + \lambda_{K} \|W_{l}^{K} - \hat{W}_{l}^{K} \|_{2}^{2}
\end{align*}
where $\lambda_{K}$ is the regularizer.
Letting $\vb{Y}_{\source} = \vb{X}_{\source}W_{l}^{K}$ 
the optimal closed form solution for the key matrix is: 
\begin{align*}
    W_{l}^{K} = (\vb{X}_{\source}^{T}\vb{X}_{\source}+\lambda_{1}I)^{-1}(\vb{X}_{\source}^{T}\vb{Y}_{\target} + \lambda_{K}\hat{W}_{l}^{K})
\end{align*}
Same is applied to get optimal matrix for value embeddings.
\subsection{Model Editing Results}
\textbf{Stable-Diffusion Variants, Open-Journey and SD-XL.}  In Figure~\ref{viz_prompt_edits} and Figure~\ref{model_edit_v1},
it becomes apparent that~\crossedit{} effectively integrates accurate edits into the locations identified by~\crossprompt{}.
Qualitatively examining the visual edits in Figure~\ref{viz_prompt_edits}, our method demonstrates the capability to remove artistic ``styles",
modify trademarked ``objects," and update outdated ``facts" within a text-to-image model with accurate information.
This visual assessment is complemented by the quantitative analysis in Figure~\ref{model_edit_v1},
where we observe that the CLIP-Score of images generated by the edited model, given prompts containing specific visual attributes, consistently registers lower than that of the clean model for ``objects" and ``style." For ``facts," we gauge the CLIP-Score of images from the model with the correct facts,
wherein a higher CLIP-Score indicates a correct edit, as illustrated in Figure~\ref{model_edit_v1}.
Combining both qualitative and quantitative findings, these results collectively underscore the effectiveness of~\crossedit{} across SD-v1-5, SD-v2-1, Open-Journey, and SD-XL.
However, it's noteworthy that the efficacy of closed-form edits varies among different text-to-image models.
Specifically, in the case of ``style," we observe the most substantial drop in CLIP-Score between the edited and unedited models for SD-v1-5 and Open-Journey,
while the drop is comparatively less for SD-v2-1 and SD-XL.
Conversely, for ``facts," we find that all models perform similarly in updating with new information.



\textbf{Limitations with DeepFloyd Closed-Form Edits. } DeepFloyd, despite revealing distinct locations through~\crossprompt \hspace{0.05cm} (albeit depending on the underlying prompt), exhibits challenges in effective closed-form edits at these locations. ~\Cref{deepfloyd_limitations} provides qualitative visualizations illustrating this limitation. The model employs a T5-encoder with bi-directional attention, diverging from other text-to-image models using CLIP-variants with causal attention. Closed-form edits, relying on mapping the last-subject token embedding to a target embedding, are typically effective in text-embeddings generated with causal attention, where the last-subject token holds crucial information. However, the T5-encoder presents a hurdle as tokens beyond the last subject token contribute essential information about the target attribute. Consequently, restricting the mapping to the last-subject token alone proves ineffective for a T5-encoder.

While~\crossprompt{} along with~\crossedit{} makes model editing more interpretable -- we also find that localized-model editing is better than updating all layers in the UNet as shown in~\Cref{update_all}. We also compare our method with existing editing methods~\citep{basu2023localizing, kumari2023ablating, gandikota2023erasing} in~\Cref{compare}. We find that our editing method is at par with existing baselines, with the added advantage of generalizability to models beyond Stable-Diffusion-v1-5. In~\Cref{robustness}, we also show the robustness of our method to generic prompts. 

\section{On Neuron-Level Model Editing}

\begin{figure}
    \hskip 0.2cm
  \includegraphics[width=\columnwidth]{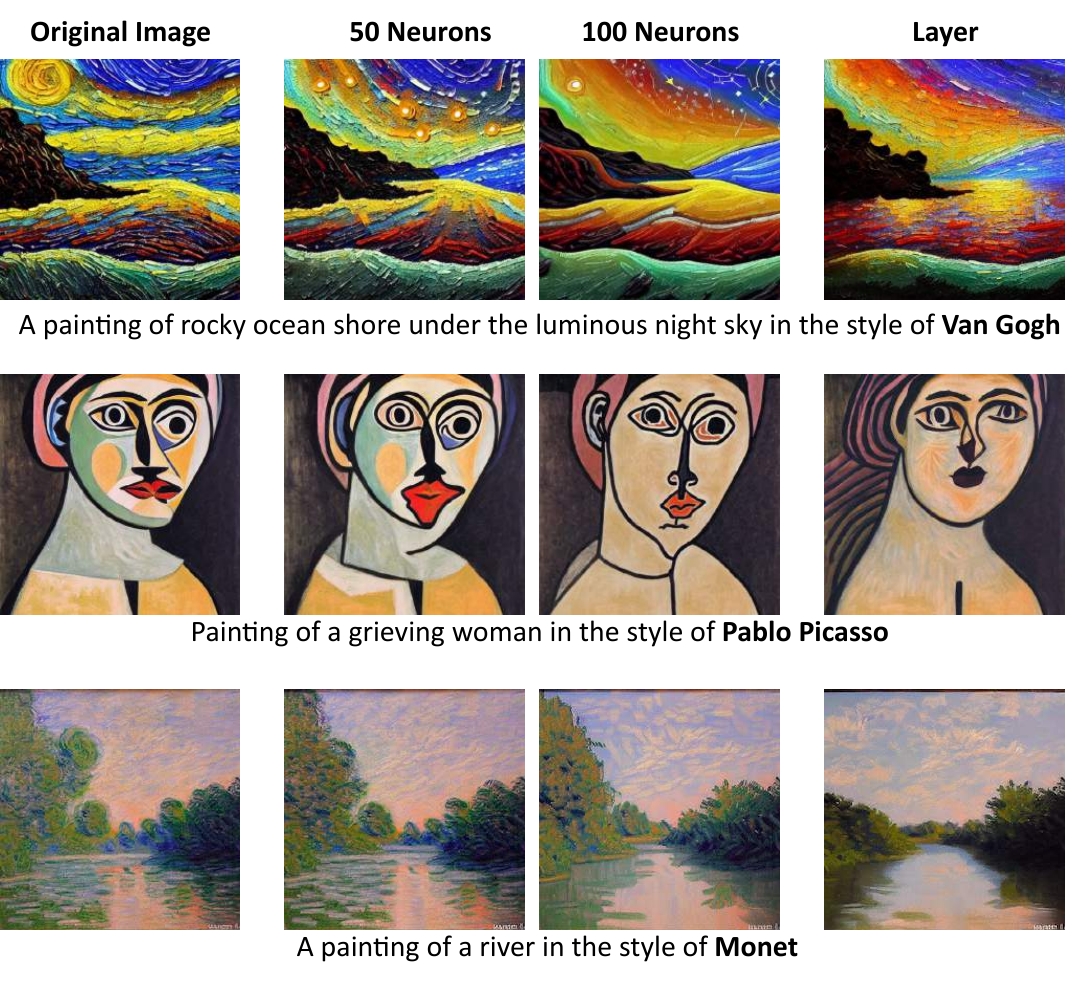}
  \vspace{-0.8cm}
    \caption{\label{zero-shot-edit}
    \textbf{Neuron-Level Model Editing - Qualitative}. Results when applying \textbf{neuron-level dropout} on identified neurons in layers specified with \crossprompt \hspace{0.04cm} on Stable Diffusion v1.5.
    The second and third columns display images with $50$ and $100$ modified neurons out of $1280$ in controlling layers, respectively.
    The last column shows images with a different embedding in controlling layers.
    }%
    \vspace{-0.8cm}
\end{figure}
In this section, we explore the feasibility of effecting neuron-level modifications to eliminate stylistic attributes from the output of text-to-image models.
According to layers identified with \crossprompt, 
our objective is to ascertain whether the selective dropout of neurons at the activation layers
within the specified cross-attention layers (key and value embeddings) can successfully eliminate stylistic elements.

To accomplish this objective, we first need to identify which neurons are responsible for the generation of particular artistic styles, e.g., {\it Van Gogh}.
We examine the activations of neurons in the embedding space of key and value matrices in identified cross-attention layers.
More specifically, we pinpoint neurons that exhibit significant variations when comparing input prompts that include a particular style with the case that input prompts do not involve the specified style.

To execute this process, we collect a set of $N_1$ prompts that feature the specific style, e.g. Van Gogh.
We gather text-embeddings of the last subject token of these prompts denoted by $\vb{c}_1, \vb{c}_2, \text{…}, \vb{c}_{N_1}$, where $\vb{c}_i \in \mathbb{R}^{d}$.
We also obtain a set of $N_2$ prompts without any particular style and analogously obtain $\{\vb{c}'_1, \vb{c}'_2, \text{…}, \vb{c}'_{N_2}\}$, where $\vb{c}'_i \in \mathbb{R}^{d}$.
Next, for the key or value matrix $W \in \mathbb{R}^{d \times d'}$, we consider key or value embedding of these input prompts, i.e.,
$\{z_i\}_{i=1}^{N_1} \cup \{z'_i\}_{i=1}^{N_2}$ where $z_i = \vb{c}_i W$ and $z'_i = \vb{c}'_i W$.
We note that $z_i, z'_i \in \mathbb{R}^{d'}$.

Subsequently, for each of these $d'$ neurons, we assess the statistical difference in their activations between input prompts that include a particular style and those without it.
Specifically, we compute the z-score for each neuron within two groups of activations: ${z_1, z_2, \text{…}, z_{N_1}}$ and ${z'_1, z'_2, \text{…}, z'_{N_2}}$.
The neurons are then ranked based on the absolute value of their z-score,
with the top neurons representing those that exhibit significant differences in activations depending on the presence or absence of a particular concept in the input prompt.
During generation, we drop-out these neurons and see if particular style is removed or not.
\begin{figure}
    \hskip 0.2cm
  \includegraphics[width=\columnwidth]{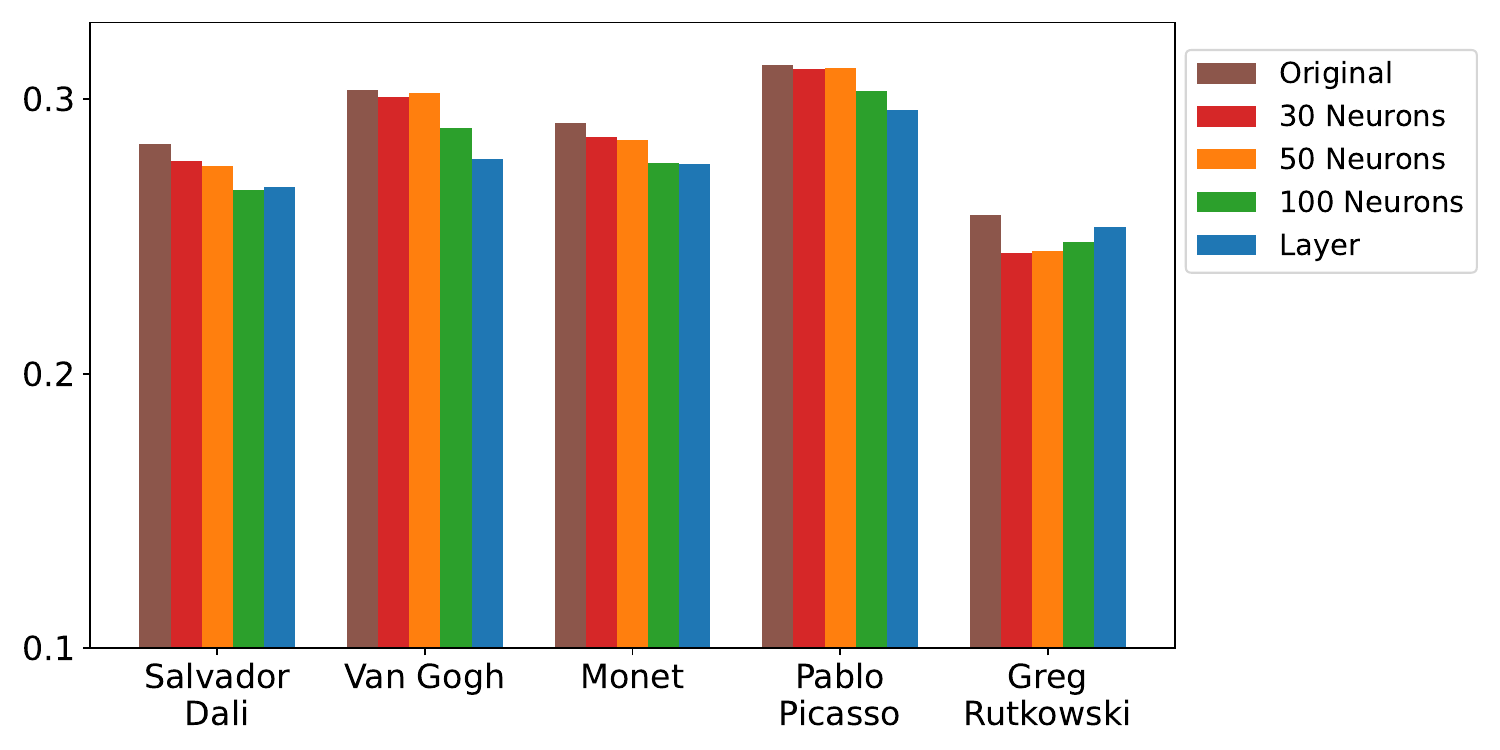}
  \vspace{-0.8cm}
    \caption{\label{zero-shot-edit-score}
    \textbf{Neuron-Level Model Editing - Quantitative}. Average CLIP-Score of generated images to text prompt {\it 'style of $<$artist$>$'}.
    Brown bars show similarity to original generated image;
    red, orange, and green bars show similarity to generated image when $30$, $50$, and $100$ neurons are modified, respectively;
    and blue bars refer to images when controlling layers receive other prompt.
    }
    \vspace{-0.7cm}
\end{figure}
As seen in Figure~\ref{zero-shot-edit}, neuron-level modification at inference time is effective at removing styles.
This shows that knowledge about a particular style can be even more localized to a few neurons.
It is noteworthy that the extent of style removal increases with the modification of more neurons, albeit with a trade-off in the quality of generated images.
This arises because modified neurons may encapsulate information related to other visual attributes.
To quantify the effectiveness of this approach,
we measure the drop in CLIP-Score for modified images across various styles.
Figure~\ref{zero-shot-edit-score} presents a bar-plot illustrating these similarity scores.
Notably, drop in CLIP-Score demonstrates that neuron-level model editing effectively removes the styles associated with different artists in the generated images.
We refer to Appendix~\ref{sec:zero-shot-apx1} for more details on neuron-level model editing experiments.

\vspace{-0.2cm}
\section{Conclusion}
In our paper, we comprehensively examine knowledge localization across various open-source text-to-image models. We initially observe that while causal tracing proves effective for early Stable-Diffusion variants, its generalizability diminishes when applied to newer text-to-image models like DeepFloyd and SD-XL for localizing control points associated with visual attributes. To address this limitation, we introduce ~\crossprompt{}, capable of effectively identifying locations within the UNet across diverse text-to-image models. Harnessing these identified locations within the UNet, we evaluate the efficacy of closed-form model editing across a range of text-to-image models leveraging~\crossedit \hspace{0.05cm}, uncovering intriguing properties. Notably, for specific visual attributes such as ``style", we discover that knowledge can even be traced to a small subset of neurons and subsequently edited by applying a simple dropout layer, thereby underscoring the possibilities of neuron-level model editing.

\section{Impact Statement}
This paper presents work to advance the understanding of the inner workings of open-source text-to-image generative models. Our interpretability method can advance the understanding of how knowledge is represented in generative models and does not have any potential negative implications on the society. Our editing method can address societal concerns (e.g., an artist asking the model owner to delete their style) in an effective way and to the best of our knowledge does not have any negative societal consequences. 



\bibliography{example_paper}
\bibliographystyle{icml2024}

\newpage
\appendix
\onecolumn
\section{Visualizations with Causal Tracing}
\label{causal_tracing_viz}
\subsection{SD-v2-1}
\begin{figure}[H]
    \hskip 0.2cm
  \includegraphics[width=\columnwidth]{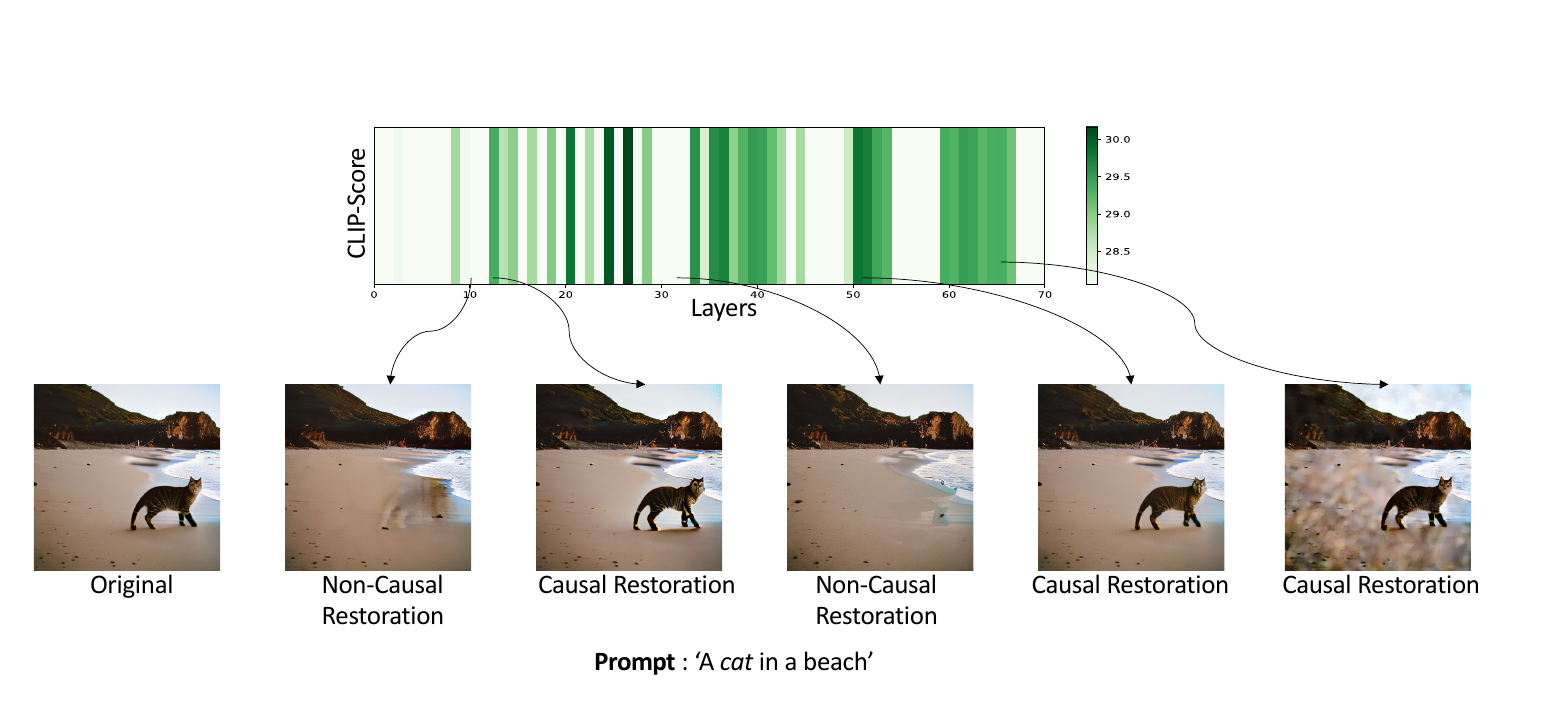}
  \vspace{-0.8cm}
    \caption{\label{sd2_causal-1}
    \textbf{Causal Tracing in the UNet for SD-v2-1} Similar to earlier works~\citep{basu2023localizing}, we find that knowledge about the attribute "objects" is distributed in the UNet amongst various layers. Given that these layers can independently restore a corrupted model to a clean model, model editing requires editing all these causal layers. Moreover these layers cannot be updated in closed-form updates due to the presence of non-linearities in the layer components.
    }
\end{figure}
\subsection{SD-XL}
\begin{figure}[H]
    \hskip 0.2cm
  \includegraphics[width=\columnwidth]{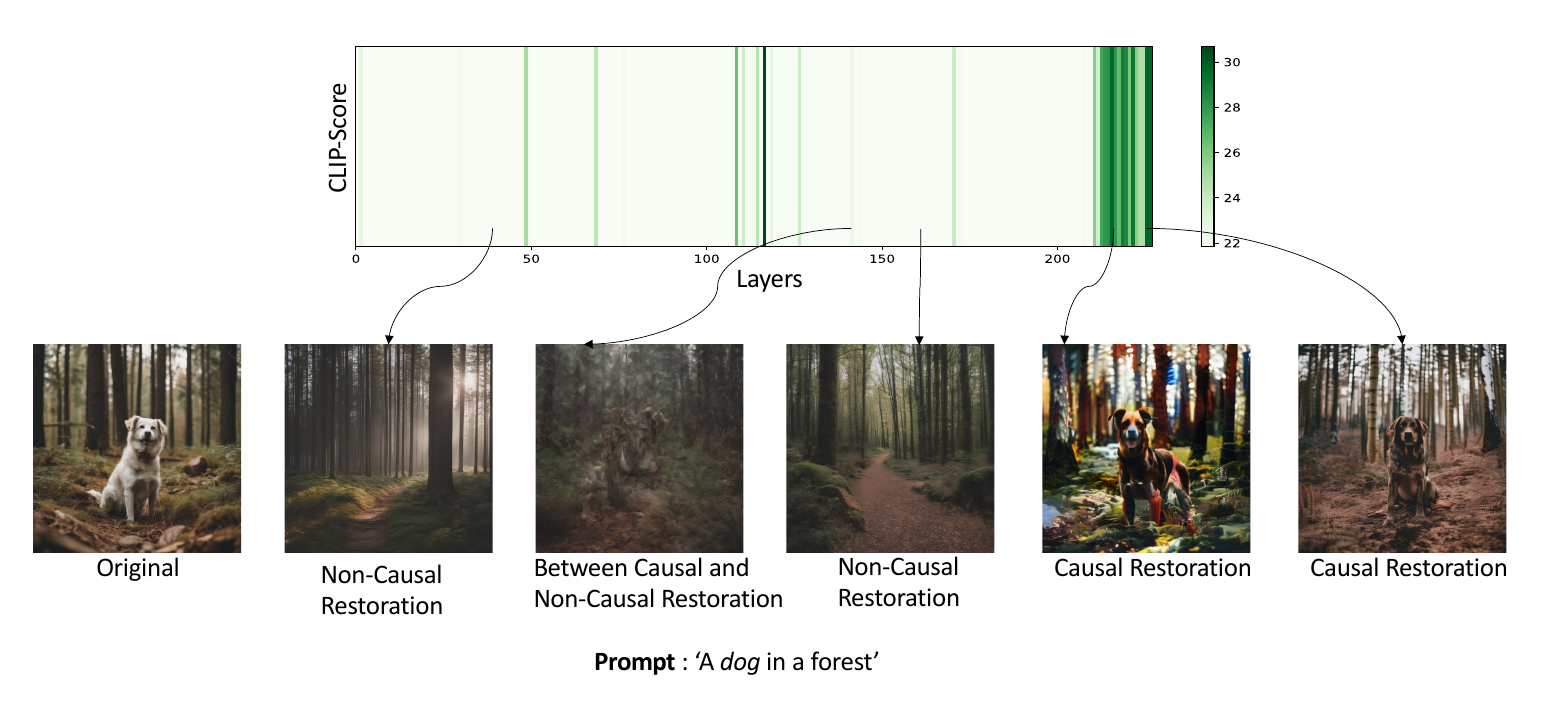}
  \vspace{-0.8cm}
    \caption{\label{sdxl_causal}
    \textbf{Causal Tracing in the UNet for SD-XL}. We find that knowledge about the attribute "objects" is distributed amongst the various layers in the UNet. However, compared to other models such as SD-v2-1, the distribution is slightly sparse. 
    }
\end{figure}
\subsection{DeepFloyd}
\begin{figure}[H]
    \hskip 0.2cm
  \includegraphics[width=\columnwidth]{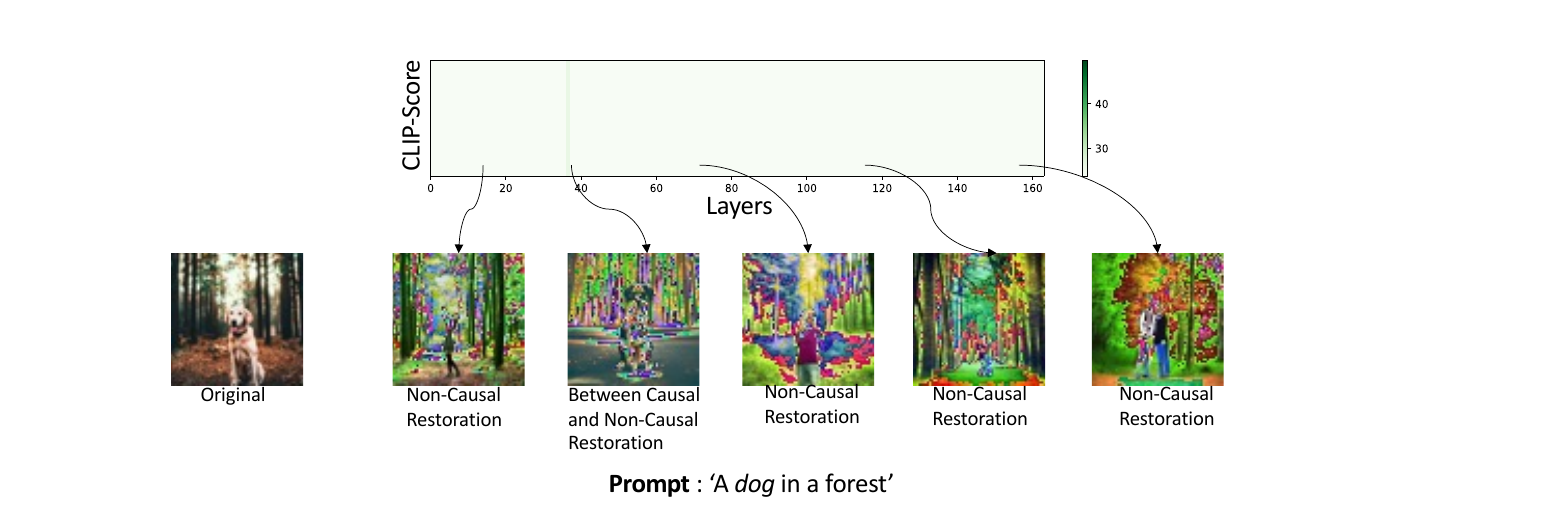}
  \vspace{-0.8cm}
    \caption{\label{sd2_causal-2}
    \textbf{DeepFloyd}. In the case of the DeepFloyd model, we find that there is no presence of strong causal state which can restore a corrupted model to its clean state. This absence of causal states rules out the possibilities of model editing in the UNet. 
    }
\end{figure}
\section{Visualizations with CrossPrompt}
\label{viz_prompt_interpret}
\subsection{SD-v1-5}
\begin{figure}[H]
    \hskip 0.2cm
  \includegraphics[width=\columnwidth]{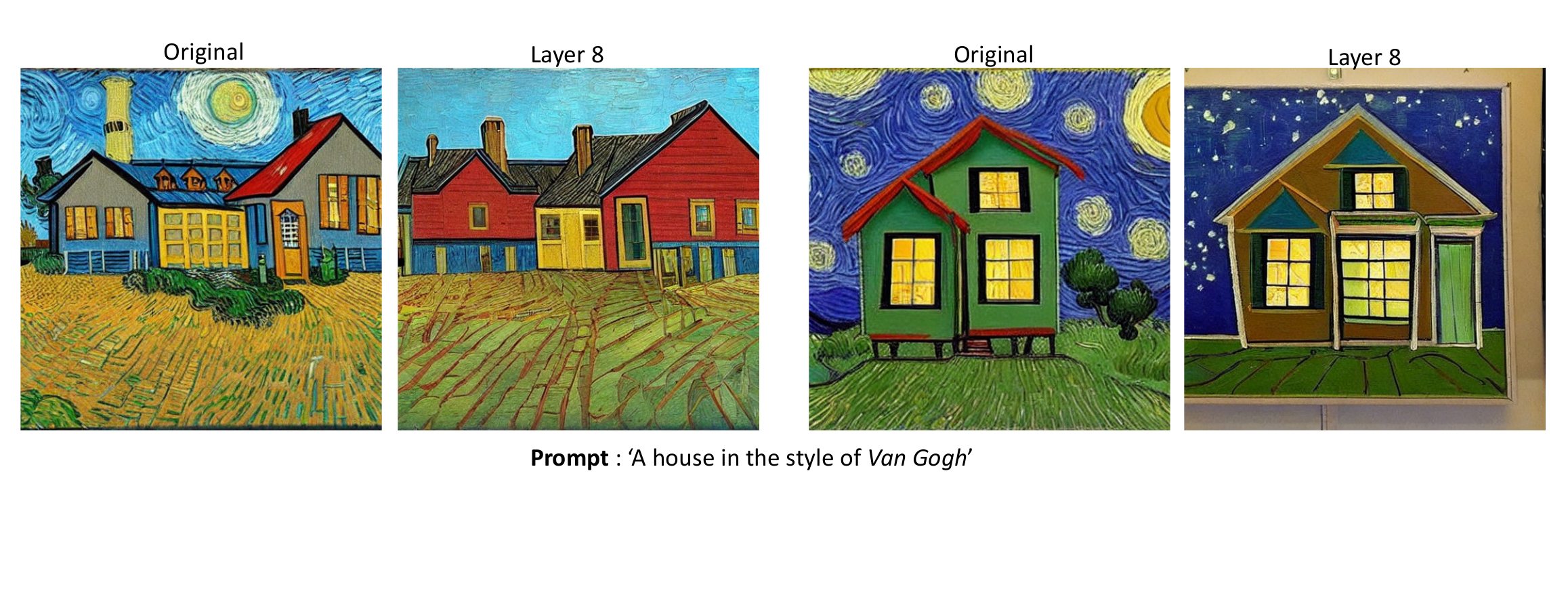}
  \vspace{-0.8cm}
    \caption{\label{sd2_causal-3}
    \textbf{Layer 8 can control ``style" in SD-v1-5}. We perform an intervention in the different cross-attention layers of Stable-Diffusion-v2-1 by using a target prompt - {\it 'a painting'} in those layers while the original prompt is used for other layers. We find that there exists an unique layer which can control output style.  
    }
\end{figure}
\begin{figure}[H]
    \hskip 0.2cm
  \includegraphics[width=\columnwidth]{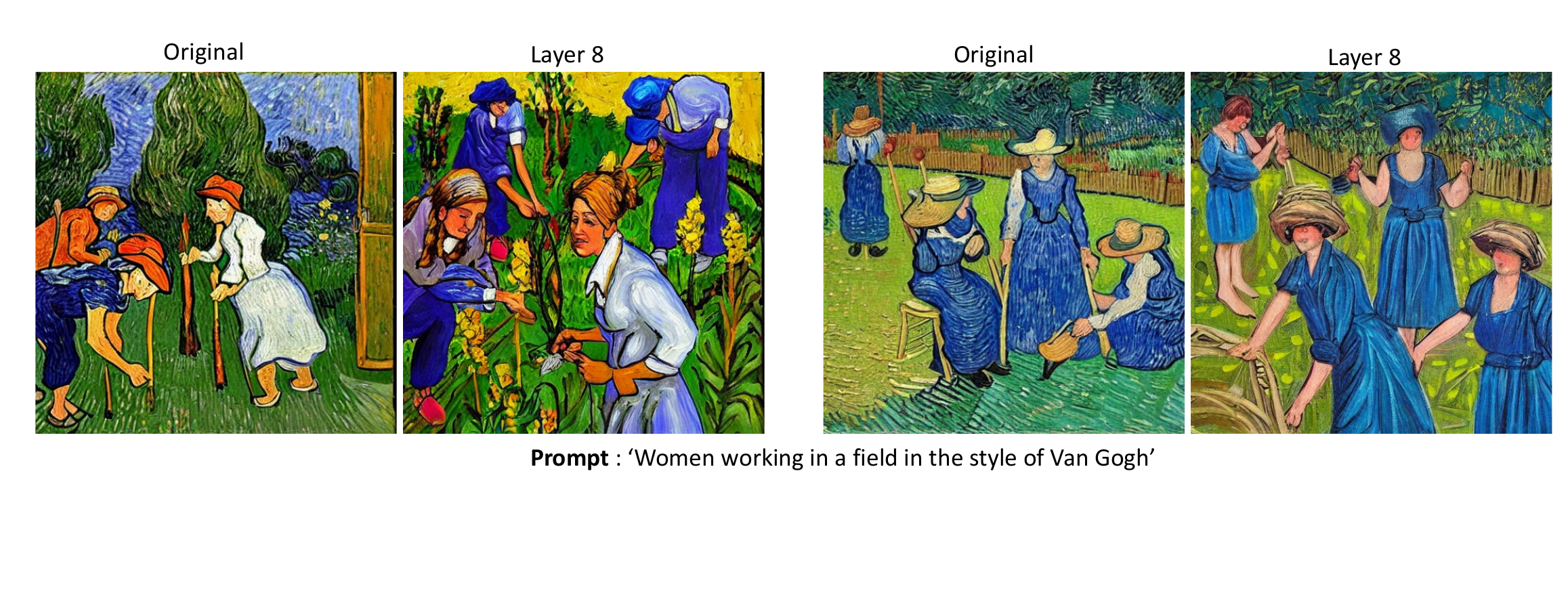}
  \vspace{-0.8cm}
    \caption{\label{sd2_causal-4}
     \textbf{Layer 8 can control ``style" in SD-v1-5}. We perform an intervention in the different cross-attention layers of Stable-Diffusion-v2-1 by using a target prompt - {\it 'a painting'} in those layers while the original prompt is used for other layers. We find that there exists an unique layer which can control output style.  
    }
\end{figure}
\begin{figure}[H]
    \hskip 0.2cm
  \includegraphics[width=\columnwidth]{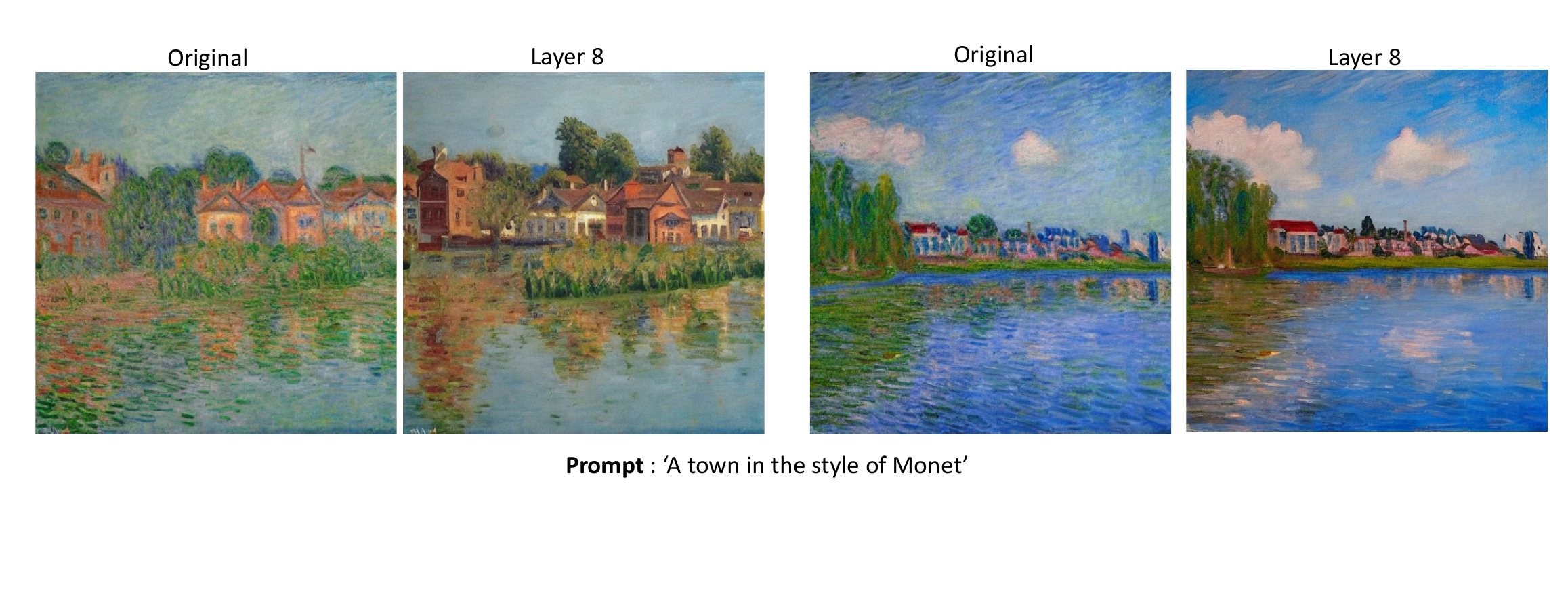}
  \vspace{-0.8cm}
    \caption{\label{sd2_causal-5}
     \textbf{Layer 8 can control ``style" in SD-v1-5}. We perform an intervention in the different cross-attention layers of Stable-Diffusion-v2-1 by using a target prompt - {\it 'a painting'} in those layers while the original prompt is used for other layers. We find that there exists an unique layer which can control output style.  
    }
\end{figure}

\begin{figure}[H]
    \hskip 0.2cm
  \includegraphics[width=\columnwidth]{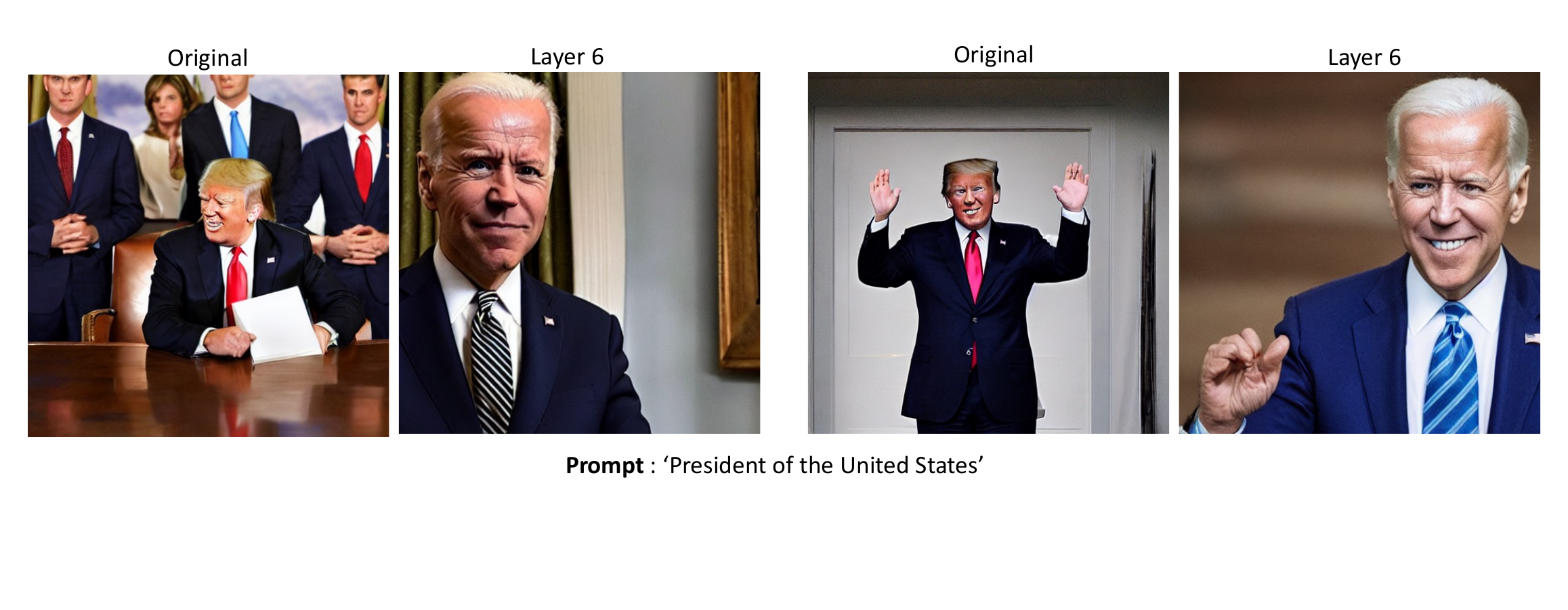}
  \vspace{-0.8cm}
    \caption{\label{sd2_causal-6}
  \textbf{Layer 6 can control ``factual knowledge" in SD-v1-5}. We perform an intervention in the different cross-attention layers of Stable-Diffusion-v2-1 by using a target prompt - {\it 'Joe Biden'} in those layers while the original prompt is used for other layers. We find that there exists an unique layer which can control output generations of ``factual knowledge". 
    }
\end{figure}

\begin{figure}[H]
    \hskip 0.2cm
  \includegraphics[width=\columnwidth]{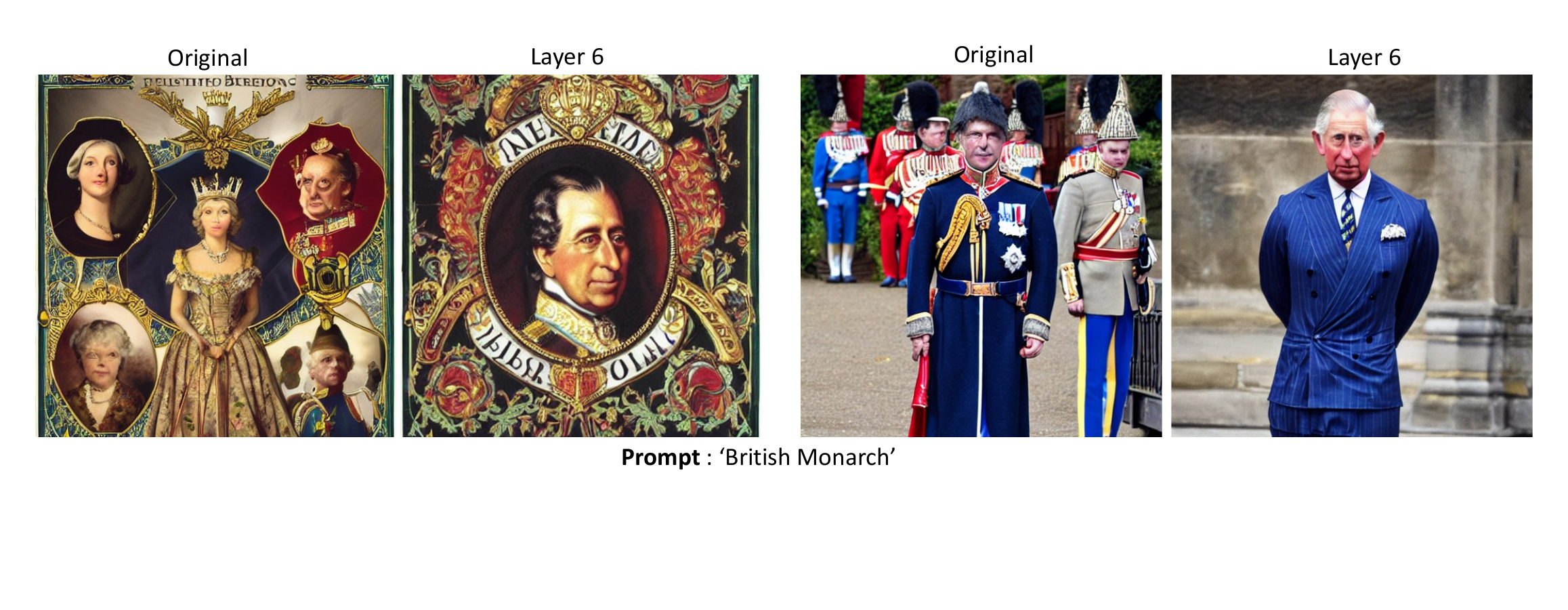}
  \vspace{-0.8cm}
    \caption{\label{sd2_causal-7}
     \textbf{Layer 6 can control ``factual knowledge" in SD-v1-5}. We perform an intervention in the different cross-attention layers of Stable-Diffusion-v2-1 by using a target prompt - {\it 'Prince Charles'} in those layers while the original prompt is used for other layers. We find that there exists an unique layer which can control output generations of ``factual knowledge".  
    }
\end{figure}

\begin{figure}[H]
    \hskip 0.2cm
  \includegraphics[width=\columnwidth]{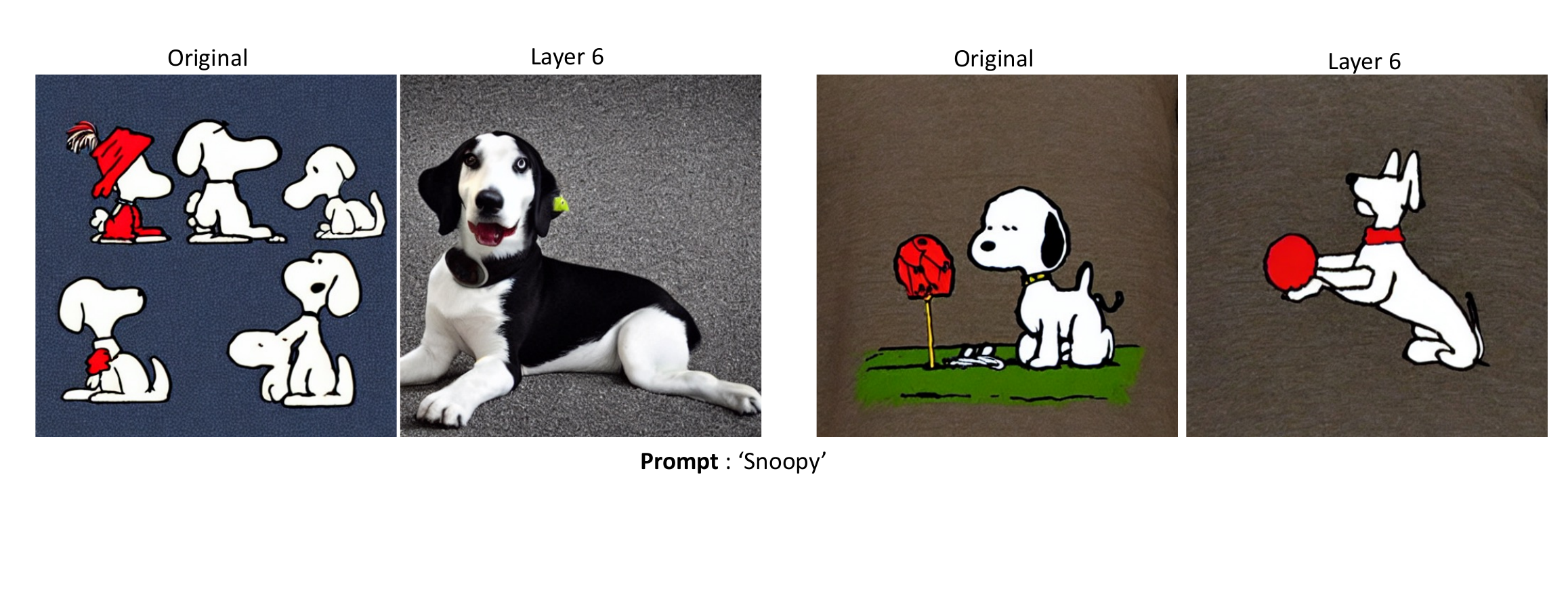}
  \vspace{-0.8cm}
    \caption{\label{sd2_causal-8}
         \textbf{Layer 6 can control "object knowledge" in SD-v1-5}. We perform an intervention in the different cross-attention layers of Stable-Diffusion-v2-1 by using a target prompt - {\it 'a dog'} in those layers while the original prompt is used for other layers. We find that there exists an unique layer which can control output generations of "object knowledge".  
    }
\end{figure}

\subsection{SD-v2-1}
\begin{figure}[H]
    \hskip 0.2cm
  \includegraphics[width=\columnwidth]{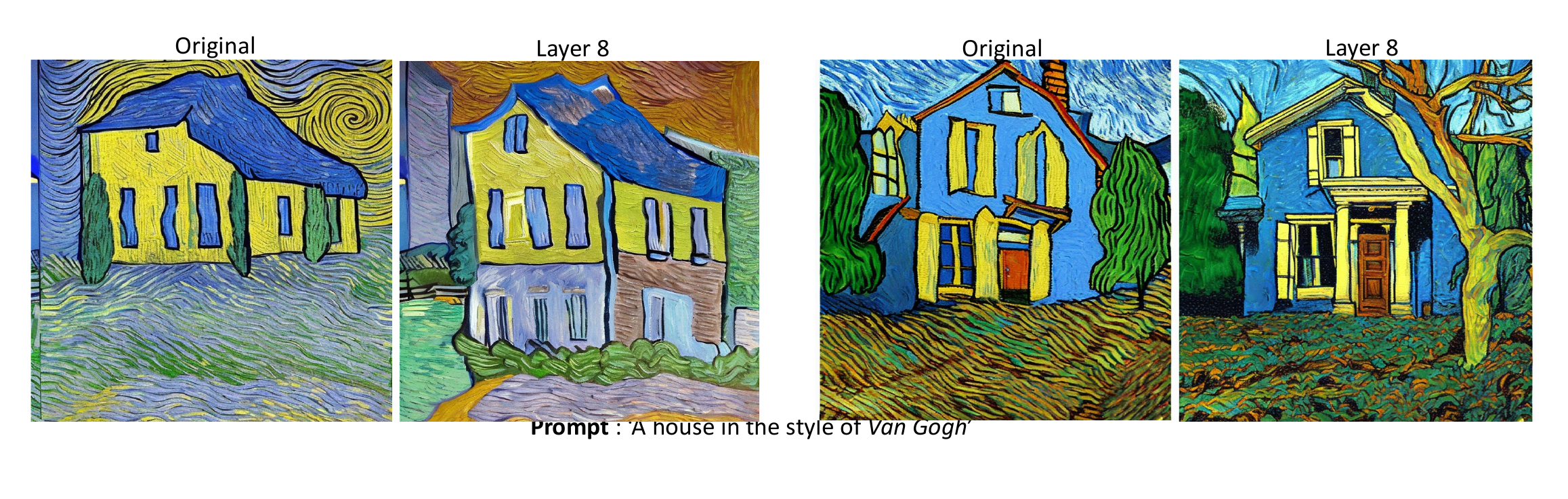}
  \vspace{-0.8cm}
    \caption{\label{sd2_causal-9}
         \textbf{Layer 8 can control ``style" in SD-v2-1}. We perform an intervention in the different cross-attention layers of Stable-Diffusion-v2-1 by using a target prompt - {\it 'a painting'} in those layers while the original prompt is used for other layers. We find that there exists an unique layer which can control output style. 
    }
\end{figure}
\begin{figure}[H]
    \hskip 0.2cm
  \includegraphics[width=\columnwidth]{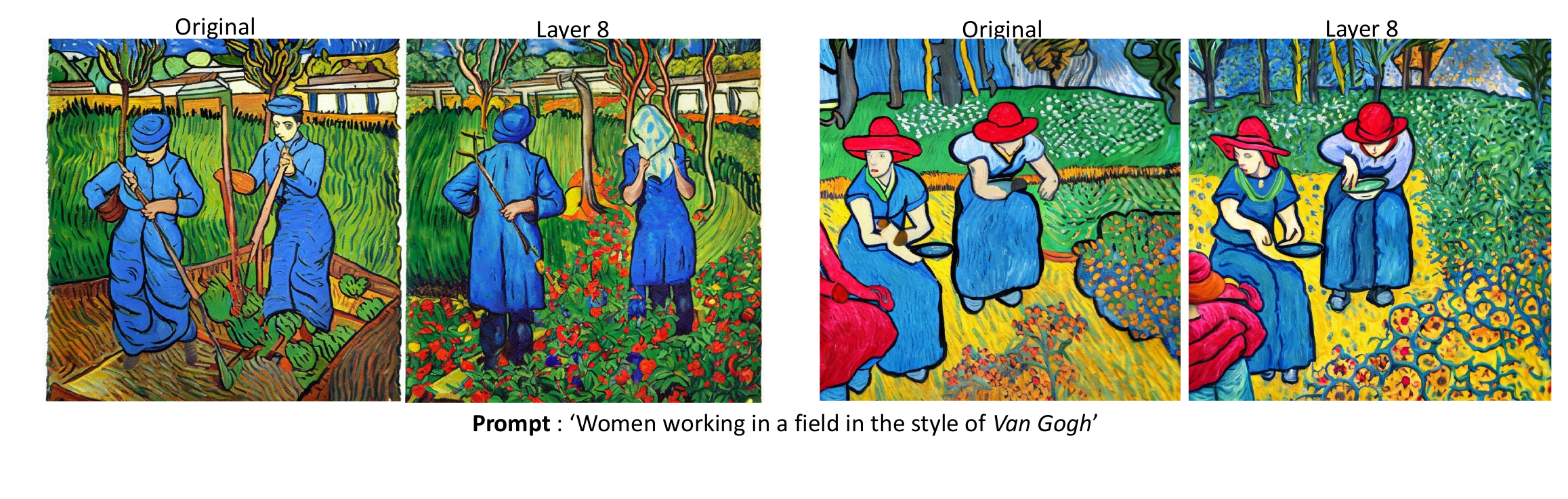}
  \vspace{-0.8cm}
    \caption{\label{sd2_causal-10}
     \textbf{Layer 8 can control ``style" in SD-v2-1}. We perform an intervention in the different cross-attention layers of Stable-Diffusion-v2-1 by using a target prompt - {\it 'a painting'} in those layers while the original prompt is used for other layers. We find that there exists an unique layer which can control output style. 
    }
\end{figure}
\begin{figure}[H]
    \hskip 0.2cm
  \includegraphics[width=\columnwidth]{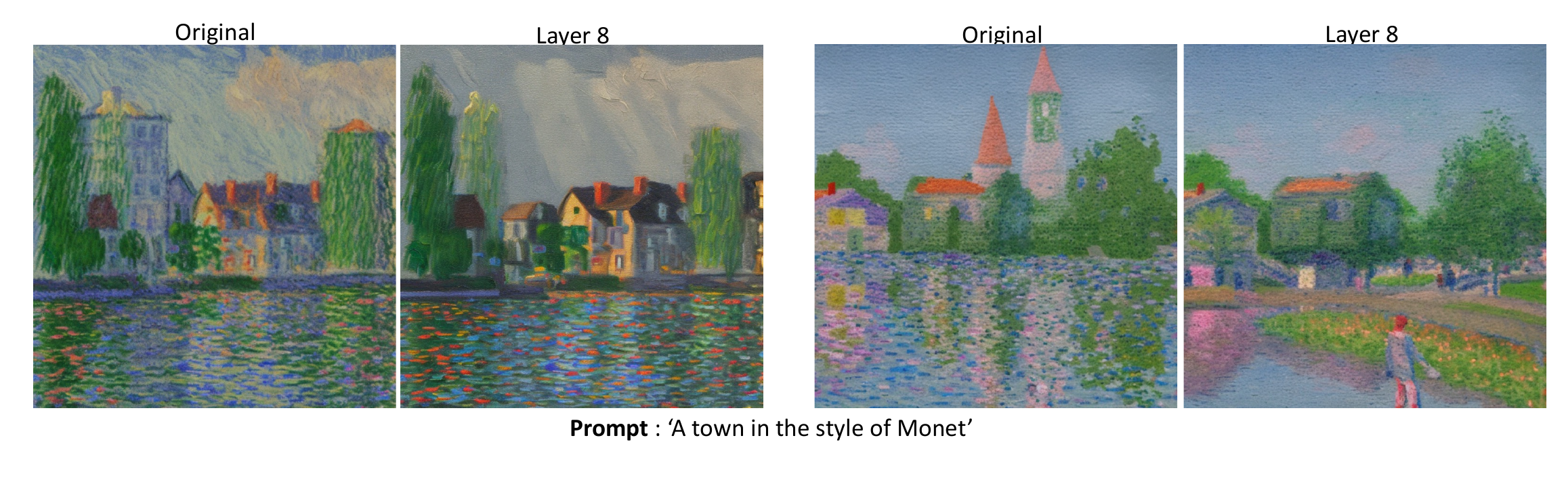}
  \vspace{-0.8cm}
    \caption{\label{sd2_causal-11}
      \textbf{Layer 8 can control ``style" in SD-v2-1}. We perform an intervention in the different cross-attention layers of Stable-Diffusion-v2-1 by using a target prompt - {\it 'a painting'} in those layers while the original prompt is used for other layers. We find that there exists an unique layer which can control output style. 
    }
\end{figure}
\begin{figure}[H]
    \hskip 0.2cm
  \includegraphics[width=\columnwidth]{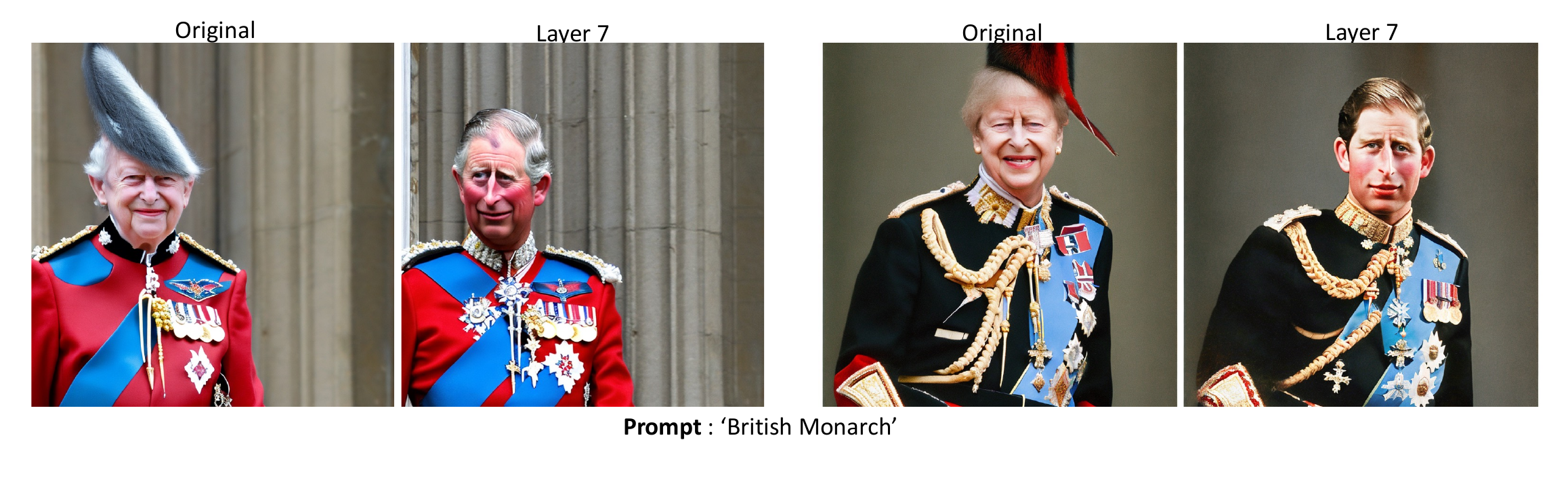}
  \vspace{-0.8cm}
    \caption{\label{sd2_causal-12}
        \textbf{Layer 7 can control ``factual knowledge" in SD-v2-1}. We perform an intervention in the different cross-attention layers of Stable-Diffusion-v2-1 by using a target prompt - {\it 'Joe Biden'} in those layers while the original prompt is used for other layers. We find that there exists an unique layer which can control output generations of ``factual knowledge".  
    }
\end{figure}
\begin{figure}[H]
    \hskip 0.2cm
  \includegraphics[width=\columnwidth]{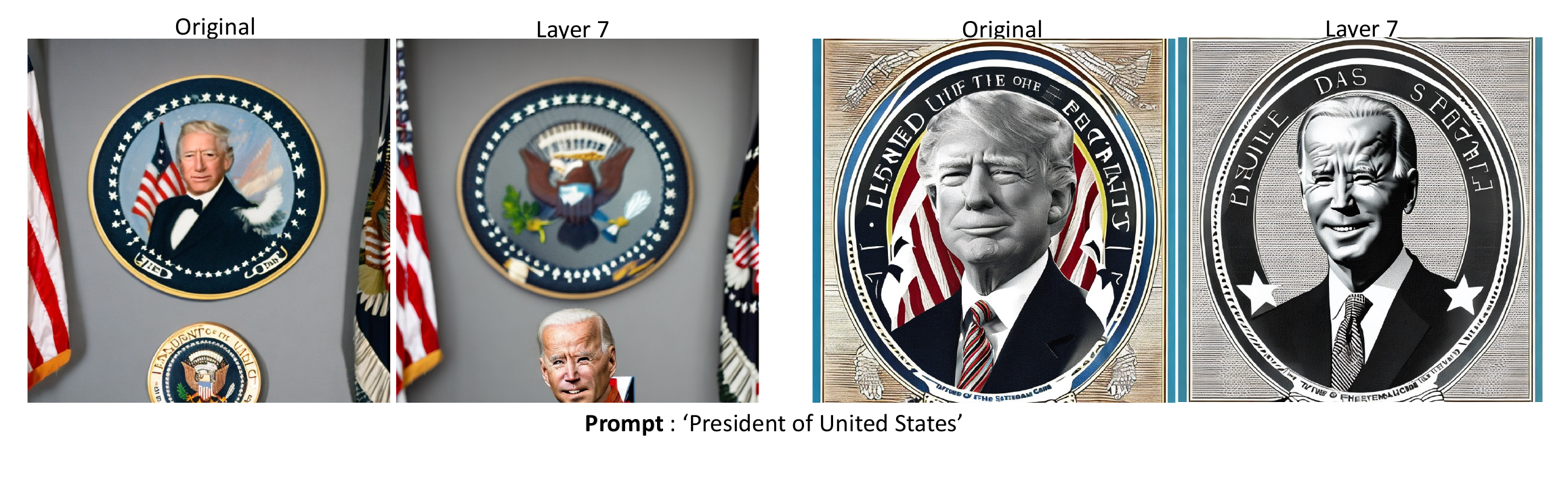}
  \vspace{-0.8cm}
    \caption{\label{sd2_causal-13}
     \textbf{Layer 7 can control ``factual knowledge" in SD-v2-1}. We perform an intervention in the different cross-attention layers of Stable-Diffusion-v2-1 by using a target prompt - {\it 'Prince Charles'} in those layers while the original prompt is used for other layers. We find that there exists an unique layer which can control output generations of ``factual knowledge".  
    }
\end{figure}
\begin{figure}[H]
    \hskip 0.2cm
  \includegraphics[width=\columnwidth]{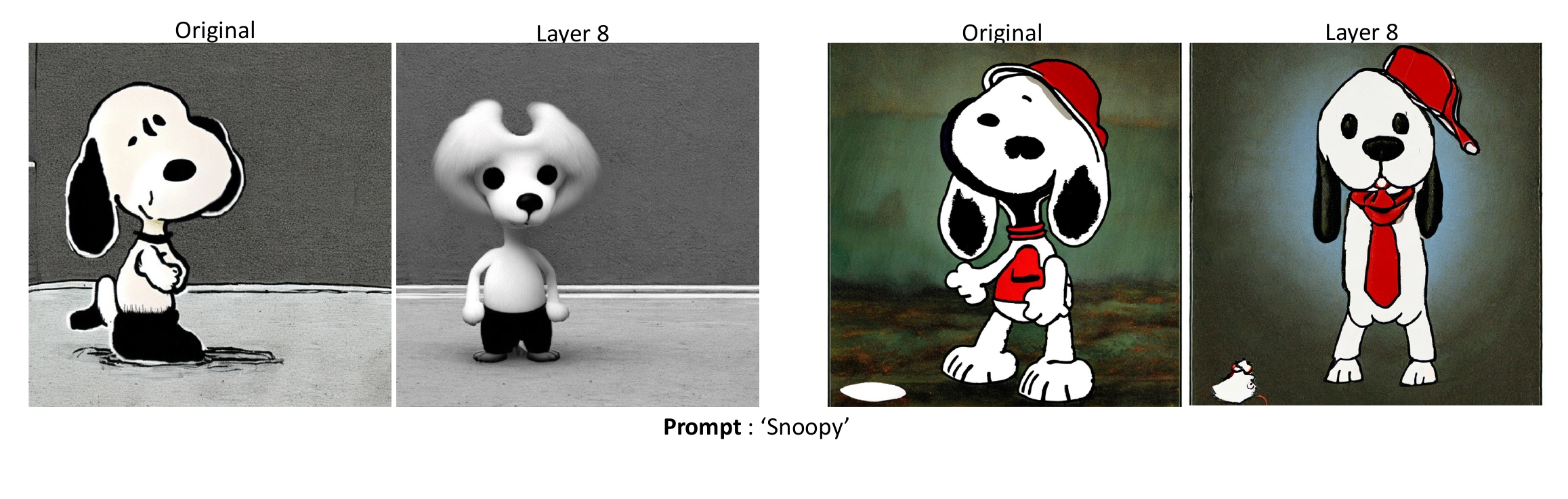}
  \vspace{-0.8cm}
    \caption{\label{sd2_causal-14}
      \textbf{Layer 8 can control "object knowledge" in SD-v2-1}. We perform an intervention in the different cross-attention layers of Stable-Diffusion-v2-1 by using a target prompt - {\it 'a dog'} in those layers while the original prompt is used for other layers. We find that there exists an unique layer which can control output generations of "object knowledge".  
    }
\end{figure}
\subsection{OpenJourney}
\begin{figure}[H]
    \hskip 0.2cm
  \includegraphics[width=\columnwidth]{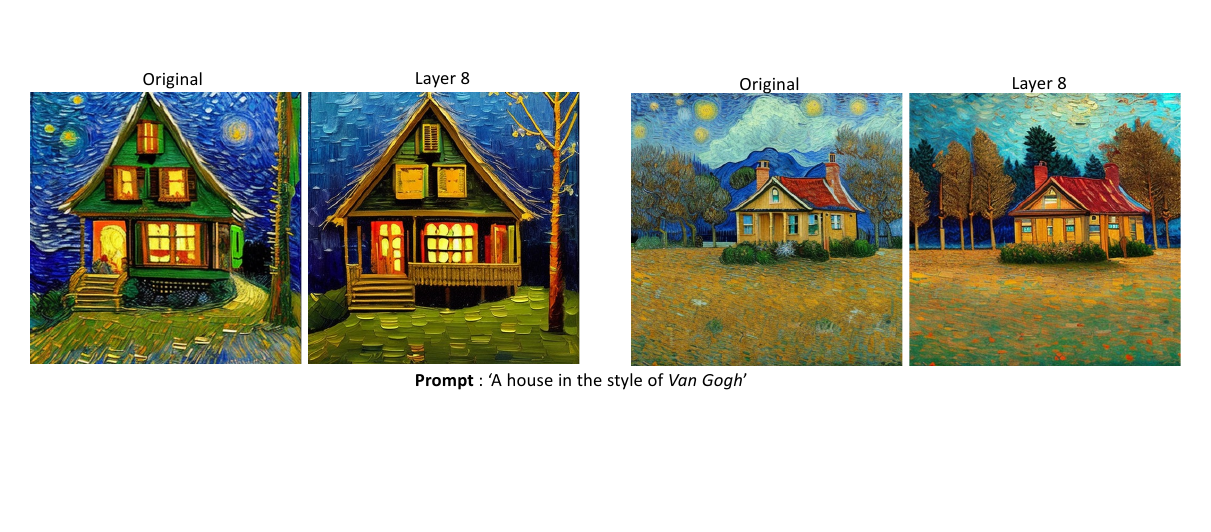}
  \vspace{-0.8cm}
    \caption{\label{sd2_causal-15}
      \textbf{Layer 8 can control ``style knowledge" in Open-Journey}. We perform an intervention in the different cross-attention layers of Open-Journey by using a target prompt - {\it 'a painting'} in those layers while the original prompt is used for other layers. We find that there exists an unique layer which can control output generations of ``style knowledge".  
    }
\end{figure}
\begin{figure}[H]
    \hskip 0.2cm
  \includegraphics[width=\columnwidth]{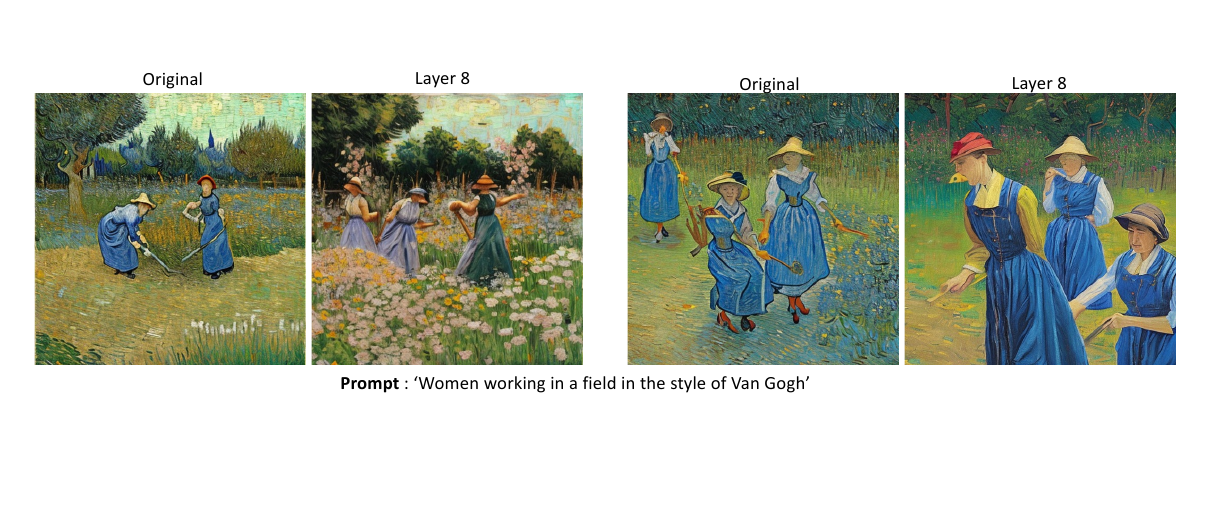}
  \vspace{-0.8cm}
    \caption{\label{sd2_causal-16}
   \textbf{Layer 8 can control ``style knowledge" in Open-Journey}. We perform an intervention in the different cross-attention layers of Open-Journey by using a target prompt - {\it 'a painting'} in those layers while the original prompt is used for other layers. We find that there exists an unique layer which can control output generations of ``style knowledge".  
    }
\end{figure}
\begin{figure}[H]
    \hskip 0.2cm
  \includegraphics[width=\columnwidth]{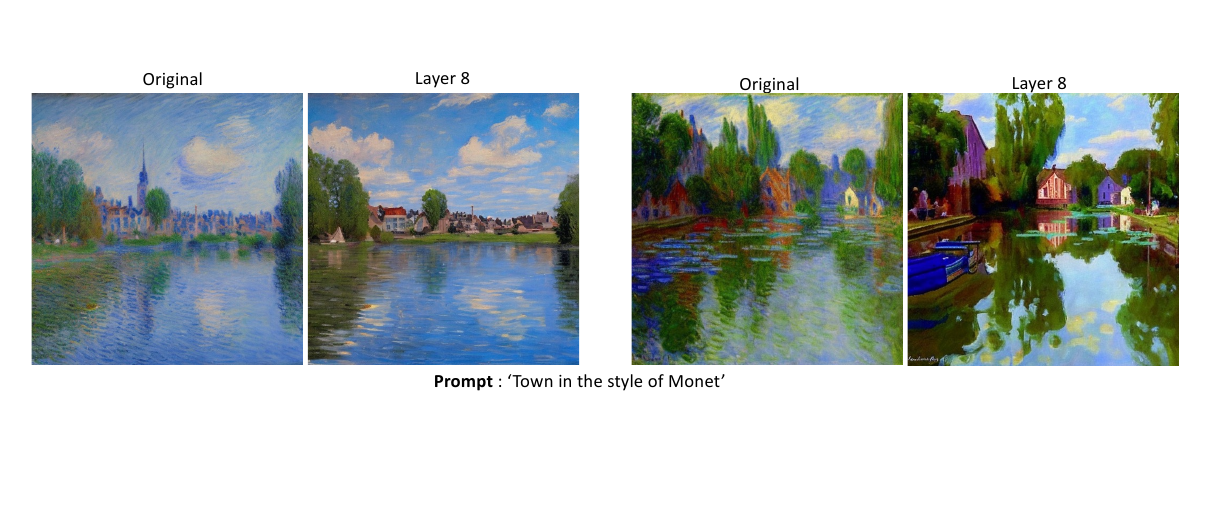}
  \vspace{-0.8cm}
    \caption{\label{sd2_causal-17}
   \textbf{Layer 8 can control ``style knowledge" in Open-Journey}. We perform an intervention in the different cross-attention layers of Open-Journey by using a target prompt - {\it 'a painting'} in those layers while the original prompt is used for other layers. We find that there exists an unique layer which can control output generations of ``style knowledge".  
    }
\end{figure}
\begin{figure}[H]
    \hskip 0.2cm
  \includegraphics[width=\columnwidth]{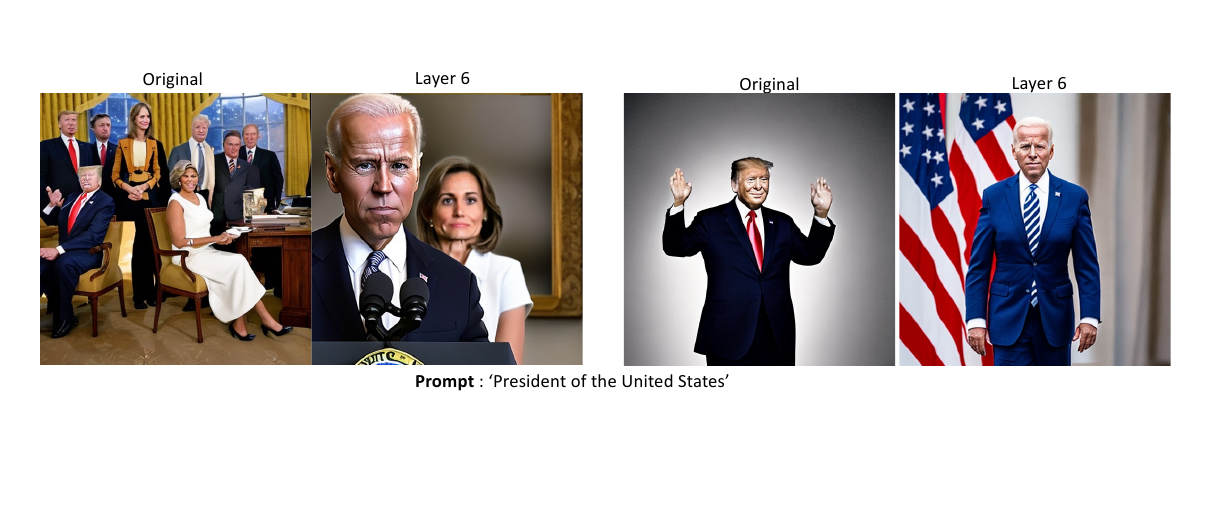}
  \vspace{-0.8cm}
    \caption{\label{sd2_causal-18}
 \textbf{Layer 6 can control ``factual knowledge" in Open-Journey}. We perform an intervention in the different cross-attention layers of Open-Journey by using a target prompt - {\it 'Joe Biden'} in those layers while the original prompt is used for other layers. We find that there exists an unique layer which can control output generations of ``factual knowledge".  
    }
\end{figure}
\begin{figure}[H]
    \hskip 0.2cm
  \includegraphics[width=\columnwidth]{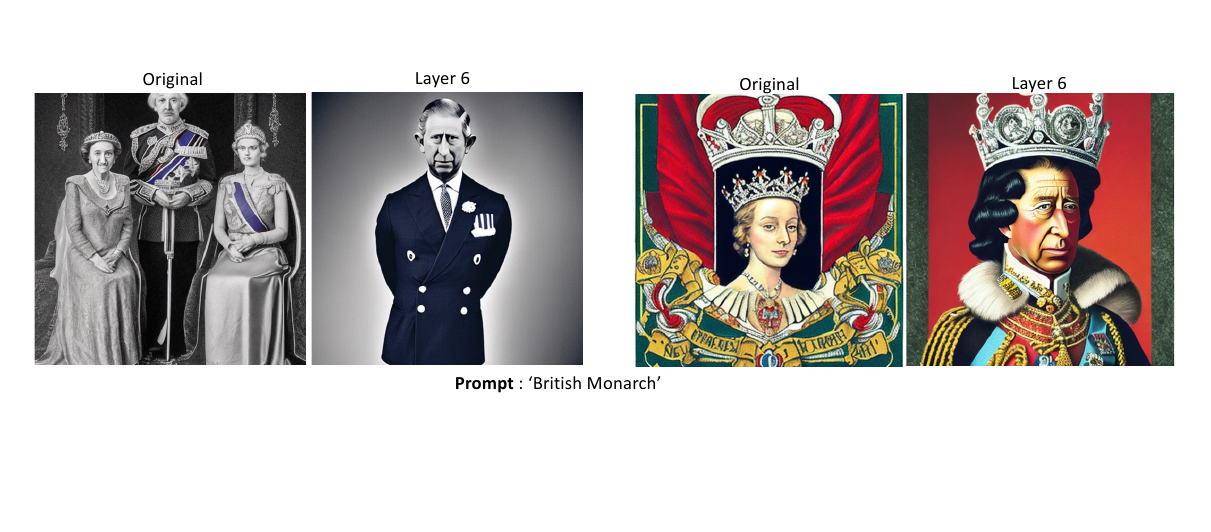}
  \vspace{-0.8cm}
    \caption{\label{sd2_causal-19}
 \textbf{Layer 6 can control ``factual knowledge" in Open-Journey}. We perform an intervention in the different cross-attention layers of Open-Journey by using a target prompt - {\it 'Prince Charles'} in those layers while the original prompt is used for other layers. We find that there exists an unique layer which can control output generations of ``factual knowledge".  
    }
\end{figure}
\begin{figure}[H]
    \hskip 0.2cm
  \includegraphics[width=\columnwidth]{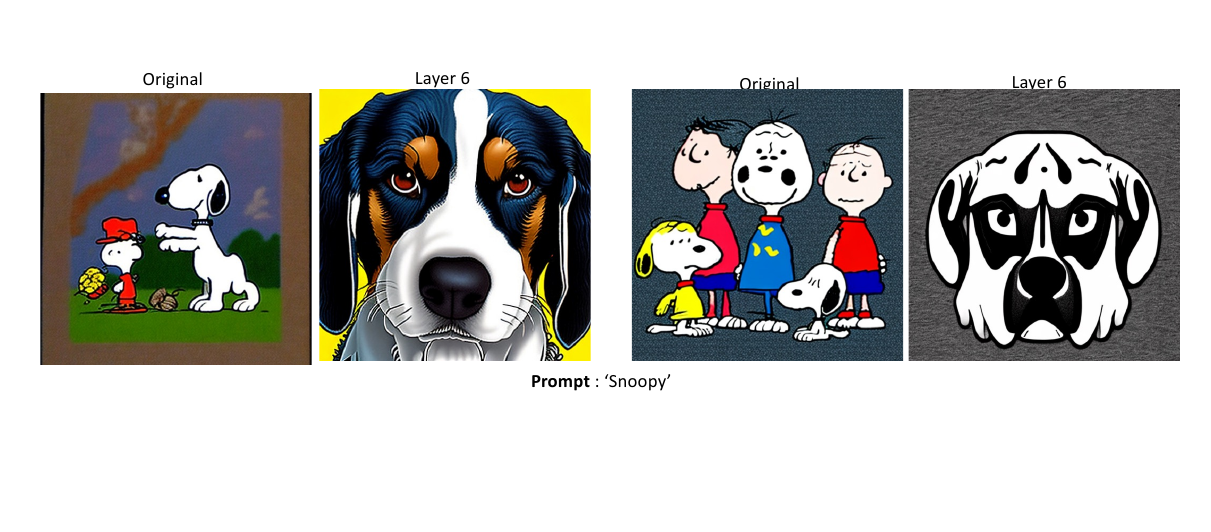}
  \vspace{-0.8cm}
    \caption{\label{sd2_causal-20}
 \textbf{Layer 6 can control "object knowledge" in Open-Journey}. We perform an intervention in the different cross-attention layers of Open-Journey by using a target prompt - {\it 'a dog'} in those layers while the original prompt is used for other layers. We find that there exists an unique layer which can control output generations of "object knowledge".  
    }
\end{figure}
\subsection{SD-XL}
\begin{figure}[H]
    \hskip 0.2cm
  \includegraphics[width=\columnwidth]{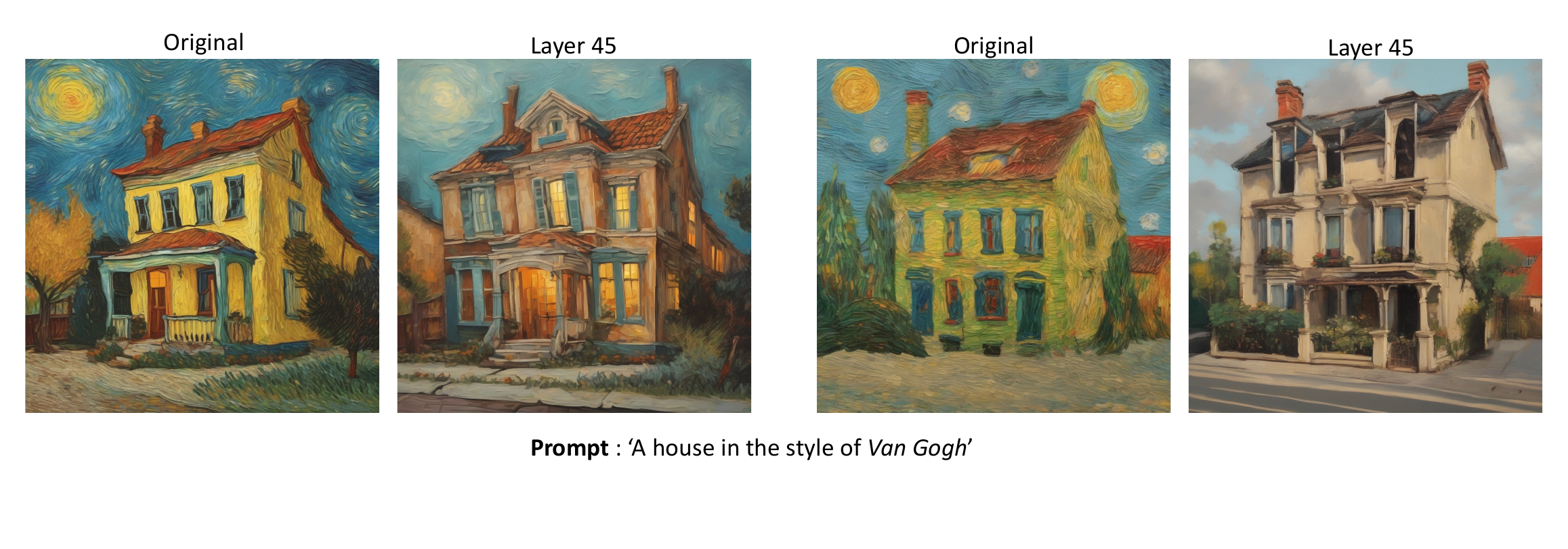}
  \vspace{-0.8cm}
    \caption{\label{sdxl_viz_1}
    \textbf{Layer 45 can control ``style knowledge" in Open-Journey}. We perform an intervention in the different cross-attention layers of Open-Journey by using a target prompt - {\it 'a painting'} in those layers while the original prompt is used for other layers. We find that there exists an unique layer which can control output generations of ``style knowledge".  
    }
\end{figure}
\begin{figure}[H]
    \hskip 0.2cm
  \includegraphics[width=\columnwidth]{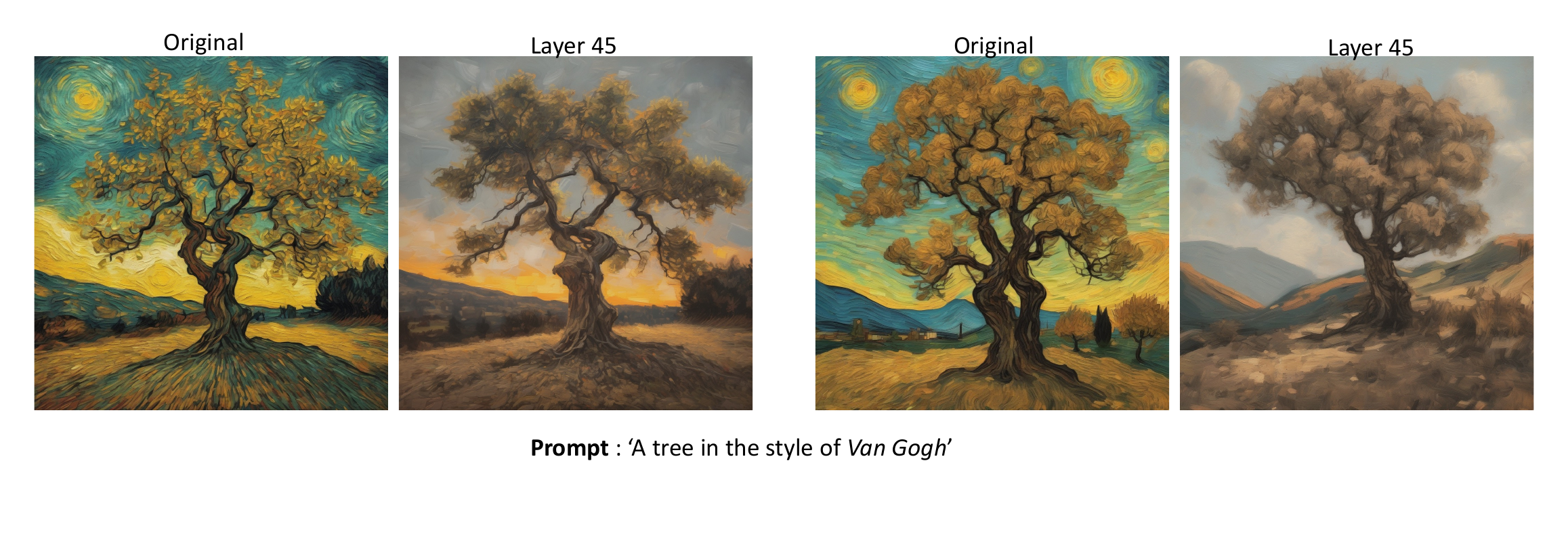}
  \vspace{-0.8cm}
    \caption{\label{sdxl_viz_2}
\textbf{Layer 45 can control ``style knowledge" in Open-Journey}. We perform an intervention in the different cross-attention layers of Open-Journey by using a target prompt - {\it 'a painting'} in those layers while the original prompt is used for other layers. We find that there exists an unique layer which can control output generations of ``style knowledge".  
    }
\end{figure}
\begin{figure}[H]
    \hskip 0.2cm
  \includegraphics[width=\columnwidth]{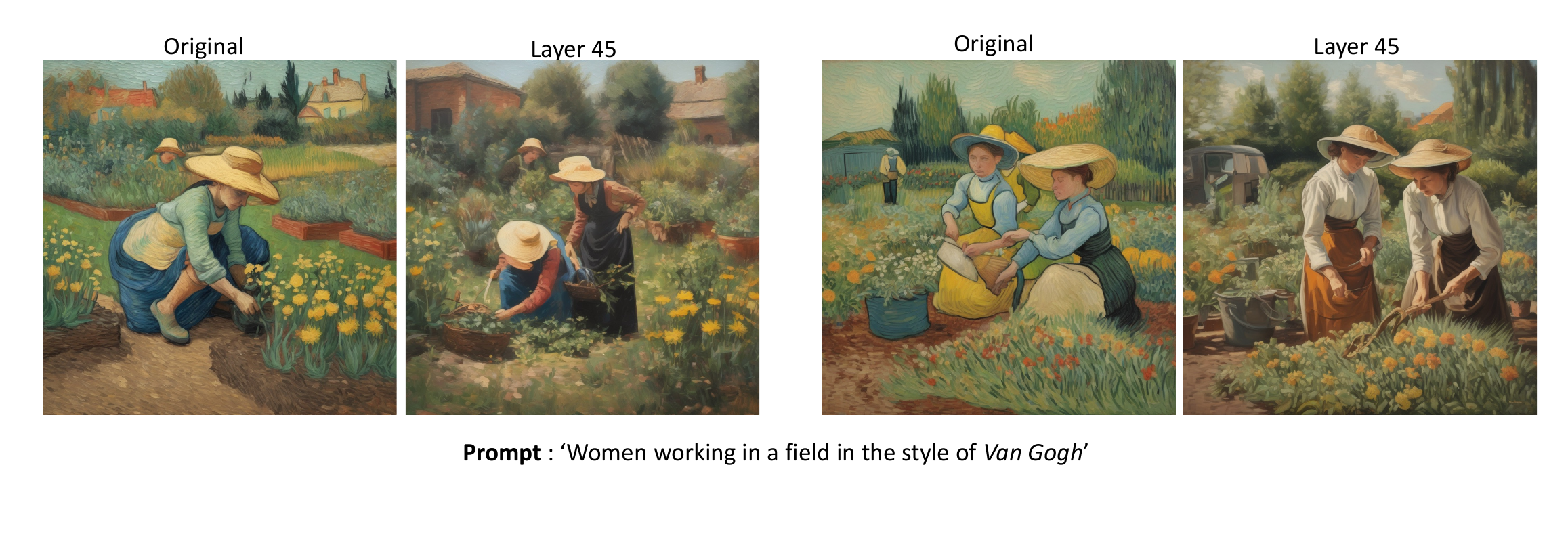}
  \vspace{-0.8cm}
    \caption{\label{sdxl_viz_3-12}
    \textbf{Layer 45 can control ``style knowledge" in Open-Journey}. We perform an intervention in the different cross-attention layers of Open-Journey by using a target prompt - {\it 'a painting'} in those layers while the original prompt is used for other layers. We find that there exists an unique layer which can control output generations of ``style knowledge".  
    }
\end{figure}
\begin{figure}[H]
    \hskip 0.2cm
  \includegraphics[width=\columnwidth]{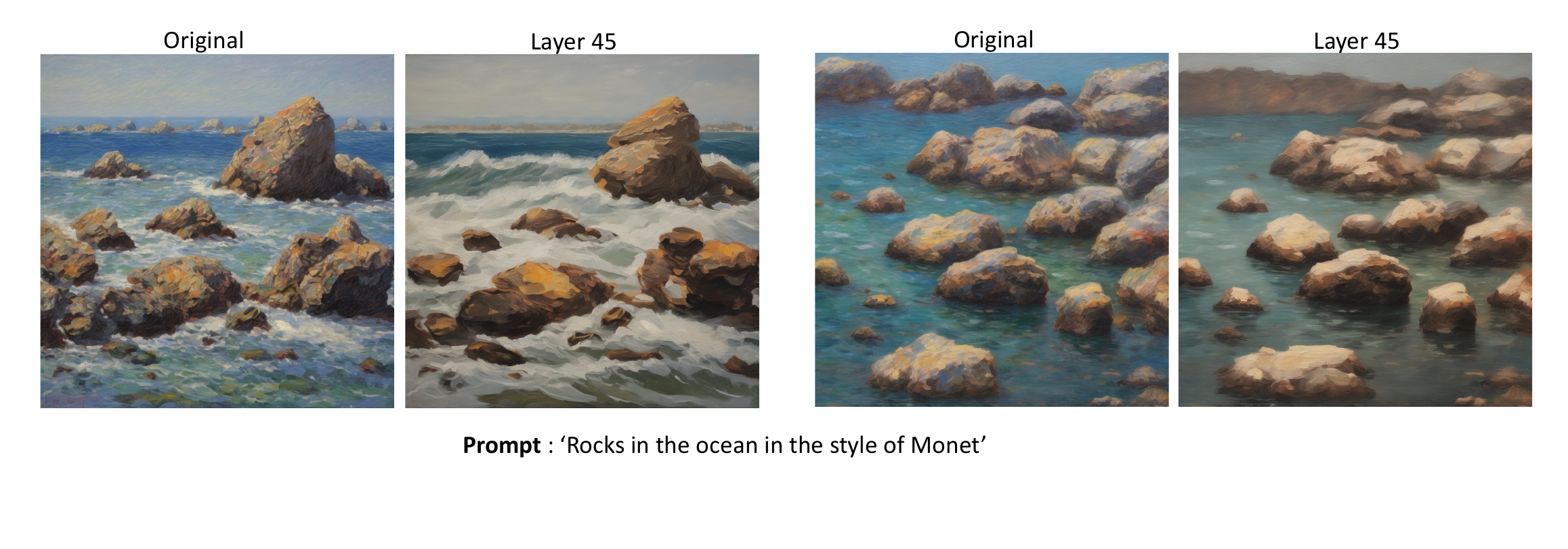}
  \vspace{-0.8cm}
    \caption{\label{sdxl_viz_3-13}
 \textbf{Layer 45 can control ``style knowledge" in Open-Journey}. We perform an intervention in the different cross-attention layers of Open-Journey by using a target prompt - {\it 'a painting'} in those layers while the original prompt is used for other layers. We find that there exists an unique layer which can control output generations of ``style knowledge".  
    }
\end{figure}
\begin{figure}[H]
    \hskip 0.2cm
  \includegraphics[width=\columnwidth]{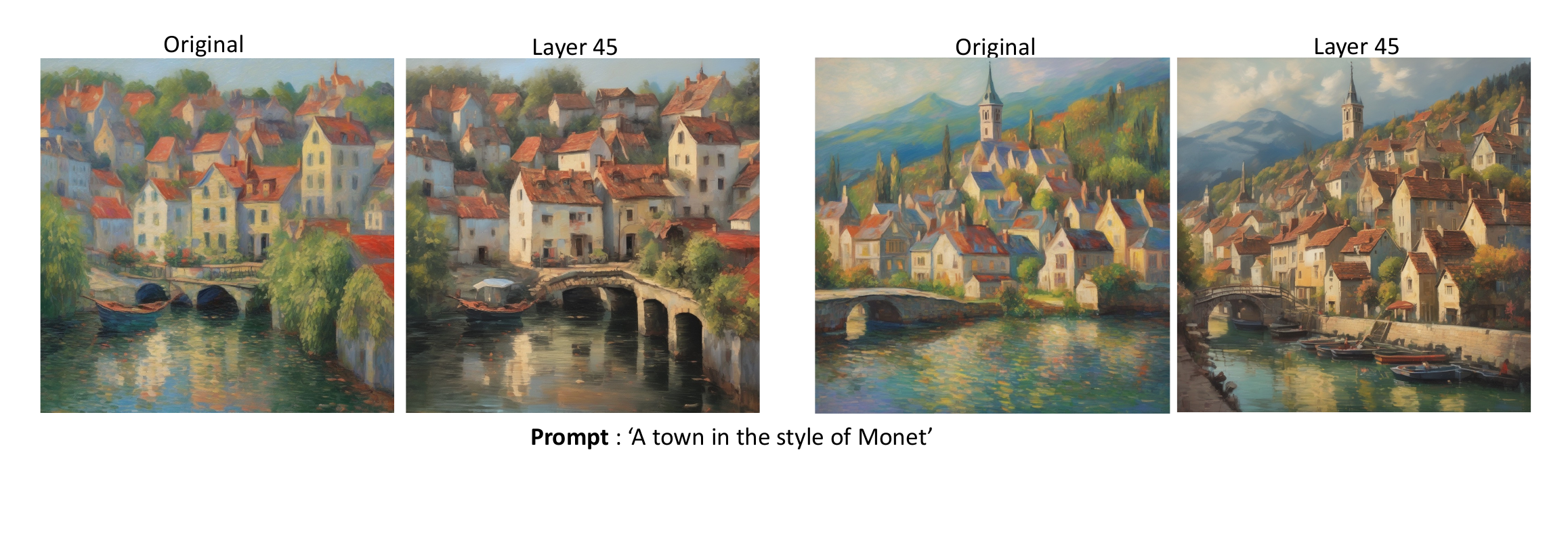}
  \vspace{-0.8cm}
    \caption{\label{sdxl_viz_3-14}
\textbf{Layer 45 can control ``style knowledge" in Open-Journey}. We perform an intervention in the different cross-attention layers of Open-Journey by using a target prompt - {\it 'a painting'} in those layers while the original prompt is used for other layers. We find that there exists an unique layer which can control output generations of ``style knowledge".  
    }
\end{figure}
\begin{figure}[H]
    \hskip 0.2cm
  \includegraphics[width=\columnwidth]{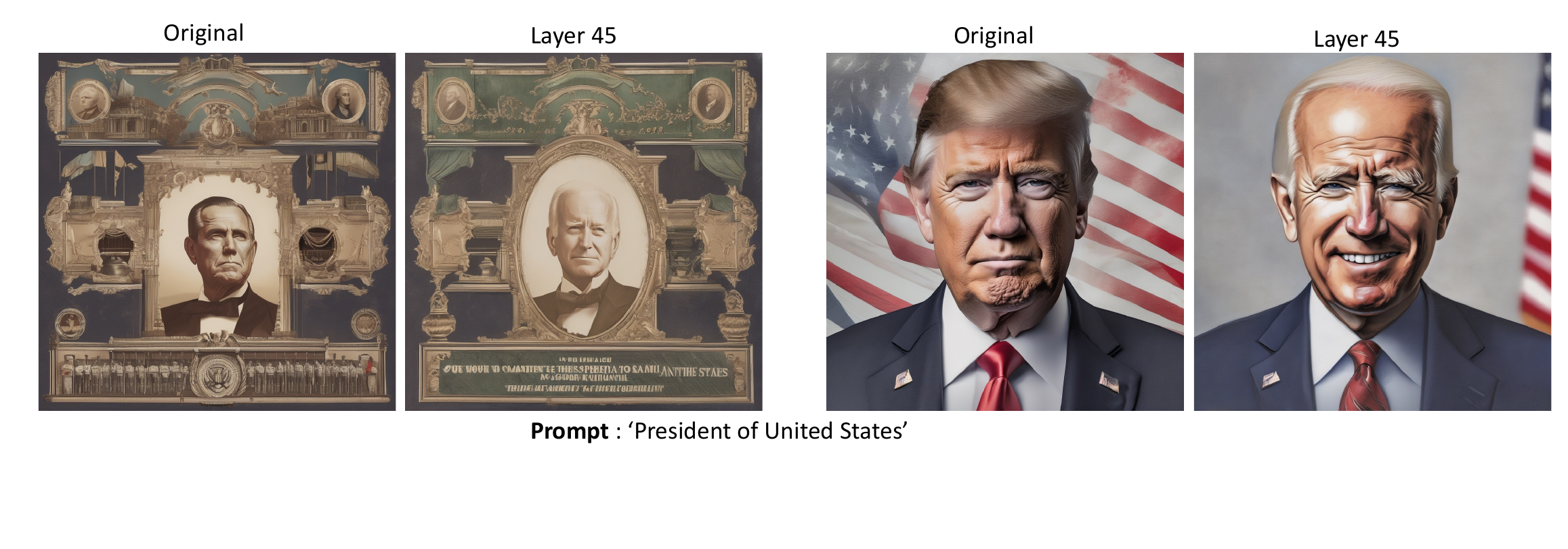}
  \vspace{-0.8cm}
    \caption{\label{sdxl_viz_3-15}
   \textbf{Layer 45 can control ``factual knowledge" in Open-Journey}. We perform an intervention in the different cross-attention layers of Open-Journey by using a target prompt - {\it 'Joe Biden'} in those layers while the original prompt is used for other layers. We find that there exists an unique layer which can control output generations of ``factual knowledge".  
    }
\end{figure}
\begin{figure}[H]
    \hskip 0.2cm
  \includegraphics[width=\columnwidth]{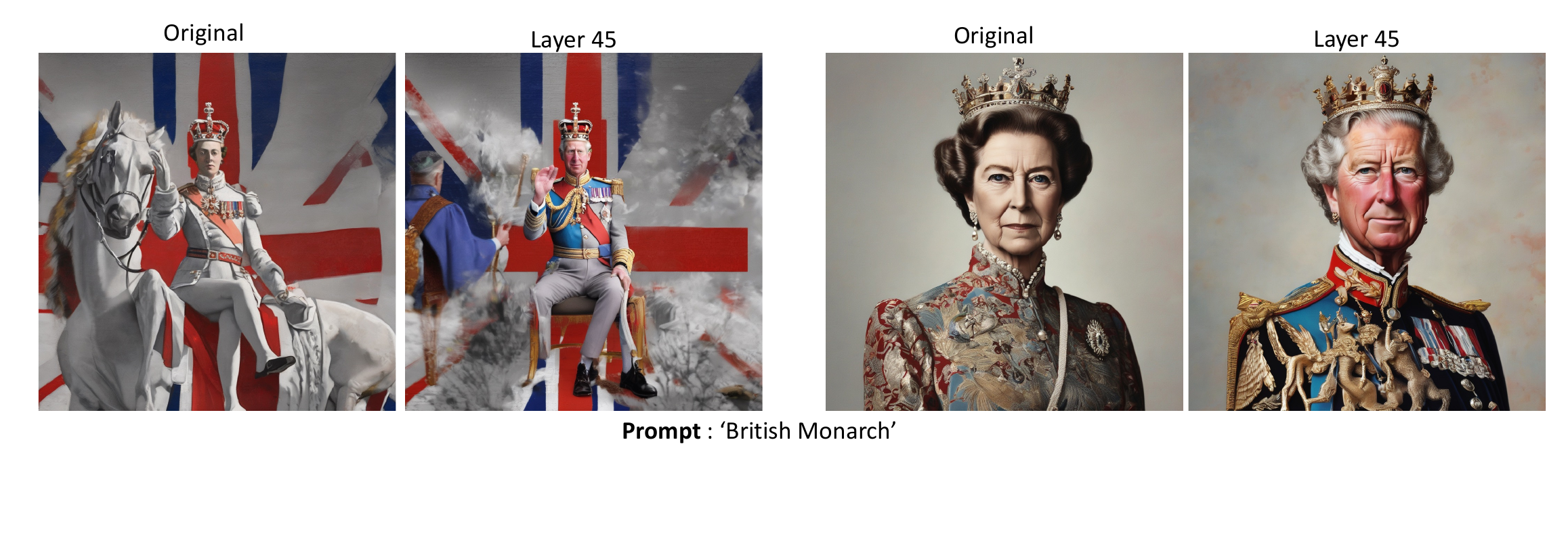}
  \vspace{-0.8cm}
    \caption{\label{sdxl_viz_3-16}
    \textbf{Layer 45 can control ``factual knowledge" in Open-Journey}. We perform an intervention in the different cross-attention layers of Open-Journey by using a target prompt - {\it 'Prince Charles'} in those layers while the original prompt is used for other layers. We find that there exists an unique layer which can control output generations of ``factual knowledge".  
    }
\end{figure}
\begin{figure}[H]
    \hskip 0.2cm
  \includegraphics[width=\columnwidth]{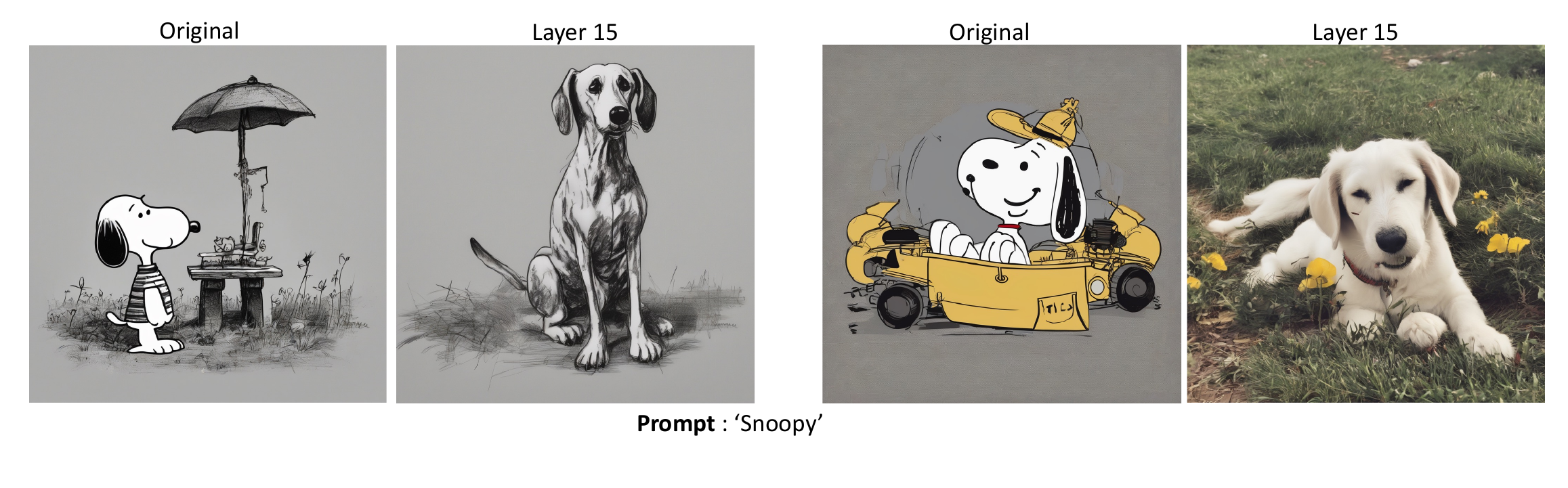}
  \vspace{-0.8cm}
    \caption{\label{sdxl_viz_3-snoop}
    \textbf{Layer 15 can control "object knowledge" in Open-Journey}. We perform an intervention in the different cross-attention layers of Open-Journey by using a target prompt - {\it 'a dog'} in those layers while the original prompt is used for other layers. We find that there exists an unique layer which can control output generations of "object knowledge".  
    }
\end{figure}
\subsection{DeepFloyd}
\label{df_interpretability}
\begin{figure}[H]
    \hskip 0.2cm
  \includegraphics[width=\columnwidth]{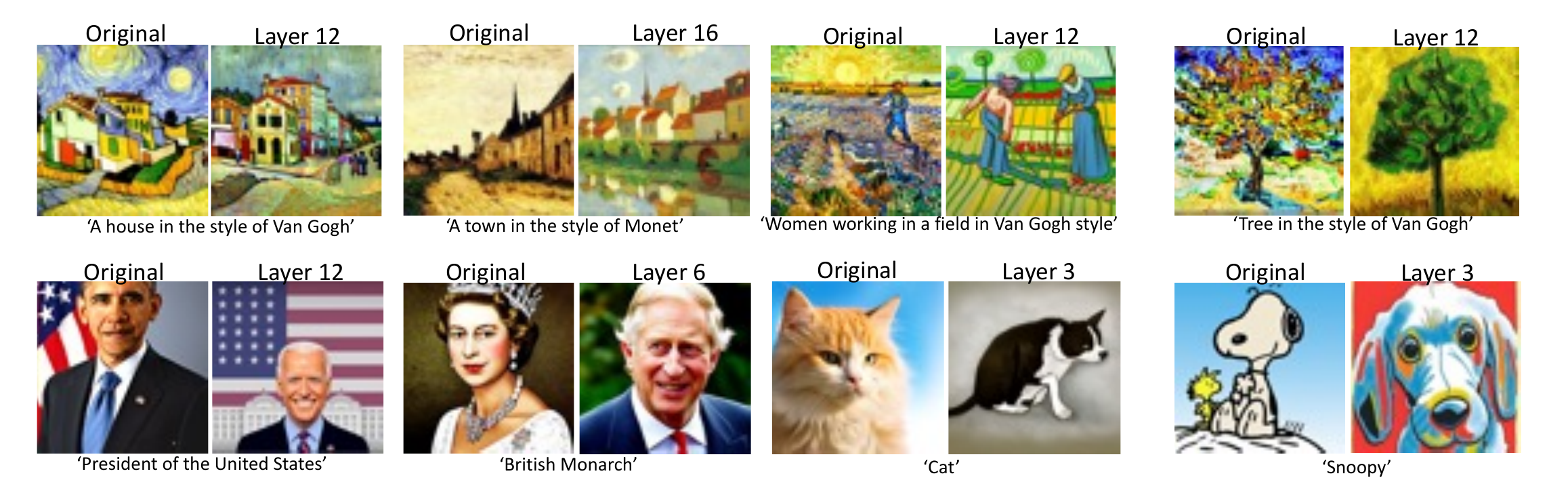}
  \vspace{-0.8cm}
    \caption{\label{sdxl_viz_3-total}
    \textbf{Knowledge Tracing in Deepfloyd}. We find that control regions for a visual attribute can be different, depending on the prompt. For e.g., prompts involving {\it Van Gogh} style can be modified from Layer 12, whereas for {\it Monet} style -- modifications are needed from Layer 16. For ``facts", prompts involving {\it The President of the United States} can be controlled from Layer 12, whereas prompts involving {\it British Monarch} can be controlled from Layer 6. 
    }
\end{figure}


\section{Prompt Dataset}
\label{prompt_dataset}
\textbf{For Interpretability and Model Editing. } We use the benchmark dataset from~\citep{basu2023localizing} and~\citep{kumari2023ablating} for obtaining prompts for ``objects", ``style" and ``facts".  For~\crossprompt{}, we select the target prompt based on the attribute of interest. For e.g., in the case of ``style",  we select the target prompt as {\it 'A painting'}. For ``facts", we use the correct answer to a given fact, as the target prompt. For e.g., the target prompt for {\it 'The President of the United States'} is {\it 'Joe Biden'} and the target prompt for {\it 'The British Monarch'} is {\it 'Prince Charles'}. For ``objects",  the target prompt for {\it 'r2d2'} is {\it 'robot'}, for {\it 'snoopy'} is {\it 'dog'}, for {\it 'nemo'} is {\it 'fish'} and for {\it 'cat'} is {\it 'dog'}. These sets of trademarked ``objects" and ``facts" are chosen from~\citep{basu2023localizing}, whereas the ``style" prompts are chosen from~\citep{kumari2023ablating}.

\section{Layer Information Across Different Text-to-Image Models}
\begin{figure}[H]
    \hskip 0.2cm
  \includegraphics[width=\columnwidth]{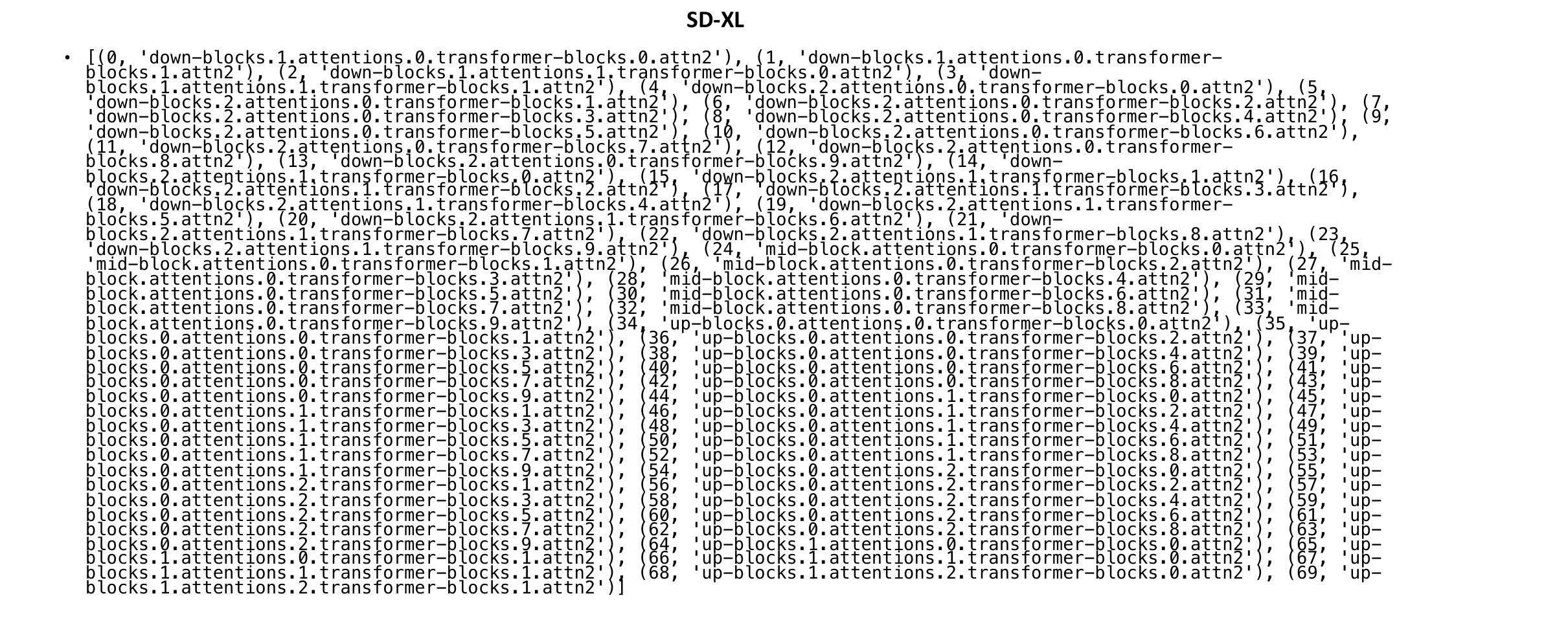}
  \vspace{-0.8cm}
    \caption{\label{sdxl_viz_3-p1}
    \textbf{Layers to Probe for SD-XL}. Indexing of cross-attention layers in the UNet. 
    }
\end{figure}
\begin{figure}[H]
    \hskip 0.2cm
  \includegraphics[width=\columnwidth]{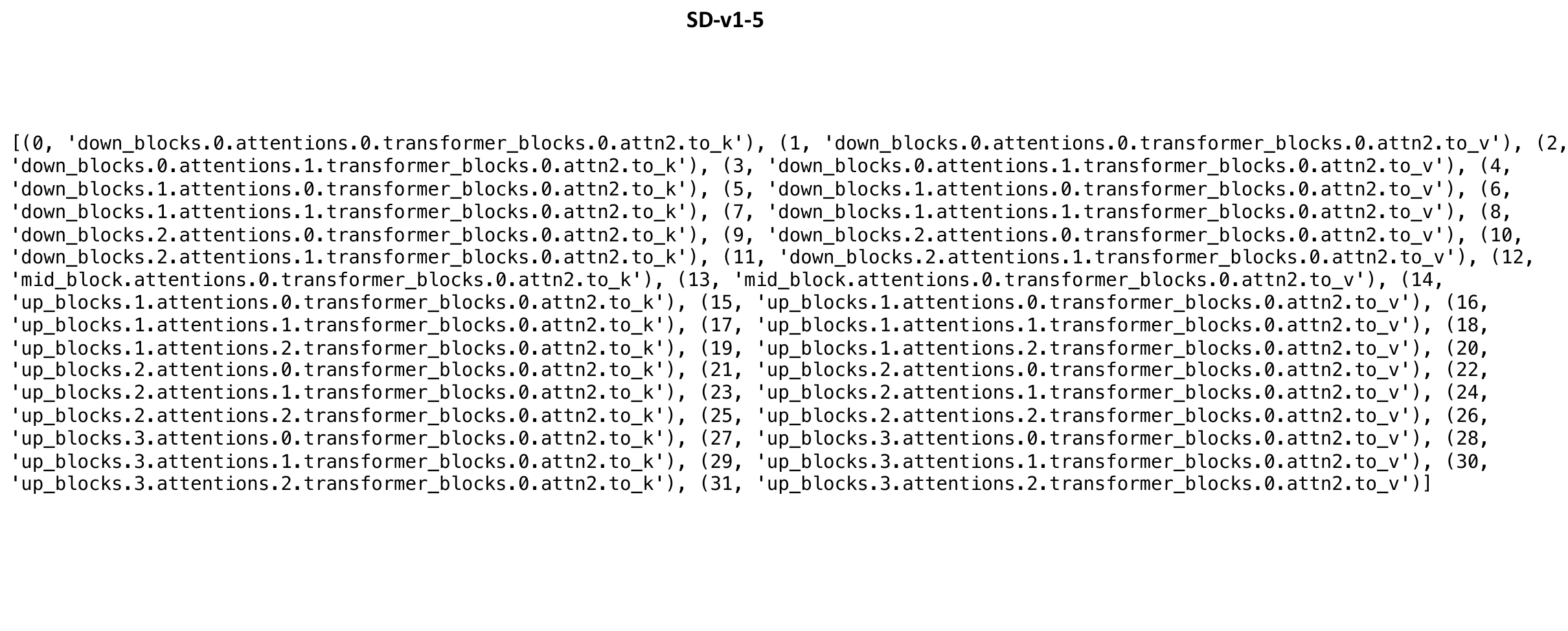}
  \vspace{-0.8cm}
    \caption{\label{sdxl_viz_3-p2}
    \textbf{Layers to Probe for SD-v1-5}. Indexing of cross-attention layers in the UNet. 
    }
\end{figure}
\begin{figure}[H]
    \hskip 0.2cm
  \includegraphics[width=\columnwidth]{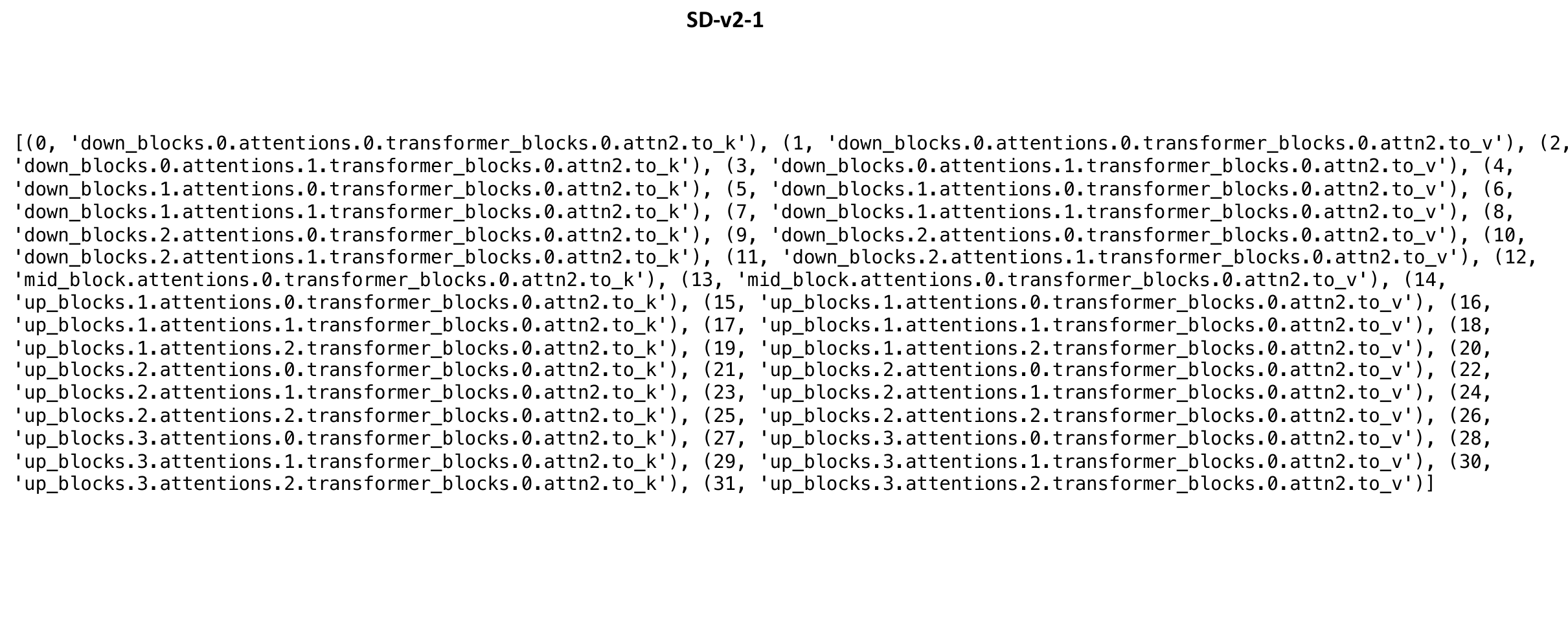}
  \vspace{-0.8cm}
    \caption{\label{sdxl_viz_3-p3}
    \textbf{Layers to Probe for SD-v2-1}. Indexing of cross-attention layers in the UNet. 
    }
\end{figure}
\begin{figure}[H]
    \hskip 0.2cm
  \includegraphics[width=\columnwidth]{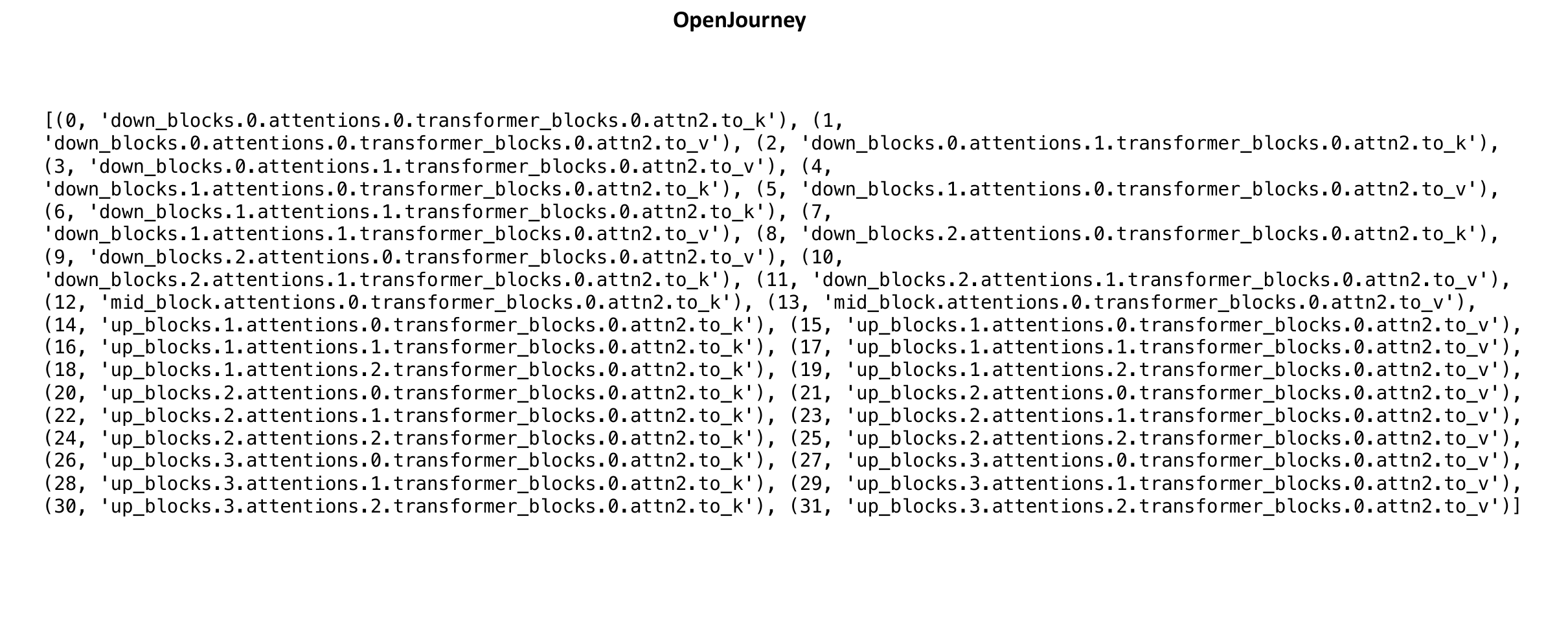}
  \vspace{-0.8cm}
    \caption{\label{sdxl_viz_3-p4}
    \textbf{Layers to Probe for OpenJourney}. Indexing of cross-attention layers in the UNet. 
    }
\end{figure}
\begin{figure}[H]
    \hskip 0.2cm
  \includegraphics[width=\columnwidth]{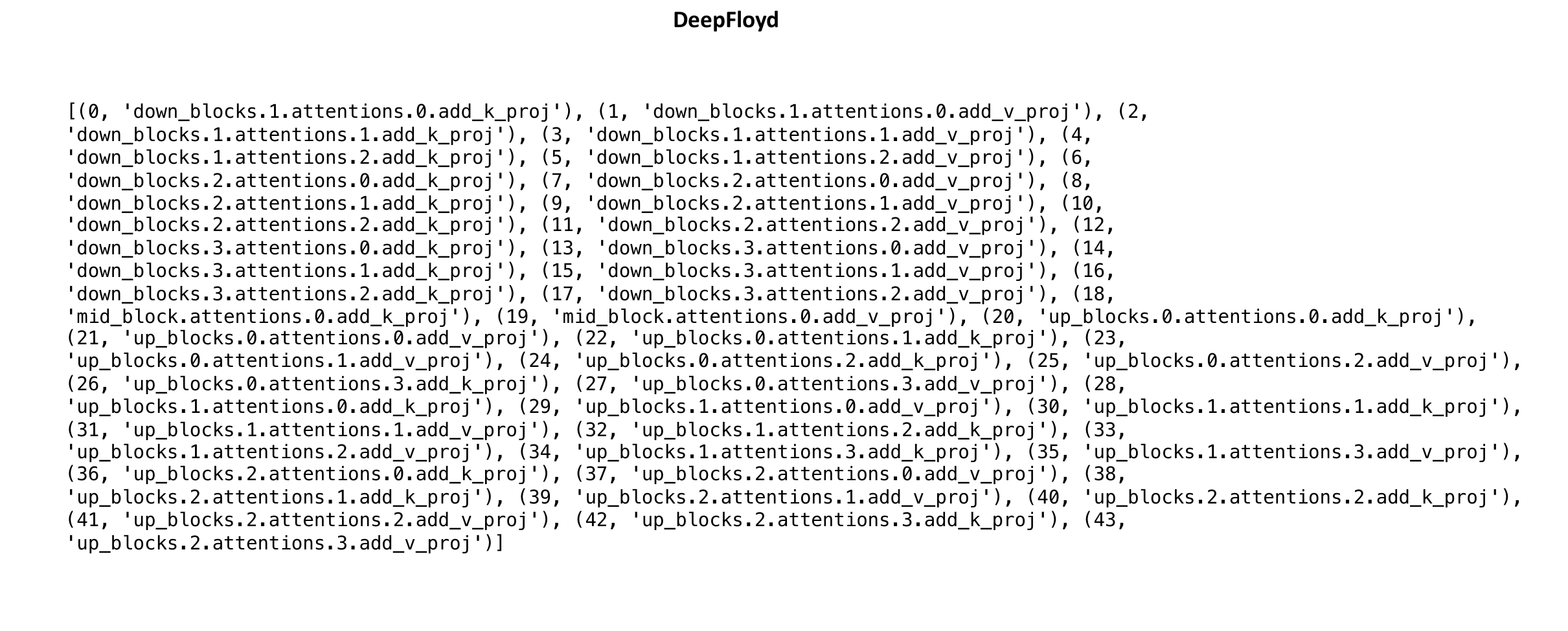}
  \vspace{-0.8cm}
    \caption{\label{sdxl_viz_3-deepfloyd}
     \textbf{Layers to Probe for DeepFloyd}. Indexing of cross-attention layers in the UNet. 
    }
\end{figure}
\section{More Visualizations for Model Editing}
\begin{figure}[H]
    \hskip 0.2cm
  \includegraphics[width=\columnwidth]{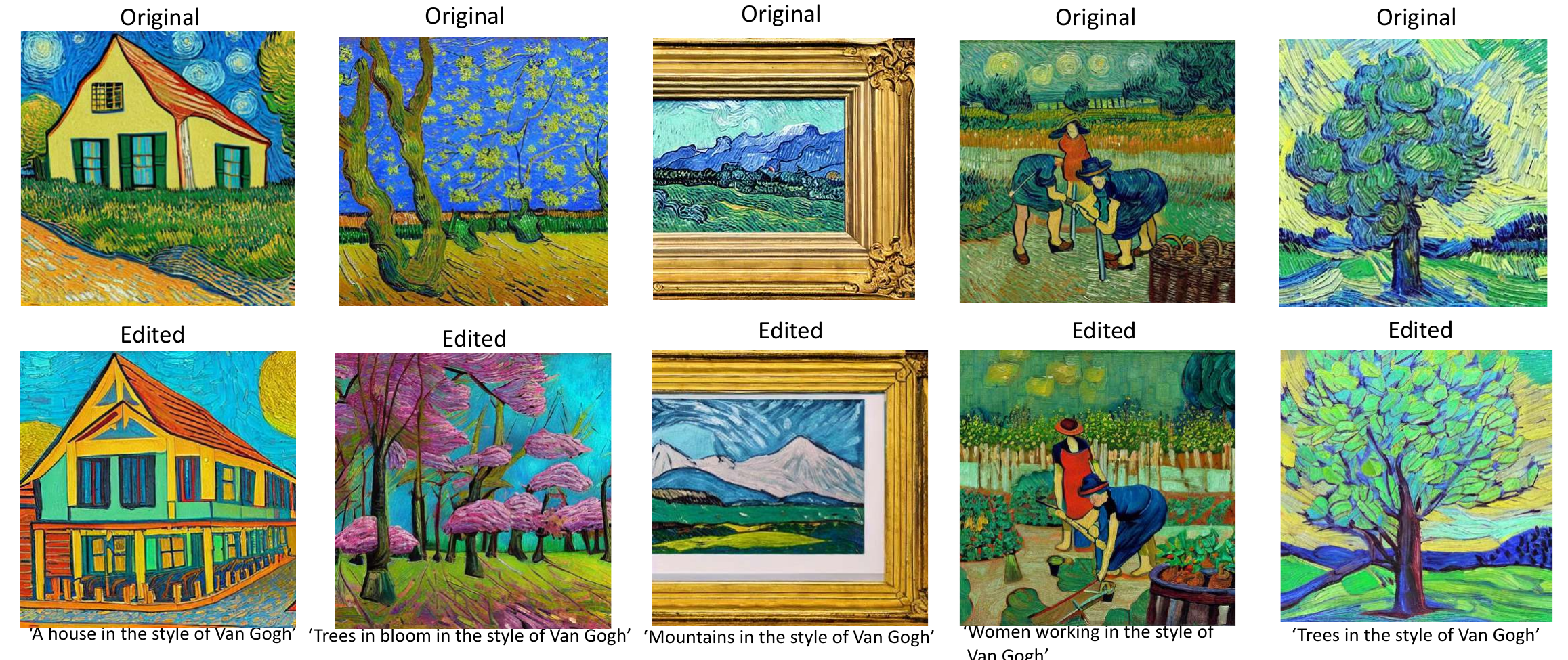}
  \vspace{-0.8cm}
    \caption{\label{sdxl_viz_3-1-p1}
    \textbf{SDv1-5 Edits for ``style"}. We show successful model editing on the layers identified by~\crossprompt. In case of the images generated by the edited model, we can observe that the trademarked brushstrokes of the artist {\it Van Gogh} are missing. 
    }
\end{figure}
\begin{figure}[H]
    \hskip 0.2cm
  \includegraphics[width=\columnwidth]{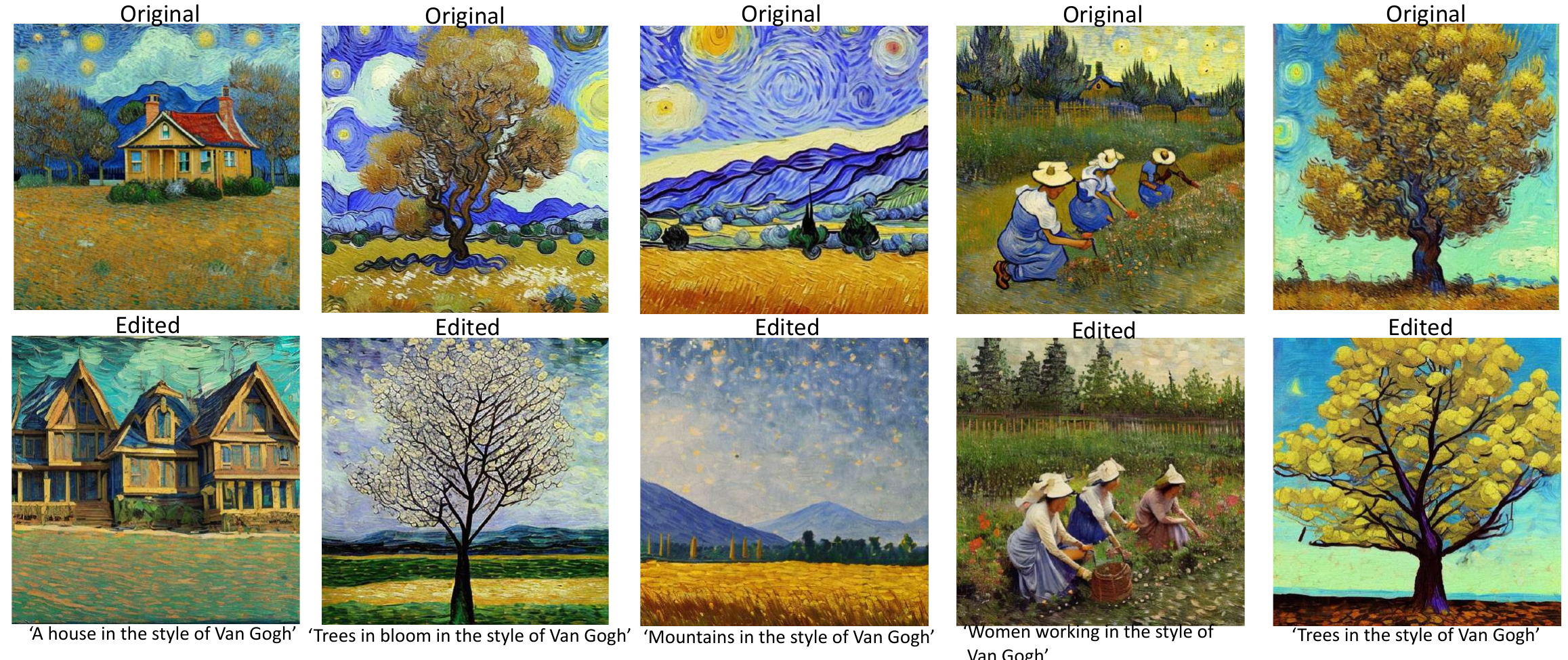}
  \vspace{-0.8cm}
    \caption{\label{sdxl_viz_3-all-p2}
  \textbf{Open-Journey Edits for ``style"}. We show successful model editing on the layers identified by~\crossprompt. In case of the images generated by the edited model, we can observe that the trademarked brushstrokes of the artist {\it Van Gogh} are missing. For some of the images from the edited model, we even find that the patterns in the sky which is another trademark {\it Van Gogh} signature have them deleted.
    }
\end{figure}
\begin{figure}[H]
    \hskip 0.2cm
  \includegraphics[width=\columnwidth]{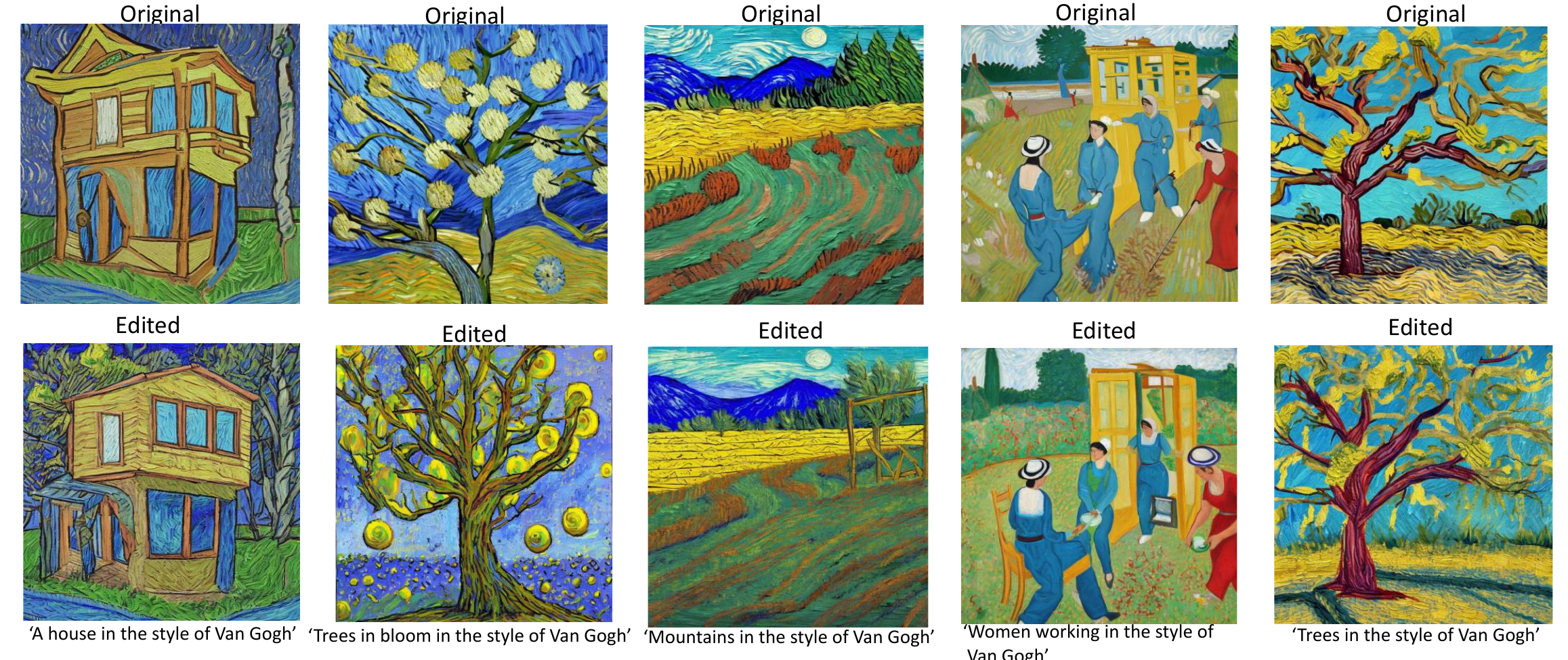}
  \vspace{-0.8cm}
    \caption{\label{sdxl_viz_3-all-p3}
    \textbf{SD-v2-1 Edits for ``style"}. We show successful model editing on the layers identified by~\crossprompt. In case of the images generated by the edited model, we can observe that the trademarked brushstrokes of the artist {\it Van Gogh} are missing. For some of the images from the edited model, we even find that the patterns in the sky which is another trademark {\it Van Gogh} signature have them deleted.
    }
\end{figure}
\begin{figure}[H]
    \hskip 0.2cm
  \includegraphics[width=\columnwidth]{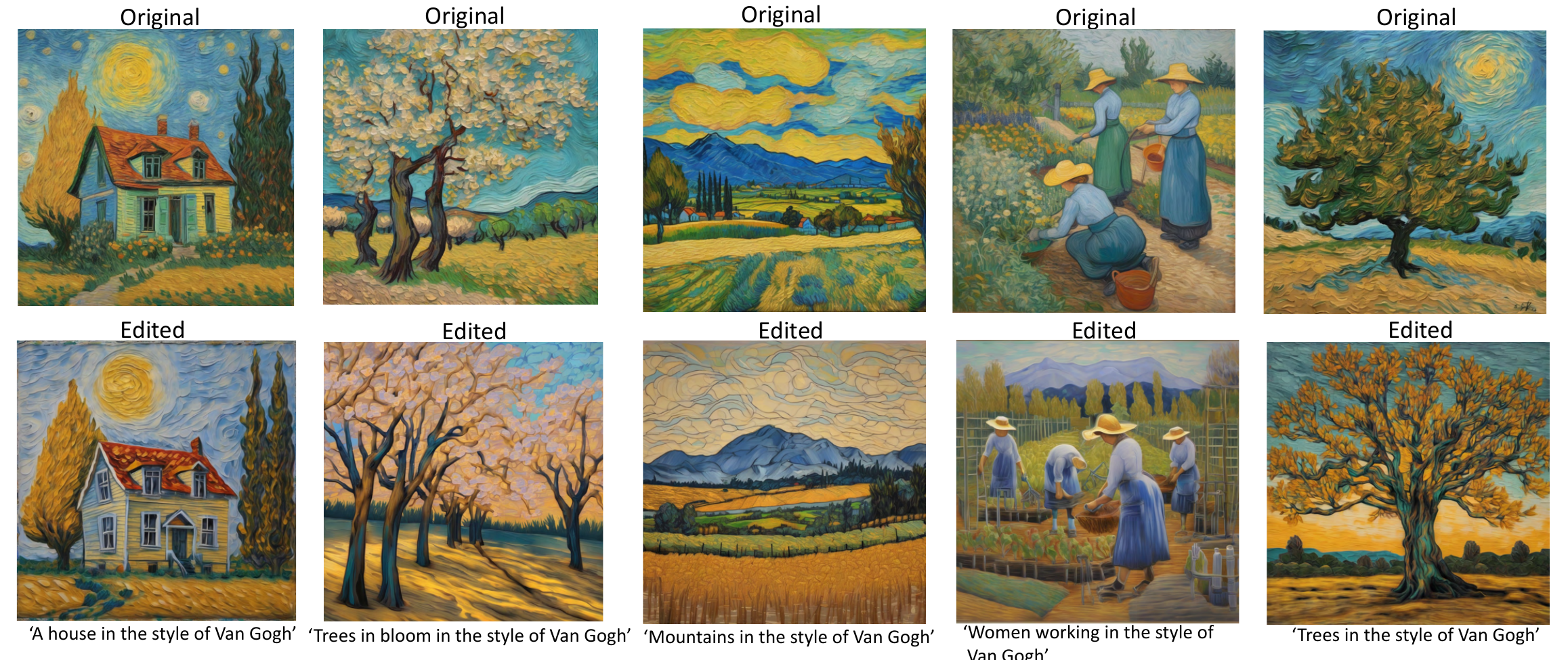}
  \vspace{-0.8cm}
    \caption{\label{sdxl_viz_3-all-p4}
    \textbf{SDXL Edits for ``style"}. We show successful model editing on the layers identified by~\crossprompt. In case of the images generated by the edited model, we can observe that the trademarked brushstrokes of the artist {\it Van Gogh} are missing. For some of the images from the edited model (e.g., {Trees in the style of Van Gogh}), we even find that the patterns in the sky which is another trademark {\it Van Gogh} signature have them deleted.
    }
\end{figure}
\begin{figure}[H]
    \hskip 0.2cm
  \includegraphics[width=\columnwidth]{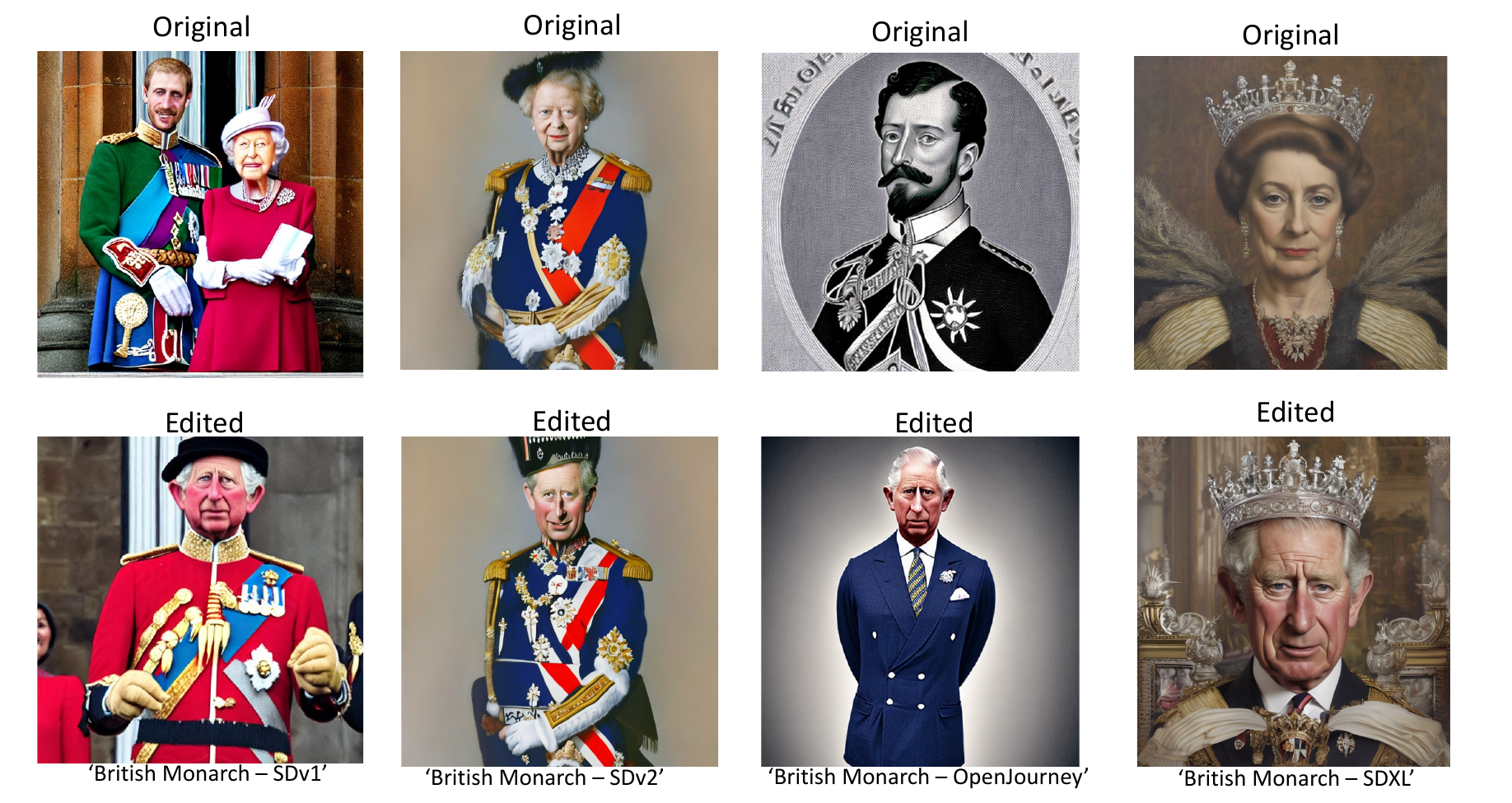}
  \vspace{-0.8cm}
    \caption{\label{sdxl_viz_3-pp1}
    \textbf{Fact Editing Across Different Text-to-Image Models.}. We show that ~\crossedit \hspace{0.05cm} can successfully update out-dated facts in text-to-image models with the correct facts. 
    }
\end{figure}
\begin{figure}[H]
    \hskip 0.2cm
  \includegraphics[width=\columnwidth]{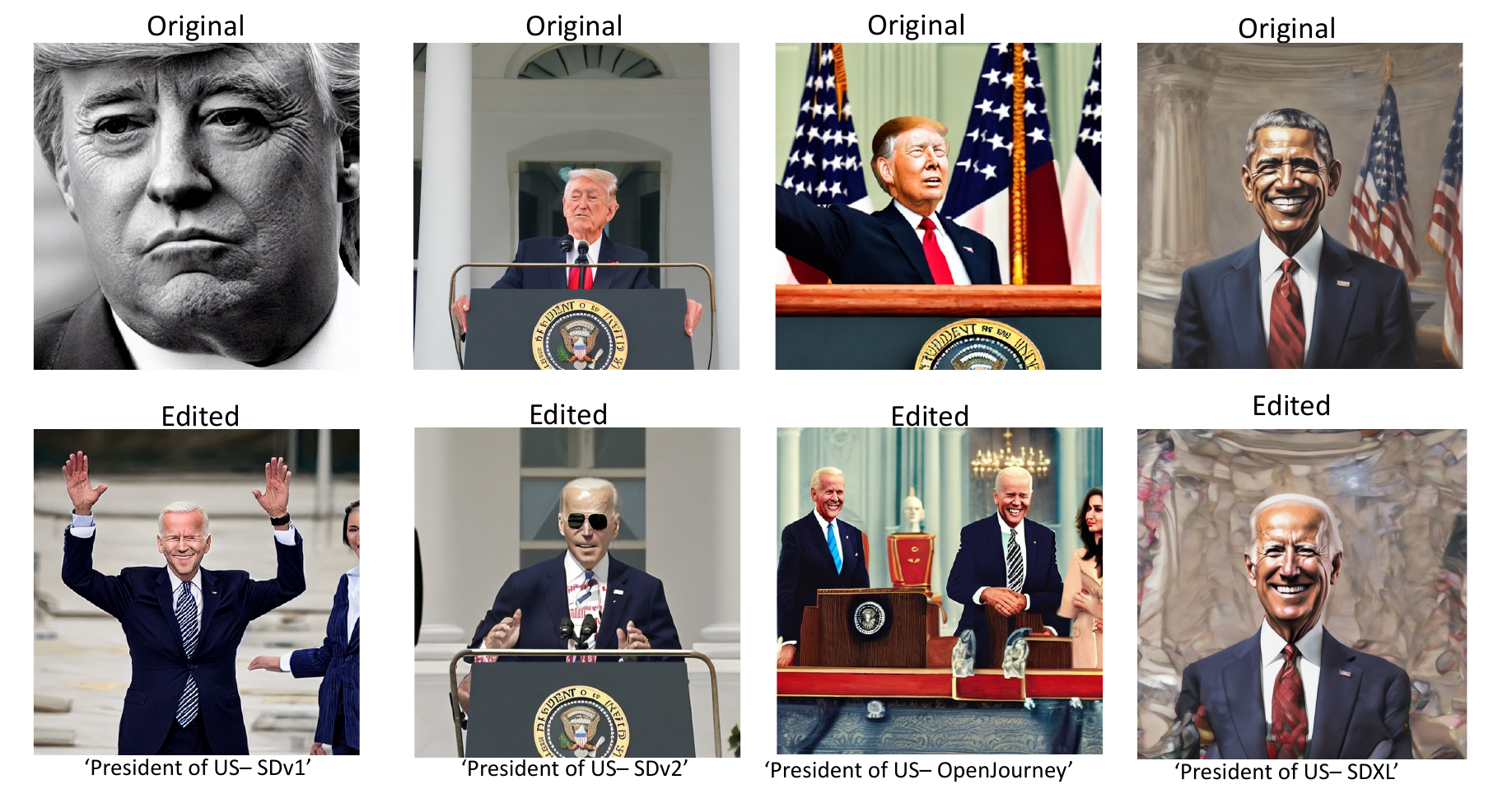}
  \vspace{-0.8cm}
    \caption{\label{sdxl_viz_3-pp2}
      \textbf{Fact Editing Across Different Text-to-Image Models.}. We show that ~\crossedit \hspace{0.05cm} can successfully update out-dated facts in text-to-image models with the correct facts. 
    }
\end{figure}
\begin{figure}[H]
    \hskip 0.2cm
  \includegraphics[width=\columnwidth]{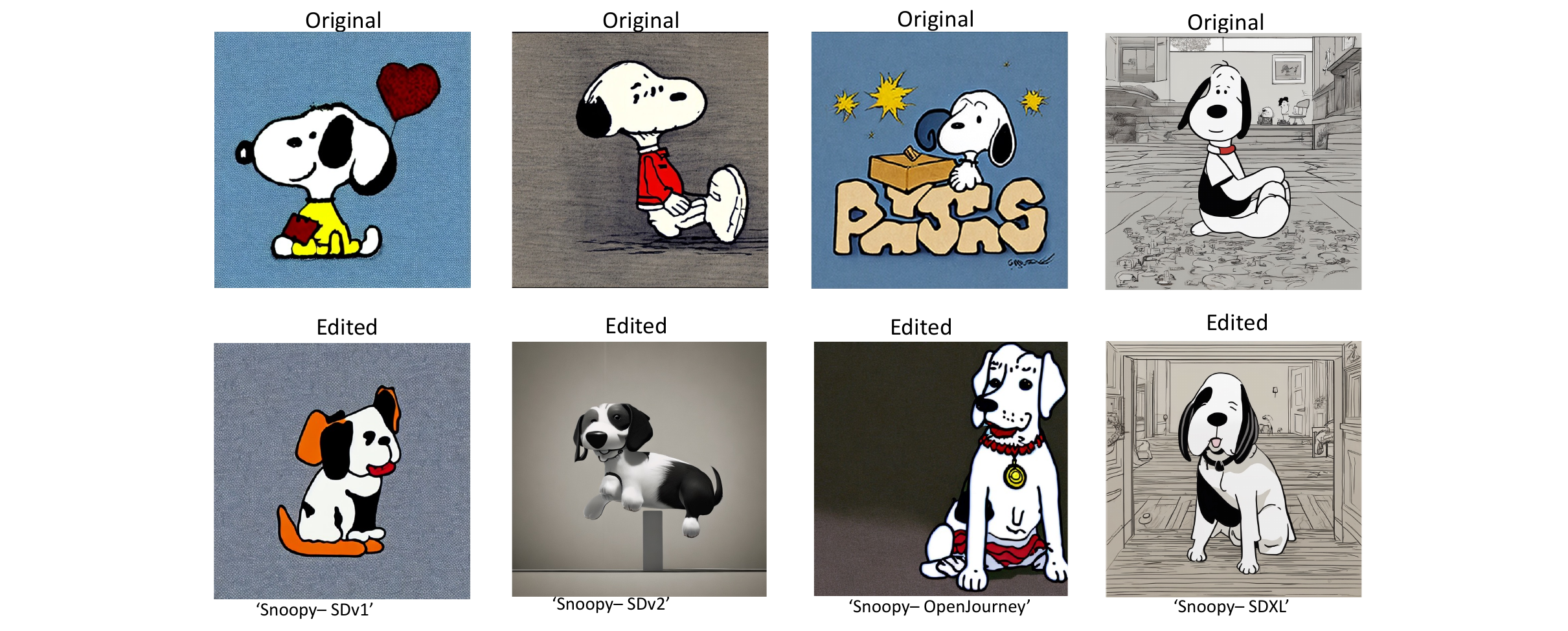}
  \vspace{-0.8cm}
    \caption{\label{sdxl_viz_3-pp3}
      \textbf{Trademarked Object Editing Across Different Text-to-Image Models.}. We show that ~\crossedit \hspace{0.05cm} can successfully modify trademarked objects by performing weight-space editing in the locations identified by~\crossprompt. 
    }
\end{figure}
\begin{figure}[H]
    \hskip 0.3cm
  \includegraphics[width=\columnwidth]{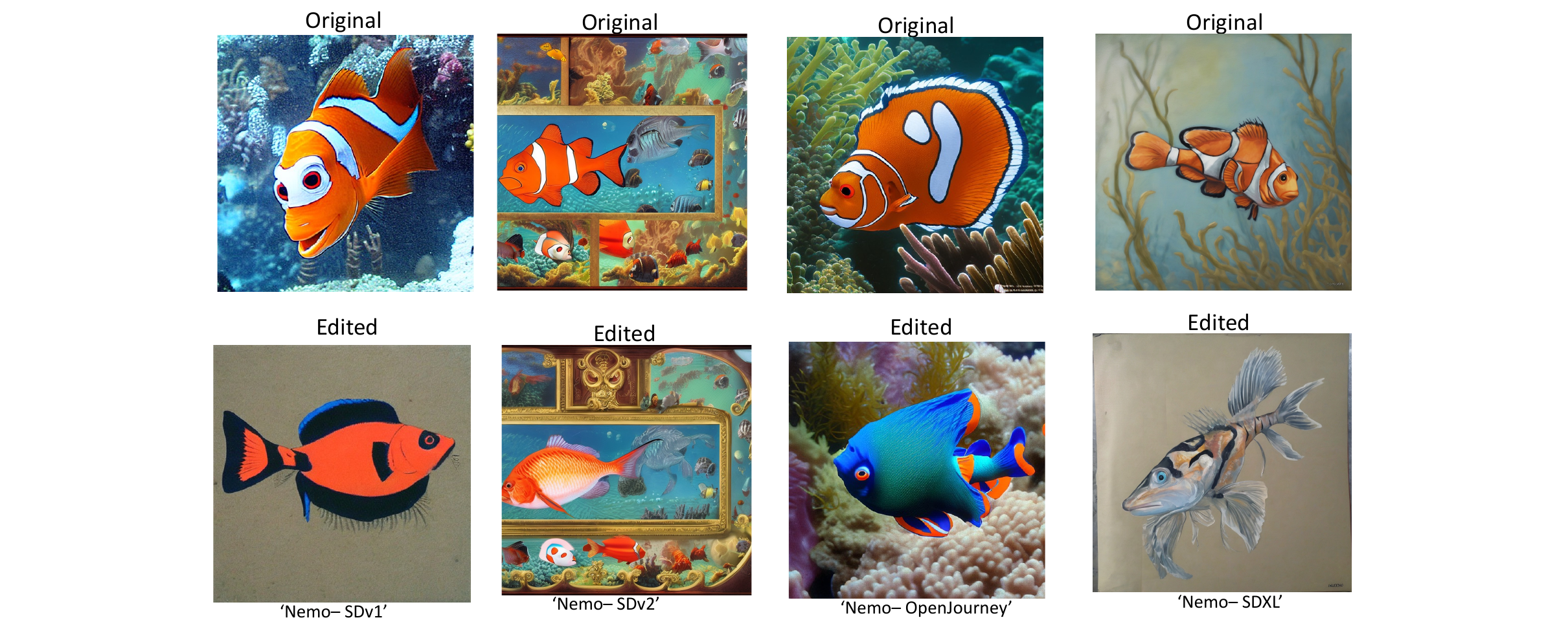}
  \vspace{-0.8cm}
    \caption{\label{sdxl_viz_3-ep2}
        \textbf{Trademarked Object Editing Across Different Text-to-Image Models.}. We show that ~\crossedit \hspace{0.05cm} can successfully modify trademarked objects by performing weight-space editing in the locations identified by~\crossprompt. 
    }
\end{figure}
\section{Updating All Layers vs. Localized Layers}
\label{update_all}
\begin{figure}[H]
    \hskip 0.2cm
  \includegraphics[width=\columnwidth]{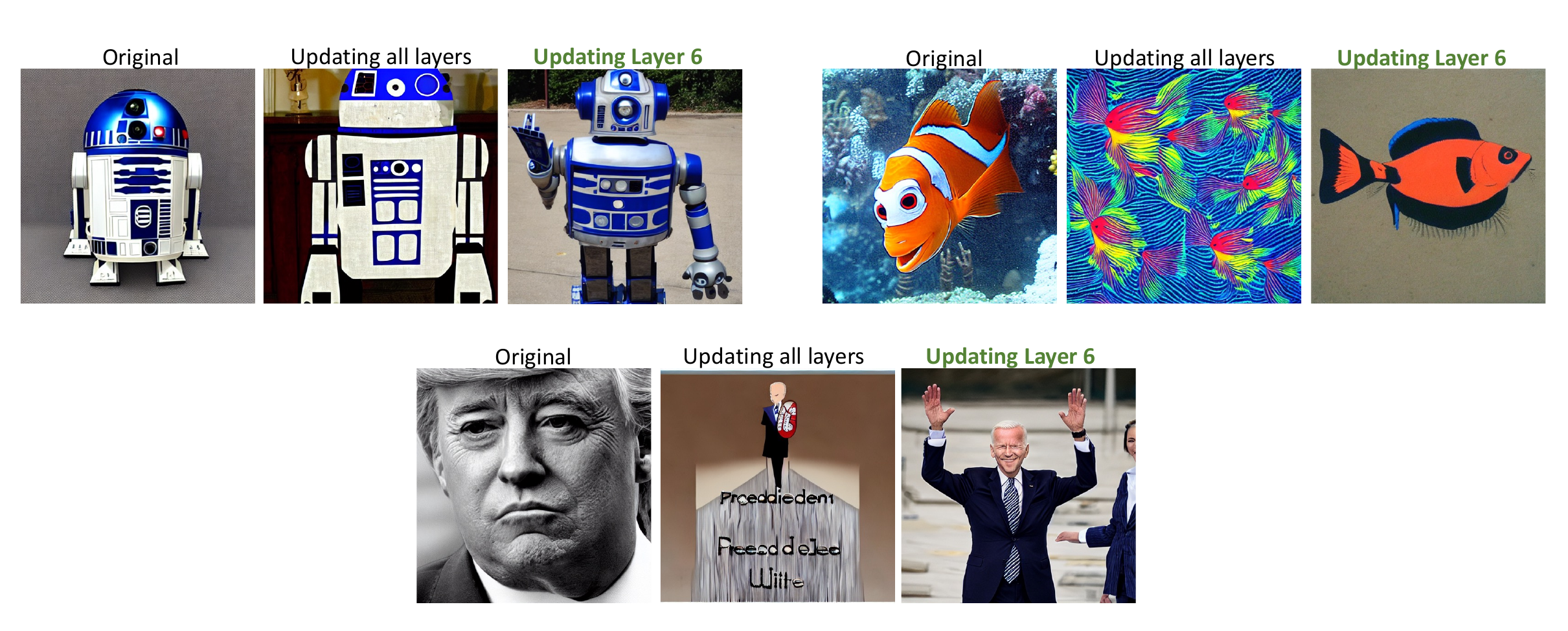}
  \vspace{-0.8cm}
    \caption{\label{sdxl_viz_3-all}
\textbf{Localized Editing vs. Non-Localized Editing}. For "objects" and ``facts", we find that updating the layers identified by~\crossprompt \hspace{0.05cm} is better than editing all the layers. For e.g., in this qualitative study, we find that updating all the layers does not lead to correct outputs for certain cases involving prompts corresponding to "objects" and ``facts". 
    }
\end{figure}
\section{Hyper-parameter Search}
\label{hyperparameter}
In this section, we enumerate the hyper-parameter $m$ for each text-to-image model. In particular, we use Stable-Diffusion-v1-5 as a base to first obtain the optimal value of $m$. For Stable-Diffusion-v1-5, we find this value to be $m=2$. Based on the percentage of the total layers, in the model it encompasses, we perform a local search around that hyper-parameter. In this way, we find the optimal value of $m=3$ for Stable-Diffusion-v2-1, $m=2$ for OpenJourney, $m=5$ for SD-XL and $m=3$ for DeepFloyd. 
\begin{figure}[H]
    \hskip 0.2cm
  \includegraphics[width=\columnwidth]{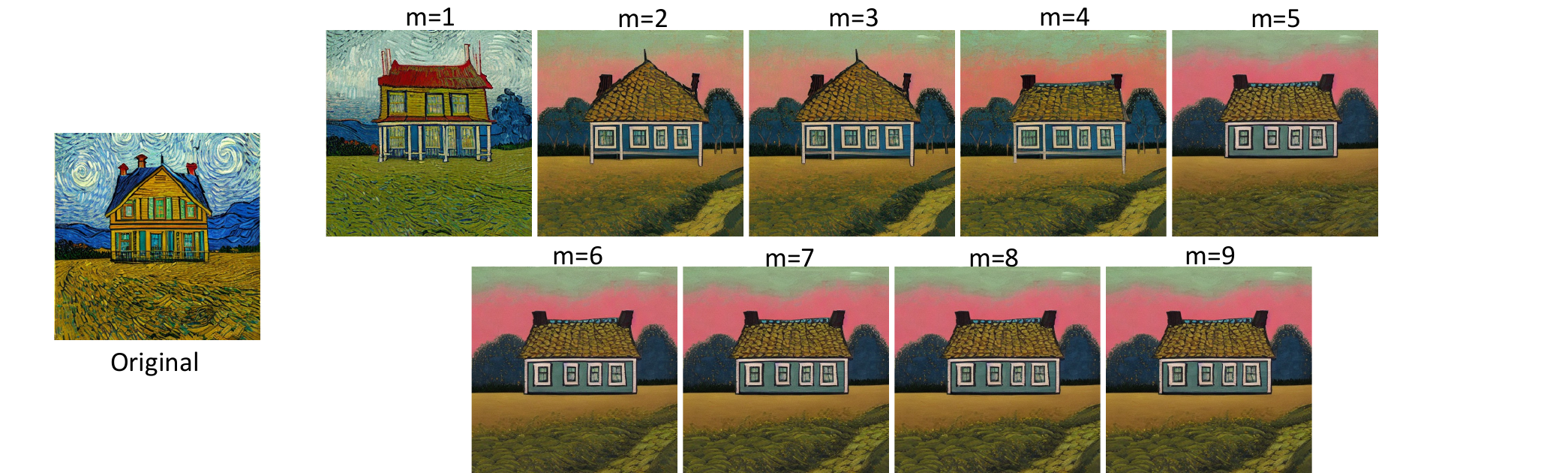}
  \vspace{-0.5cm}
    \caption{\label{sdxl_viz_3-new}
\textbf{Selection of $m$ for SD-v1-5.} At Layer 8, which encodes ``style'', we vary the value of $m$ from $1$ to the maximum value of $m$ for the given model (to $m=9$ for a total of 16 layers, starting from Layer 8). We find that at $m=2$, the style of {\it Van Gogh} for the prompt {\it A 'house in the style of Van Gogh'} is significantly removed. We therefore choose $m=2$ as the layers to edit. 
    }
\end{figure}
\begin{figure}
    \hskip 4cm
  \includegraphics[width=9cm, height = 6cm]{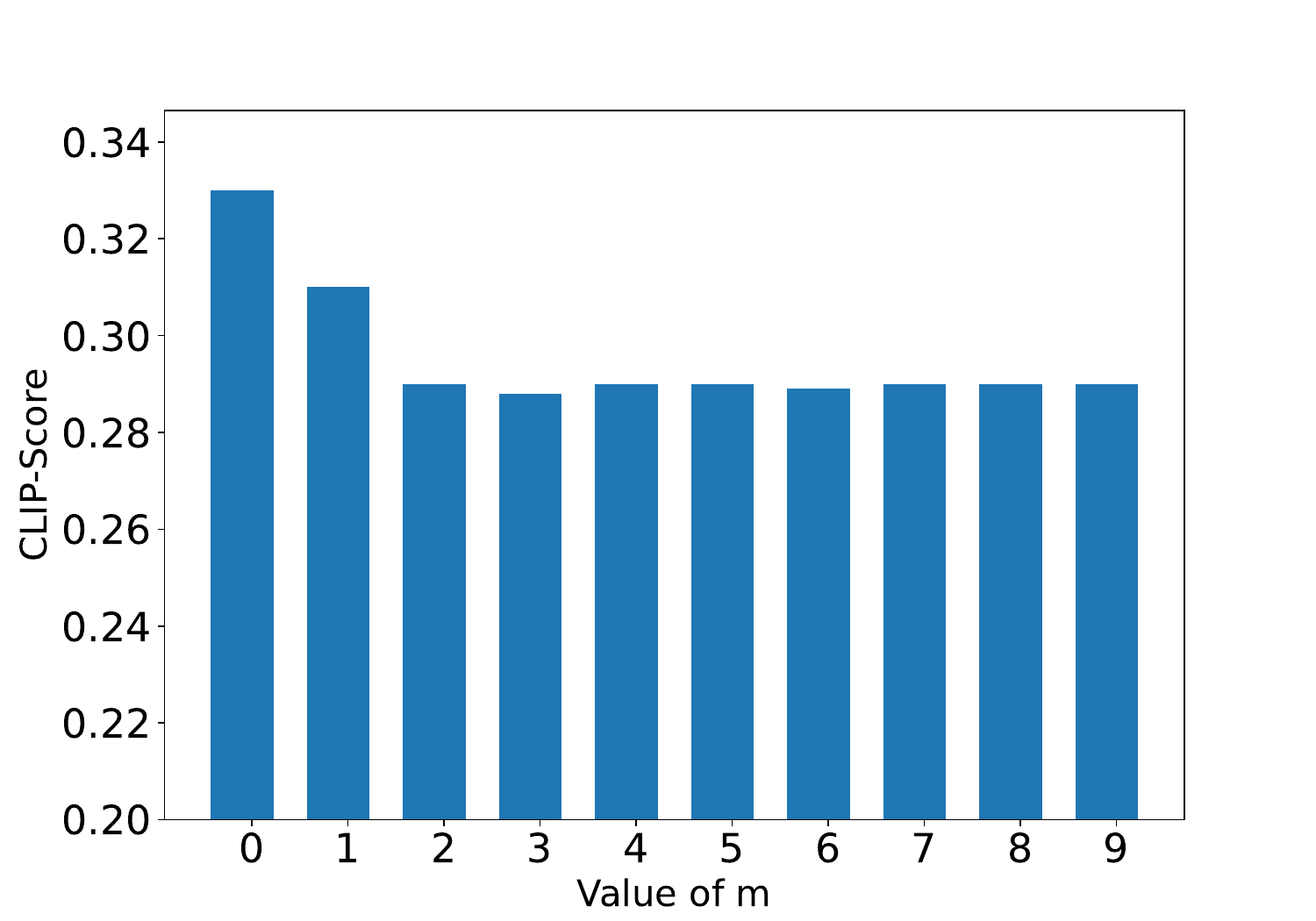}
  \vspace{-0.2cm}
    \caption{\label{sdxl_viz_3-sd}
\textbf{Selection of $m$ for SD-v1-5 - CLIP-Score.} m=0 corresponds to the original generation. We find that the CLIP-Score saturates after $m=2$, therefore we choose $m=2$ in our experiments. This figure is for the ``style" attribute; We choose the value of $m$ in a similar way for "objects" and ``facts", where the optimal $m$ comes out to be 2. 
    }
    \vspace{-0cm}
\end{figure}
\section{Model Editing Hyper-parameters}
We set the following hyper-parameters for $\lambda_{K}$ and $\lambda_{V}$ in~\crossedit{} as 0.01 for all the text-to-image models, as it led to the best editing results. 
\section{Comparison with Other Model Editing Baselines}
\label{compare}
\begin{figure}[H]
    \hskip 0.4cm
  \includegraphics[width=\columnwidth]{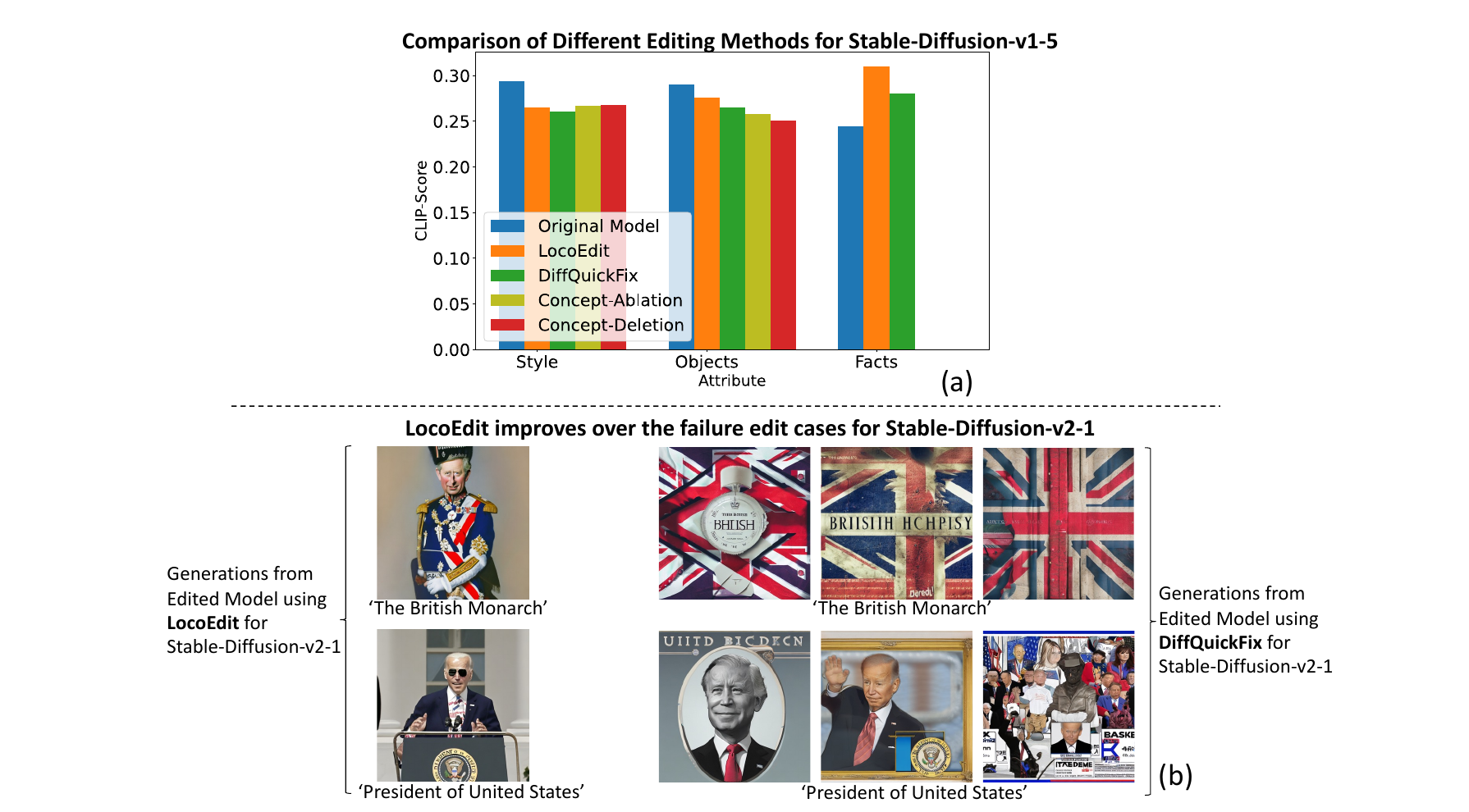}
  \vspace{-0.2cm}
    \caption{\label{baseline_comparison}
    \textbf{Comparison of our editing method when compared to other baselines.} (a) We compare our method with existing model editing methods -- DiffQuickFix~\citep{basu2023localizing}, Concept-Ablation~\citep{kumari2023ablating} and Concept-Deletion~\citep{gandikota2023erasing}. Amongst them only DiffQuickFix is a closed-form update method, whereas the other methods are fine-tuning based approaches. We find that~\crossedit{} has comparable performance to existing methods for style, better performance for facts and for objects slightly worse performance. The drop in CLIP-Score is lesser in our method as model edits via ~\crossedit{} modifies certain characteristics of the object, rather than the whole object which is enough to remove the trademarked characteristics of the object. For facts, we find that our method has a better increase in the CLIP-Score than~\citep{basu2023localizing} therefore highlighting effective factual edits. We note that both Concept-Ablation and Concept-Deletion do not perform factual editing. (b) Although DiffQuickFix from~\citep{basu2023localizing} was primarily built on Stable-Diffusion-v1 versions, we extend their framework to Stable-Diffusion-v2-1. In particular, we identify certain failure cases primarily on factual edits, where our method~\crossedit{} performs better. }
\end{figure}

Given that other model editing baselines~\citep{kumari2023ablating, gandikota2023erasing, basu2023localizing} primarily operate on SD-v1 versions, we compare our method with these baselines on SD-v1-5. In~\Cref{baseline_comparison} -- we provide a comprehensive comparison and analysis of how~\crossedit{} compares to other methods.

\section{Human-Study}
\label{sec:human-study-apx}
\begin{figure}
    \centering
    \vspace{1cm}
    \includegraphics[width=0.5\columnwidth]{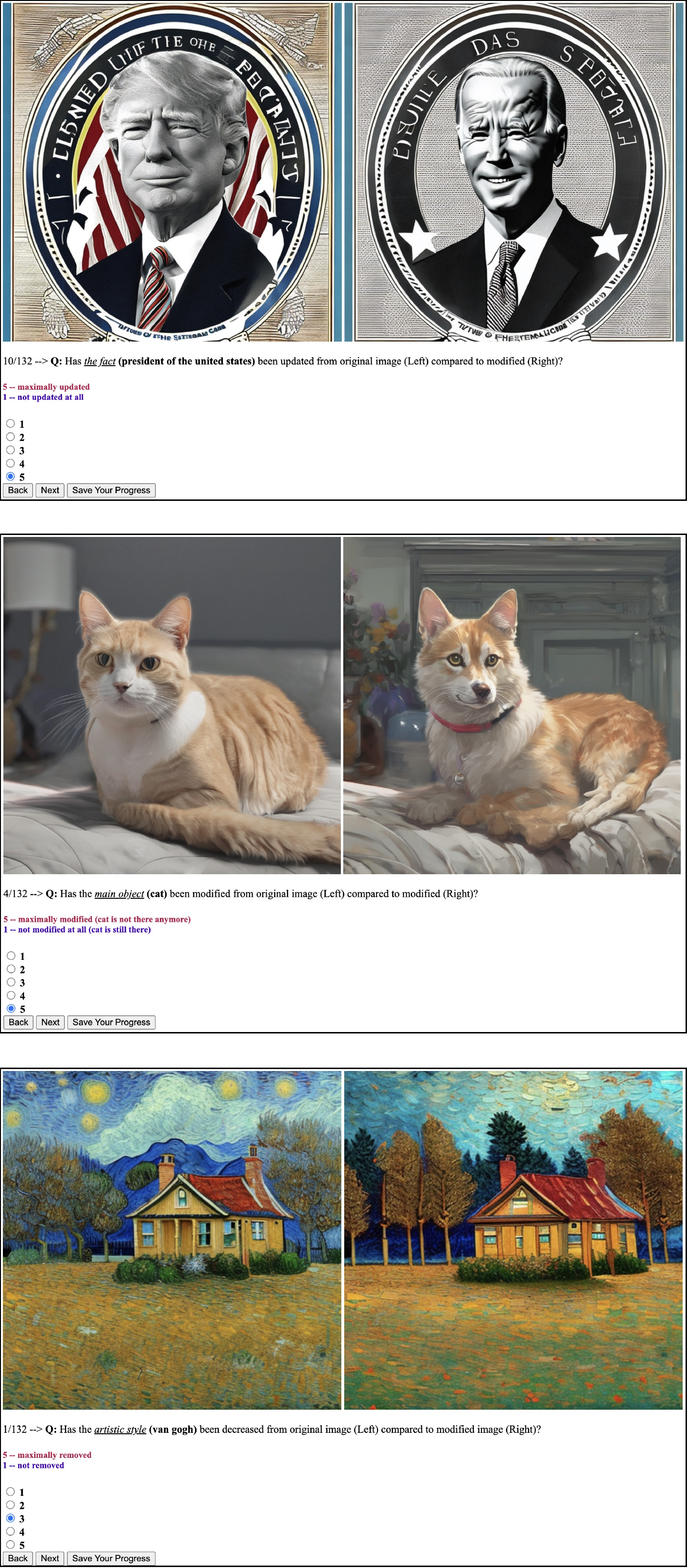}
    \caption{\label{fig:human-study-apx1}
    Examples of some tests that evaluators see in human-study.}%
\end{figure}

In this section, we report additional details about human experiments we did to verify \crossprompt{}.
Figure~\ref{fig:human-study-apx1} shows visualization of some pairs that we used in our experiments.
We used $11$ different prompts listed below (visual attribute is written in paranthesis) and $3$ generations per each prompt for each of the models.

\begin{itemize}[noitemsep,nolistsep]
    \item ``a house in the style of van gogh" \textbf{(style)}
    \item ``a town in the style of monet" \textbf{(style)}
    \item ``a town in the style of monet" \textbf{(style)}
    \item ``british monarch" \textbf{(fact)}
    \item   ``cat" \textbf{(object)}
    \item  ``elephant painting the style of salvador dali" \textbf{(style)}
    \item ``nemo" \textbf{(object)}
    \item ``president of the united states" \textbf{}
    \item ``rocks in the ocean in the style of monet" \textbf{(style)}
    \item ``snoopy" \textbf{(object)}
    \item ``women working in a garden in the style of van gogh" \textbf{(style)}
\end{itemize}

For each image pair based on the visual attribute of the corresponding prompt, we ask one of the following questions from evaluators and 
they have give a rating from $1$ to $5$.

\begin{itemize}
    \item \textbf{style}: Has the artistic style (`$<$artist name$>$') been decreased from original image (Left) compared to modified image (Right)?\\
    \textbf{$5$ --} \underline{maximally removed}; \ \ \ \  \textbf{$1$ --} \underline{not removed (modified image still has the artistic style)}.
    \item \textbf{object}: Has the main object (`$<$object name$>$') been modified from original image (Left) compared to modified (Right)? \\
    \textbf{$5$ --} \underline{maximally modified (other object is there)}; \ \ \ \  \textbf{$1$ --} \underline{not modified at all (object is still there)}.
    \item \textbf{fact}: Has the fact (`$<$fact name$>$') been updated from original image (Left) compared to modified (Right)? \\
    \textbf{$5$} -- \underline{maximally updated (other person is there)};\ \ \ \  \textbf{$1$ --} \underline{not updated at all (previous person is still there)}.
\end{itemize}

in $92.58\%$ of pairs, evaluators give a rating \textbf{above $1$} which shows that the edit is effective.
Now, we report more detailed scores for each of different attributes:
\begin{itemize}
    \item \underline{style}
    \begin{itemize}
        \item in $90.28\%$ of cases, evaluators give a rating of at least $2$.
        \item in $63.61\%$ of cases, evaluators give a rating of at least $3$.
        \item in $37.50\%$ of cases, evaluators give a rating of at least $4$.
        \item in $13.89\%$ of cases, evaluators give the rating $5$.
    \end{itemize}
    \item \underline{fact}
    \begin{itemize}
        \item in $100.00\%$ of cases, evaluators give a rating of at least $2$.
        \item in $95.83\%$ of cases, evaluators give a rating of at least $3$.
        \item in $86.67\%$ of cases, evaluators give a rating of at least $4$.
        \item in $59.17\%$ of cases, evaluators give the rating $5$.
    \end{itemize}
    \item \underline{object}
    \begin{itemize}
        \item in $92.22\%$ of cases, evaluators give a rating of at least $2$.
        \item in $81.11\%$ of cases, evaluators give a rating of at least $3$.
        \item in $62.22\%$ of cases, evaluators give a rating of at least $4$.
        \item in $39.44\%$ of cases, evaluators give the rating $5$.
    \end{itemize}
\end{itemize}

For different models, ratings are as follows:
\begin{itemize}
    \item \underline{SD-v1}
    \begin{itemize}
        \item evaluators give the rating of at least $2$ in $95.76\%$ of cases.
        \item evaluators give the rating of at least $3$ in $82.42\%$ of cases.
        \item evaluators give the rating of at least $4$ in $57.58\%$ of cases.
        \item evaluators give the rating of at least $5$ in $32.73\%$ of cases.
    \end{itemize}
    \item \underline{SD-v2}
    \begin{itemize}
        \item evaluators give the rating of at least $2$ in $87.27\%$ of cases.
        \item evaluators give the rating of at least $3$ in $61.82\%$ of cases.
        \item evaluators give the rating of at least $4$ in $43.03\%$ of cases.
        \item evaluators give the rating of at least $5$ in $13.33\%$ of cases.
    \end{itemize}
    \item \underline{SD-XL}
    \begin{itemize}
        \item evaluators give the rating of at least $2$ in $90.91\%$ of cases.
        \item evaluators give the rating of at least $3$ in $69.70\%$ of cases.
        \item evaluators give the rating of at least $4$ in $49.70\%$ of cases.
        \item evaluators give the rating of at least $5$ in $31.52\%$ of cases.
    \end{itemize}
    \item \underline{OpenJourney}
    \begin{itemize}
        \item evaluators give the rating of at least $2$ in $96.36\%$ of cases.
        \item evaluators give the rating of at least $3$ in $83.03\%$ of cases.
        \item evaluators give the rating of at least $4$ in $62.42\%$ of cases.
        \item evaluators give the rating of at least $5$ in $38.79\%$ of cases.
    \end{itemize}
\end{itemize}

Above results shows that \crossprompt{} is effective and works better for facts and for models OpenJourney and SD-v1.

\section{Neuron-Level Model Editing}
In this section we discuss more details about experiments for neuron-level model editing.

\subsection{Prompts}
\label{sec:zero-shot-apx1}
In this section, we provide more details about the set of prompts used to report CLIP-Score in Figure~\ref{zero-shot-edit-score}.
\begin{itemize}
    \item \textbf{Van Gogh}
    \begin{itemize}
        \item ``a painting of rocky ocean shore under the luminous night sky in the style of Van Gogh"
        \item ``A painting of lone figure contemplates on a cliff, surrounded by swirling skies in the style of Van Gogh"
        \item ``Majestic mountains take on an ethereal quality in sky, painted by Van Gogh"
        \item ``Two trees in a sunlit field, painted by Van Gogh"
        \item ``A flower-filled field in the style of Van Gogh"
        \item ``painting of a river on a warm sunset in the style of Van Gogh"
        \item ``painting of olive trees in the style of Van Gogh"
        \item ``painting of a field with mountains in the background in the style of Van Gogh"
    \end{itemize}
    \item \textbf{Monet}
    \begin{itemize}
        \item ``painting of women working in the garden in the style of Monet",
        \item ``rocks in the ocean, in the style of Monet",
        \item ``a painting of a city in the style of Monet",
        \item ``a painting of a river in the style of Monet",
        \item ``Monet style painting of a person on a cliff",
        \item ``a painting of a town, in the style of Monet",
        \item ``a painting of a sunset, in the style of Monet",
        \item ``a painting of mountains, in the style of Monet",
        \item ``Monet style painting of flowers in a field",
        \item ``a painting of a landscape in the style of Monet",
        \item ``two trees in a field, painting in the style of Monet"
    \end{itemize}
    \item \textbf{Salvador Dali}
    \begin{itemize}
        \item ``the persistence of memory painting in the style of Salvador Dali",
        \item ``the elephant painting in the style of Salvador Dali",
        \item ``soft construction with boiled beans painting in the style of Salvador Dali",
        \item ``galatea of the spheres painting in the style of Salvador Dali",
        \item ``the temptation of st. anthony painting in the style of Salvador Dali",
        \item ``swans reflecting elephants painting in the style of Salvador Dali",
        \item ``enigma of desire painting in the style of Salvador Dali",
        \item ``slave market with the disappearing bust of voltaire painting of Salvador Dali",
        \item ``the meditative rose painting in the style of Salvador Dali",
        \item ``melting watch painting in the style of Salvador Dali",
    \end{itemize}
    \item {\textbf{Jeremy Mann}}
    \begin{itemize}
        \item ``In the style of Jeremy Mann, a view of a city skyline at sunset, with a warm glow spreading across the sky and the buildings below",
        \item ``In the style of Jeremy Mann, an urban scene of a group of people gathered on a street corner, captured in a moment of quiet reflection",
        \item ``In the style of Jeremy Mann, a surreal composition of floating objects, with a dreamlike quality to the light and color",
        \item ``In the style of Jeremy Mann, a view of a city street at night, with the glow of streetlights and neon signs casting colorful reflections on the wet pavement",
        \item ``In the style of Jeremy Mann, a moody, atmospheric scene of a dark alleyway, with a hint of warm light glowing in the distance",
        \item ``In the style of Jeremy Mann, an urban scene of a group of people walking through a park, captured in a moment of movement and energy",
        \item ``In the style of Jeremy Mann, a landscape of a forest, with dappled sunlight filtering through the leaves and a sense of stillness and peace",
        \item ``In the style of Jeremy Mann, a surreal composition of architectural details and organic forms, with a sense of tension and unease in the composition",
        \item ``In the style of Jeremy Mann, an abstract composition of geometric shapes and intricate patterns, with a vibrant use of color and light",
        \item ``In the style of Jeremy Mann, a painting of a bustling city at night, captured in the rain-soaked streets and neon lights",
    \end{itemize}
    \item \textbf{Greg Rutkowski}
    \begin{itemize}
        \item ``a man riding a horse, dragon breathing fire, painted by Greg Rutkowski",
        \item ``a dragon attacking a knight in the style of Greg Rutkowski",
        \item ``a demonic creature in the wood, painting by Greg Rutkowski",
        \item ``a man in a forbidden city, in the style of Greg Rutkowski",
        \item ``painting of a group of people on a dock by Greg Rutkowski",
        \item ``a king standing, with people around in a hall, painted by Greg Rutkowski",
        \item ``two magical characters in space, painting by Greg Rutkowski",
        \item ``a man with a fire in his hands in the style of Greg Rutkowski",
        \item ``painting of a woman sitting on a couch by Greg Rutkowski",
        \item ``a man with a sword standing on top of a pile of skulls, in the style of Greg Rutkowski",
    \end{itemize}
    \item \textbf{Pablo Picasso}
    \begin{itemize}
        \item ``Painting of nude figures by Pablo Picasso",
        \item ``painting of a grieving woman in the style of Pablo Picasso",
        \item ``painting of three dancers by Pablo Picasso",
        \item ``portrait of a girl in front of mirror, painted by Pablo Picasso",
        \item ``painting of a bird, in the style of Pablo Picasso",
        \item ``painting of a blind musician, by Pablo Picasso",
        \item ``painting of a room in the style of Pablo Picasso",
        \item ``painting of an acrobat in performance, by Pablo Picasso",
    \end{itemize}
\end{itemize}

Prompts that we used to compute z-score are same as above except that for each artist, we put the artist name in all of the above prompts.
To generate prompts withput any artistic style, we remove artist name from the prompts listed above.

\subsection{Visualization}
\label{sec:zero-shot-apx2}
\begin{figure}
    \centering
  \includegraphics[width=0.7\columnwidth]{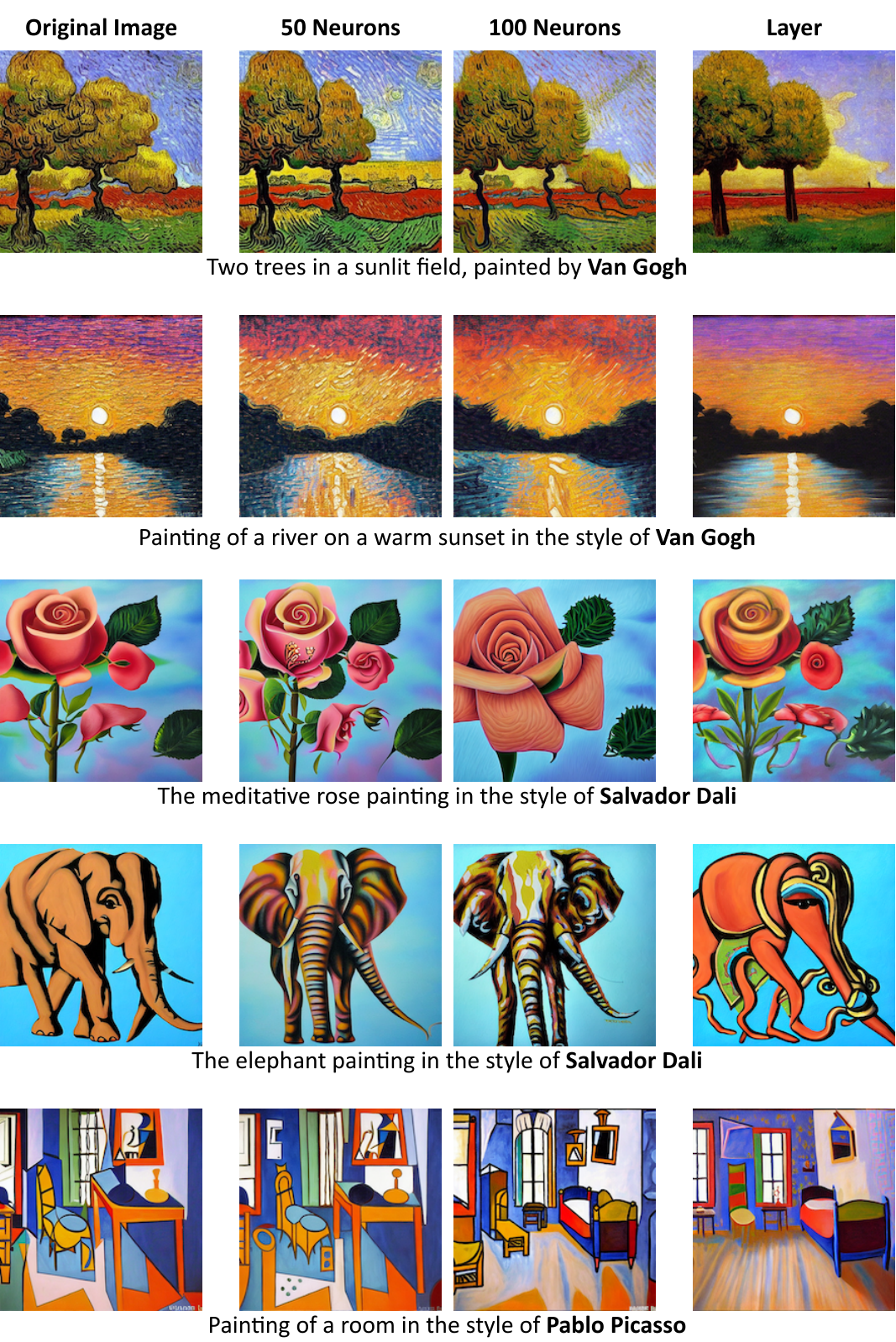}
    \caption{\label{zero-shot-edit-apx1}
    \textbf{Neuron Level Model Editing - Qualitative}. Results when applying \textbf{neuron-level dropout} on identified neurons in layers specified with \crossprompt on Stable Diffusion v1.5.
    Each row corresponds to an input text prompt featuring a particular artistic style.
    First column shows the image generated without any intervention while second and third column visualize images when $50$ and $100$ neurons out of $1280$ neurons in controlling layers are modified, respectively.
    Last column shows images when a different embedding is given to controlling layers. 
    }%
\end{figure}

\begin{figure}
    \centering
  \includegraphics[width=0.7\columnwidth]{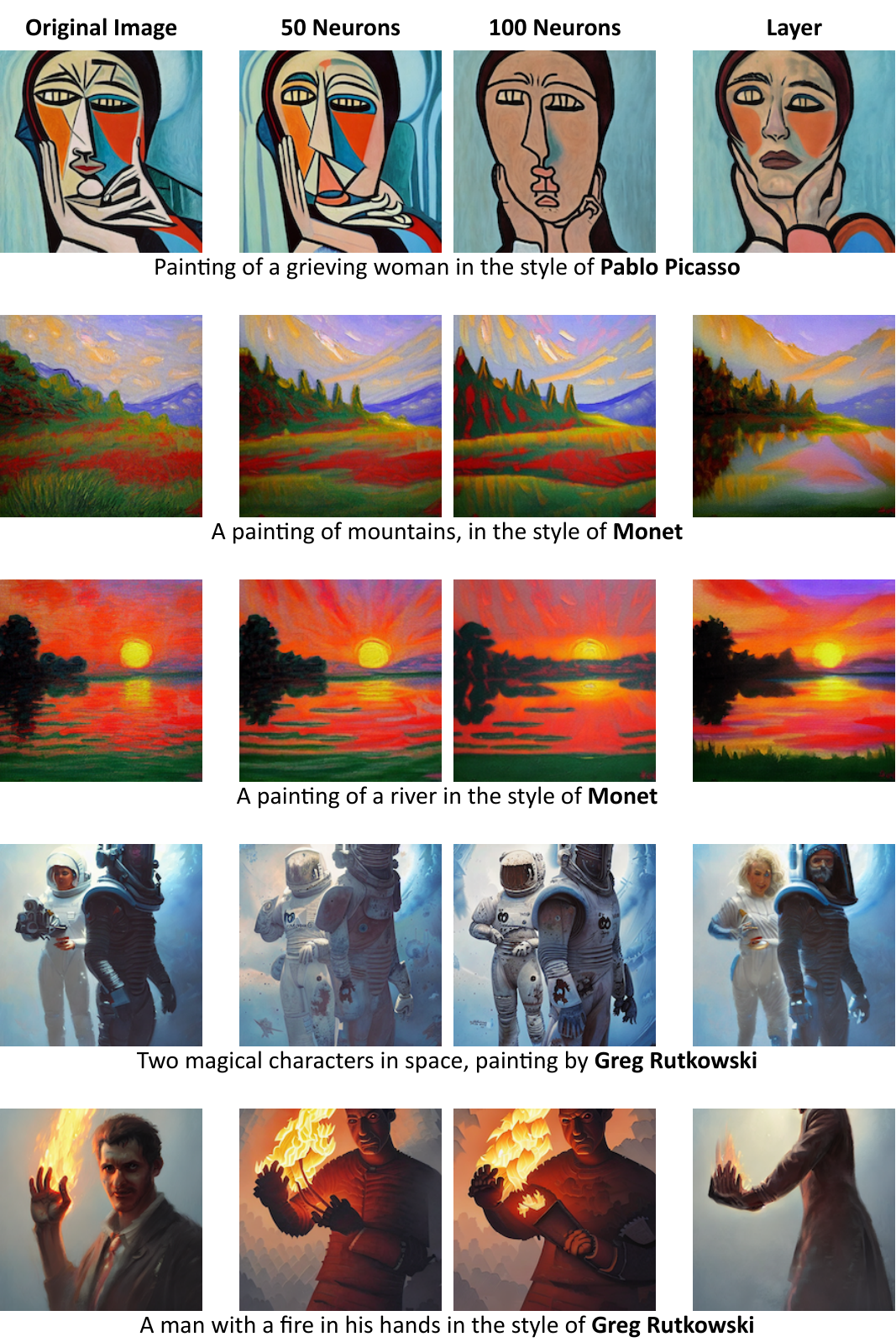}
    \caption{\label{zero-shot-edit-apx2}
    \textbf{Neuron Level Model Editing - Qualitative}. Results when applying \textbf{neuron-level dropout} on identified neurons in layers specified with \crossprompt on Stable Diffusion v1.5.
    Each row corresponds to an input text prompt featuring a particular artistic style.
    First column shows the image generated without any intervention while second and third column visualize images when $50$ and $100$ neurons out of $1280$ neurons in controlling layers are modified, respectively.
    Last column shows images when a different embedding is given to controlling layers. 
    }%
\end{figure}
\section{Robustness of the Edited Model}
\label{robustness}
In this section, we discuss the robustness of the edited model to generic prompts. In particular, we curate a set of 320 prompts from MS-COCO with 80 objects and 4 locations ("beach", "forest", "city", "house") for each. We then compare the average CLIP-Score of the generations from original model vs. edited model across "style" and "object" attributes. For edited models, we compare our method with DiffQuickFix, as it also edits models using a closed-form update. Overall from~\Cref{clip_score_surrounding} -- we find that~\crossedit{} does not harm generic prompts, highlighting the robustness of our method. 
\begin{table}
\centering
\begin{tabularx}{0.7\textwidth}   { 
  | >{\raggedright\arraybackslash}X 
  | >{\centering\arraybackslash}X 
  | >{\raggedleft\arraybackslash}X | }

 \hline
 \textbf{Editing Method} & \textbf{Style Edited Model}  & \textbf{Object Edited Model} \\
 \hline
 \text{No Editing}  & 30.04  & 30.04  \\
 \hline 
  \text{DiffQuickFix}  & 29.61  & 29.32  \\
\hline
\text{LocoEdit(Ours)}  & 29.99  & 29.40  \\
\hline 
\end{tabularx}
\caption{\label{clip_score_surrounding} \textbf{CLIP-Score of the generated images with the original prompt.} We find that while both DiffQuickFix~\citep{basu2023localizing} and our method leads to a slight drop in the CLIP-Score for the edited model, the decrease in the CLIP-Score from our method is slightly lesser -- thereby highlighting that our localized editing method does not affect generic prompts significantly. }
\end{table}
\section{DeepFloyd Edit Limitations}
\label{deepfloyd_limitations}
\begin{figure}
    \centering
    \begin{subfigure}[t]
    \centering
    \includegraphics[width=0.9\textwidth]{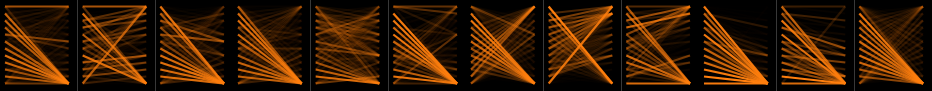}
    \end{subfigure}
    \begin{subfigure}[t]
    \centering
    \includegraphics[width=0.4\textwidth]{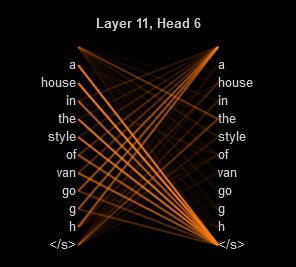}
    \end{subfigure}    
    \caption{\label{fig:attention-map} Attention map from the last layer of T5 encoder for the prompt ``a house in the style of van gogh''. Figure (a) shows the attention map between the tokens in the prompt for each of the 12 heads in the last layer of T5 encoder. The non causal nature of the mask is evident from the significant attention weights from later tokens to previous tokens in the prompt. This is more easily observed in the Figure (b) which highlights the same attention map from head 6.}
\end{figure}

While~\crossprompt{} is adept at localizing knowledge in DeepFloyd, we find that the closed-form edits are not suitable for the DeepFloyd model. In particular, we find that performing edits using~\crossedit{} at the locations identified by~\crossprompt{} does not lead to semantically correct edits. We hypothesize that this is due to the difference in the text-encoder between [SD-v1, SD-v2, OpenJourney, SDXL] and DeepFloyd. Text-to-image models (except DeepFloyd) consist of text-encoders (e.g., CLIP) which utilizes a causal attention mechanism. Given that closed-form edits require a mapping between the last-subject token embedding to a target embedding, there is less information leakage compared to a T5 text-encoder which implements a bi-directional self-attentio (see~Figure. \ref{fig:attention-map} for a visualization of this non-causal attention). In the case of T5, mapping {\it only} the last-subject token embedding to a target embedding might be insufficient and mapping all the tokens can lead to huge informational loss from the model. Below we provide some examples from editing the DeepFloyd model -- we primarily observe non-semantic generations from the edited model. Designing fast editing methods for DeepFloyd like models which utilizes bi-directional attention is an important future course of study. 
\begin{figure}[H]
    \centering
  \includegraphics[width=0.7\columnwidth]{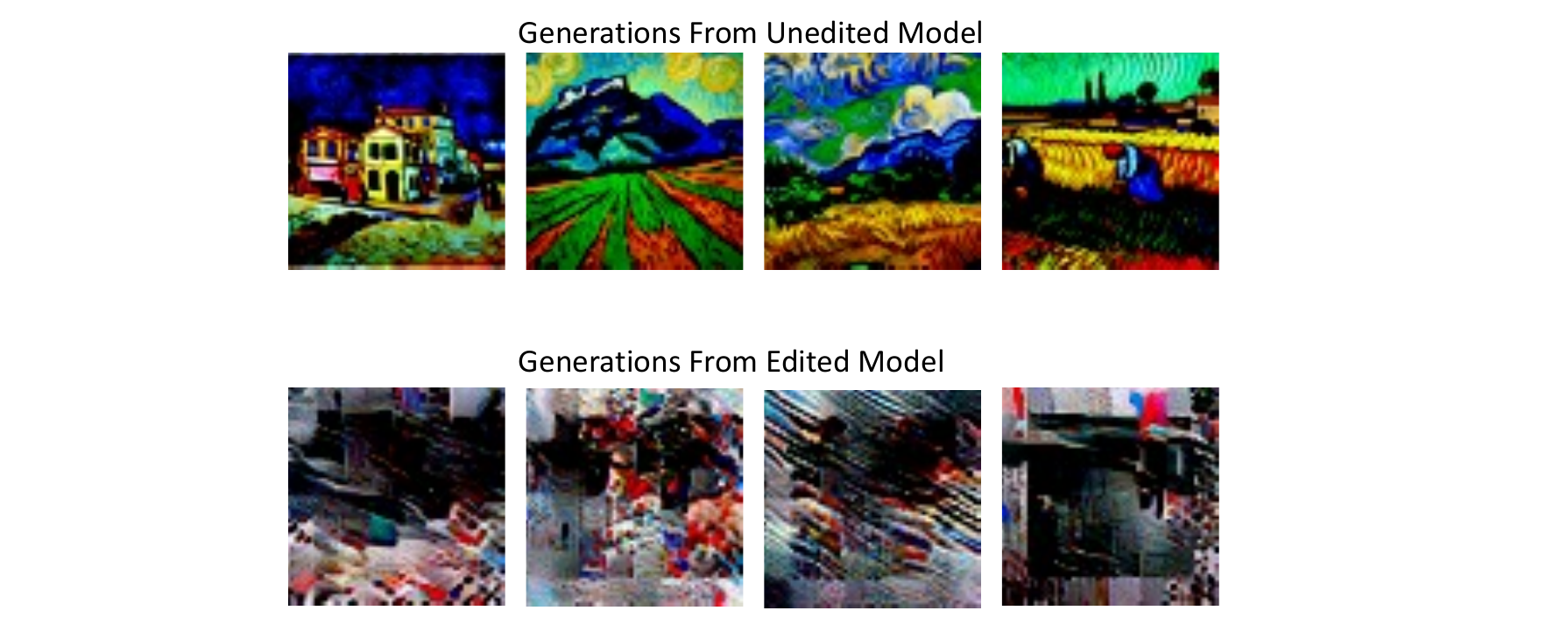}
    \caption{\label{zero-shot-edit-apx-deepfloyd}
    \textbf{Model Editing for DeepFloyd.} We find that the edited DeepFloyd model leads to generations of non-semantic outputs. In this figure, we show results for editing ``style". However, we observe similar results for other attributes such as "objects" and ``facts". Prompts used are corresponding to {\it Van Gogh}. 
    }%
\end{figure}

%

\end{document}